\documentclass{article}

\usepackage{amsmath,amsfonts,bm}

\def\vx{{\bm{x}}}

\PassOptionsToPackage{square,numbers, sort}{natbib}

\usepackage[final]{neurips_2024}

\usepackage[utf8]{inputenc} 
\usepackage[T1]{fontenc}    
\usepackage{hyperref}       
\usepackage{url}            
\usepackage{booktabs}       
\usepackage{amsfonts}       
\usepackage{nicefrac}       
\usepackage{microtype}      
\usepackage[dvipsnames]{xcolor} 

\usepackage{multirow}
\usepackage{graphicx}
\usepackage{amsmath}
\usepackage{xspace} 
\newcommand{\ourmethod}{\textsc{MotionCraft}\xspace}
\newcommand{\virg}[1]{``#1''}
\newcommand{\bigO}[1]{\mathcal{O}\left( #1 \right)}
\AtBeginDocument{\colorlet{defaultcolor}{.}}
\usepackage[linesnumbered,ruled,vlined]{algorithm2e}
\SetNlSty{bfseries}{\color{defaultcolor}}{}
\usepackage{subcaption}

\title{\ourmethod:\\ Physics-based Zero-Shot Video Generation}

\author{
Luca Savant Aira\footnotemark[1] \And Antonio Montanaro\footnotemark[1] \AND Emanuele Aiello \And Diego Valsesia \And Enrico Magli \AND \\
Politecnico di Torino
\\
{\tt\small \{name.surname\}@polito.it}
}

\begin{document}

\maketitle
\def\thefootnote{*}\footnotetext{indicates equal contribution.}\def\thefootnote{\arabic{footnote}}
\def\thefootnote{}\footnotetext{Project page: \url{https://mezzelfo.github.io/MotionCraft/}.}\def\thefootnote{\arabic{footnote}}

\begin{abstract}

Generating videos with realistic and physically plausible motion is one of the main recent challenges in computer vision. 
While diffusion models are achieving compelling results in image generation, video diffusion models are limited by heavy training and huge models, resulting in videos that are still biased to the training dataset. In this work we propose MotionCraft, a new zero-shot video generator to craft physics-based and realistic videos. MotionCraft is able to warp the noise latent space of an image diffusion model, such as Stable Diffusion, by applying an optical flow derived from a physics simulation. We show that warping the noise latent space results in coherent application of the desired motion while allowing the model to generate missing elements consistent with the scene evolution, which would otherwise result in artefacts or missing content if the flow was applied in the pixel space.
We compare our method with the state-of-the-art Text2Video-Zero reporting qualitative and quantitative improvements, demonstrating the effectiveness of our approach to generate videos with finely-prescribed complex motion dynamics.
\end{abstract}

\section{Introduction}
\label{sec:intro}

As human beings, we have always exploited our creativity to generate art, in different forms such as visual art, music or poetry. In vision, we are often inspired by the natural world since our visual system continuously acquire images perceived as a video sequence. Indeed, videos or movies are one of the best visual stimuli since they contain images, motion and audio. 

Recent generative models for still images based on diffusion models \cite{sohl2015deep, song2020denoising, rombach2022high} achieved remarkable results with quality almost indistinguishable from real images. It is therefore clear that the next big goal is video generation. However, it seems that including the dimension of time remains challenging. Some works such as Sora \cite{videoworldsimulators2024} achieve astonishing temporal consistency and photorealism at the expense of enormous computational and data requirements. Moreover, we argue that fine-grained control over the motion dynamics is impossible with a simple text prompt. If one wants to synthesize a video according to some precise physical dynamics, they would not be able to do it with current models. Interestingly, explicitly controlling the motion dynamics also allows to decouple temporal evolution from content generation. Indeed, explicitly injecting the physics of the real world as motion dynamics allows to develop more parsimonious models, that do not need to brute-force learn them from data.

For this reason, in this paper, we investigate the possibility to create a zero-shot video generation model that only requires a pretrained still image generator and knowledge of physical laws regarding motion. Indeed, since videos are temporal sequences of images correlated by physical laws, we only need to devise a way to include physical laws in the diffusion prior to animate a starting image.
We thus advocate for physics simulators as appropriate sources of motion, output as a sequence of optical flows, while also being completely user-controllable, plausible, and explainable. 

We propose \ourmethod, a physics-based zero-shot video generator that uses optical flow extracted from a physical simulation to warp the noise latent space of a pretrained image diffusion model to generate videos with complex dynamics without the need to train anything. 
While using a projection of motion onto the camera plane as a pixelwise displacement field (optical flow) may seem limiting due to the fact that, if applied in the pixel space, it would not be able to synthesise novel coherent content but only displace pixels, the trick lies in its application in the noise latent domain. Backed by evidence that motion vectors correlate between pixel and noise space, warping of the latter by means that \ourmethod allows to simultaneously apply the desired motion and exploit the powerful image prior of the generative model. This is capable of adapting the scene to the prescribed motion without significant artefacts, generate novel content and shows impressive global consistency (reflections, illumination, etc., consistent with the desired evolution).

We present quantitative and qualitative experimental results where we show that our zero-shot \ourmethod is capable of synthesising realistic videos with finely controlled temporal evolution governed by fluid-dynamics equations, rigid body physics, and multi-agent interaction models, while zero-shot state-of-art techniques cannot. 

\begin{figure*}[t]
\centering
\begin{subfigure}[t]{\textwidth}
    \centering
    \includegraphics[width=0.16\textwidth]{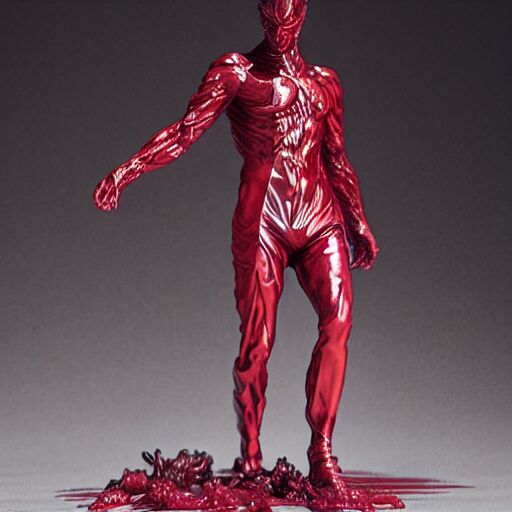}
    \includegraphics[width=0.16\textwidth]{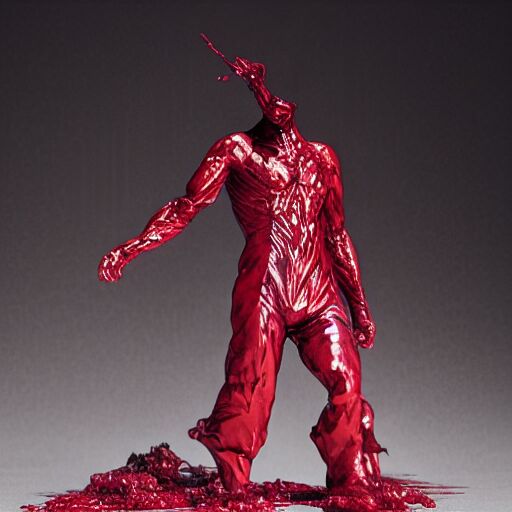}
    \includegraphics[width=0.16\textwidth]{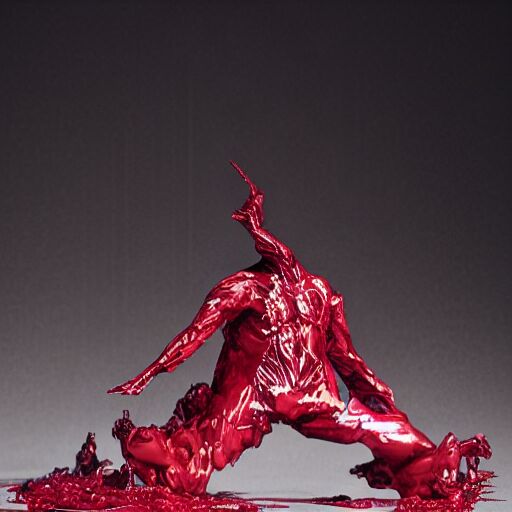}
    \includegraphics[width=0.16\textwidth]{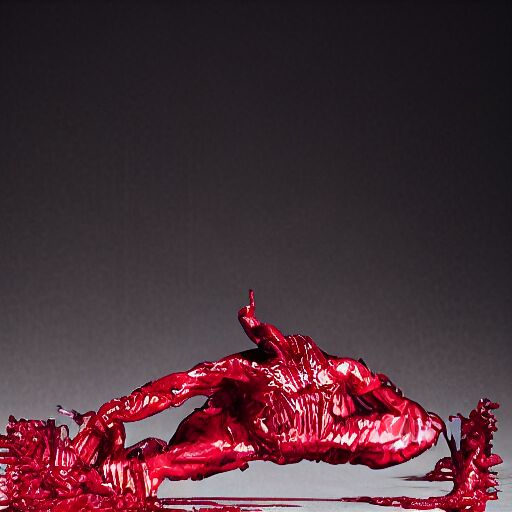}
    \includegraphics[width=0.16\textwidth]{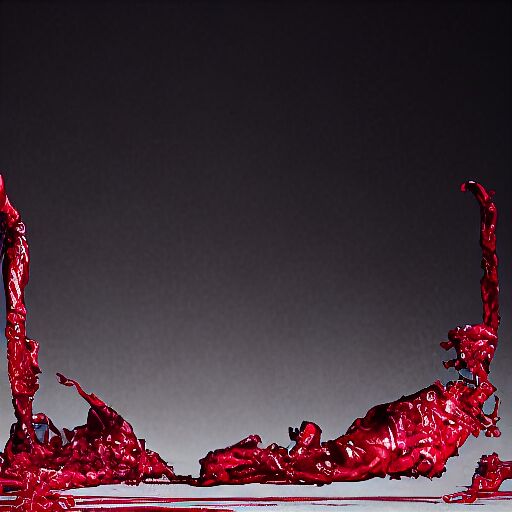}
    \includegraphics[width=0.16\textwidth]{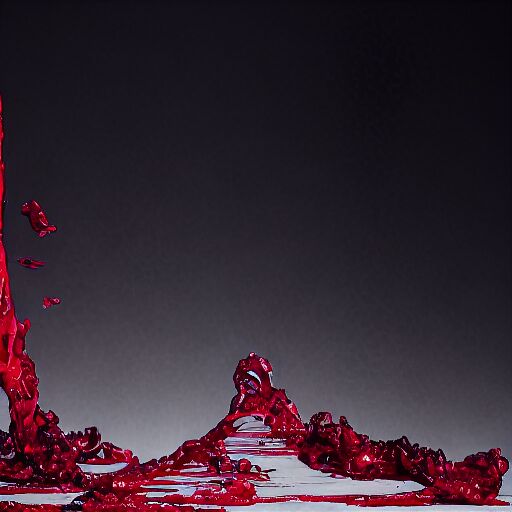}
\end{subfigure}\hfill
\begin{subfigure}[t]{\textwidth}
    \centering
    \includegraphics[width=0.16\textwidth]{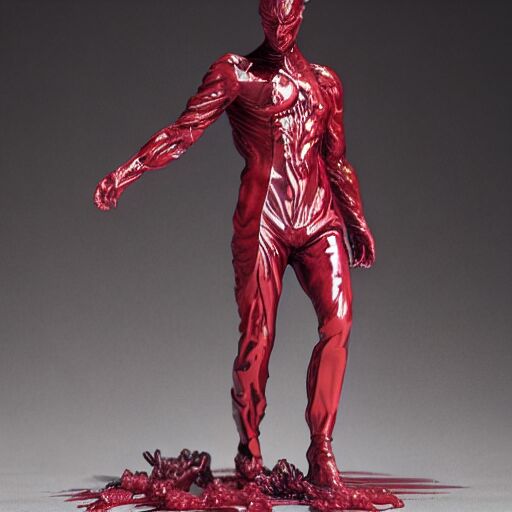}
    \includegraphics[width=0.16\textwidth]{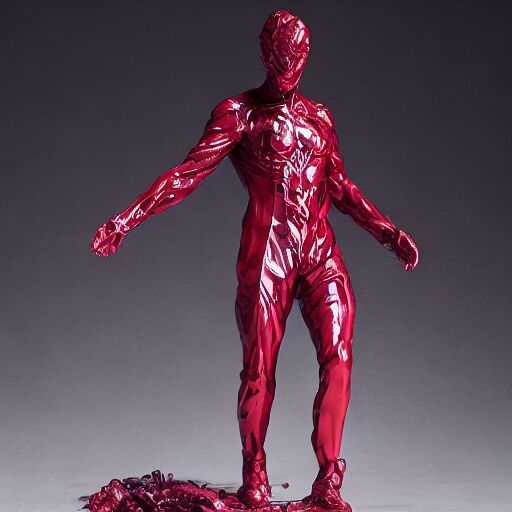}
    \includegraphics[width=0.16\textwidth]{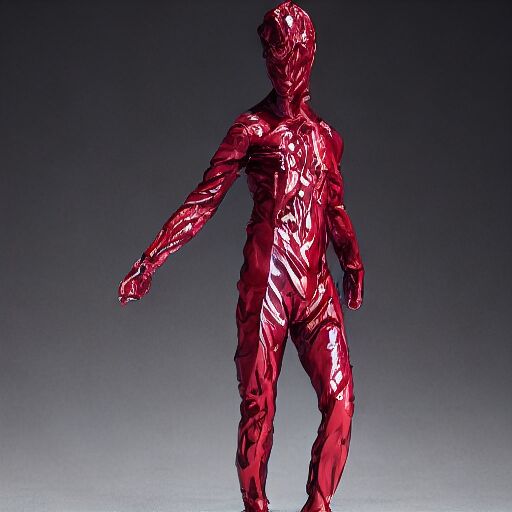}
    \includegraphics[width=0.16\textwidth]{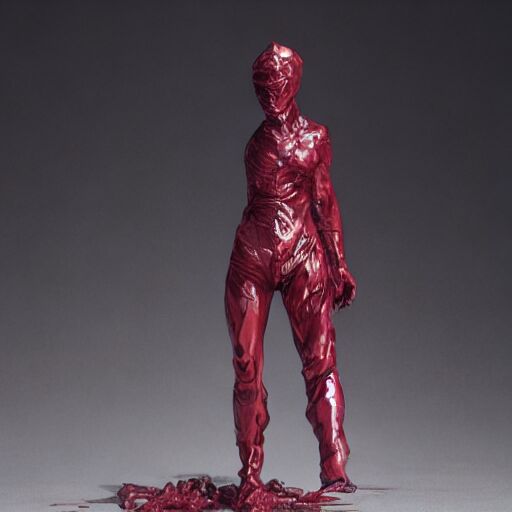}
    \includegraphics[width=0.16\textwidth]{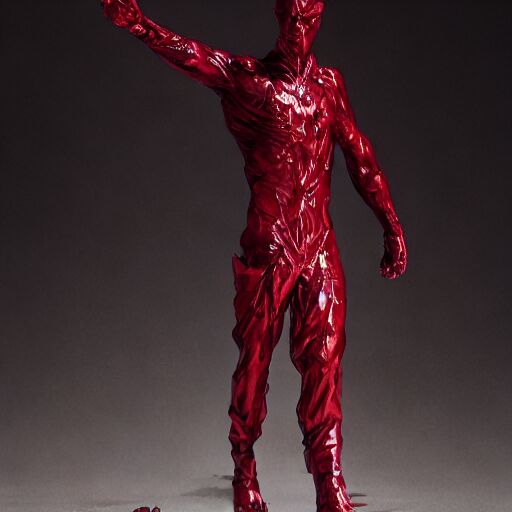}
    \includegraphics[width=0.16\textwidth]{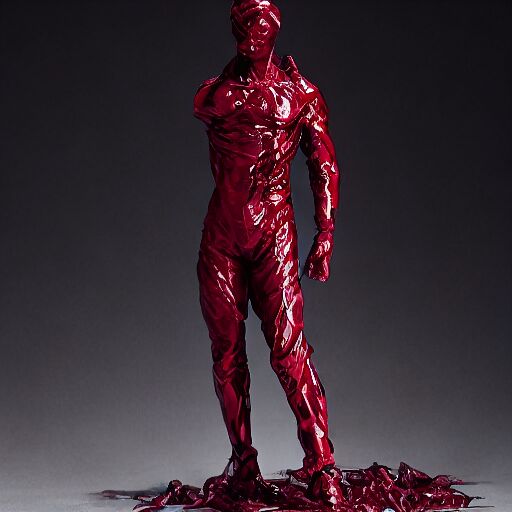}
\end{subfigure}
\caption{Melting man simulation. Top: \ourmethod; Bottom: T2V0 \cite{khachatryan2023text2video}. \ourmethod uses a fluid dynamics simulation to warp noise latents and synthetize video frames.  T2V0 is unable to simulate the evolution of the melting statue and simply moves the object towards the bottom of the frame.}
\label{fig:meltingman}
\end{figure*}

\section{Related work}
\label{sec:related_work}
\paragraph{Diffusion Based Video Generation}
Video Generation \cite{video_survey} is a longstanding problem in computer vision aiming to learn the distribution of and synthesise realistic videos. Recently, text-based Denoising Diffusion Probabilistic  Models (DDPM) \cite{sohl2015deep, song2020score} have been studied to tackle this challenge delivering impressive results. 
These approaches include Sora \cite{videoworldsimulators2024},  Video Diffusion models \cite{ho2022video}, Imagen-video \cite{ho2022imagen} and Align your Latents \cite{blattmann2023align}. They require sophisticated spatio-temporal denoising architectures at the expense of huge computational requirements and large amounts of paired text-video data for training. 
To reduce the data requirements, different approaches investigate few-shot and unsupervised learning techniques. Make-a-Video \cite{singer2022make} proposes an unsupervised training with only videos, coupled with a retrieval strategy to sample using text.
On the other hand, Ni et al. \cite{ni2023conditional} train a diffusion-based optical flow generator that outputs a flow conditioned on a reference image and a textual prompt, that reduces the computational burden of generating videos by training the diffusion process on small flow fields. Differently from them, our approach is zero-shot and we do not train anything. 

To the best of our knowledge,  Text-to-video-Zero  \cite{khachatryan2023text2video}  and Generative Rendering \cite{cai2023generative} are the only zero-shot video generators. 
However, Generative Rendering (concurrent work, with no code available) has significant extra requirements beyond Stable Diffusion (SD) as image generator, in the form of a depth-conditioned ControlNet \citep{zhang2023adding}, and a 3D mesh manually animated, leveraging UV maps to render the scene. Moreover, Generative Rendering cannot render fluids, since they are difficult to represent as 3D meshes.  

In this paper, we compare our method to Text-to-video-Zero (T2V0), as zero-shot video generator baseline. T2V0 applies a constant shift (with a fixed direction) to the initial latent noise of SD, sampling each frame sequentially by means of DDPM.
As shown in our work, since the motion in the noise latent space directly translates into the motion of the pixel space, the generated videos result in a overall shift in the same fixed direction. The largest part of the motion is caused by the stochastic fluctuations of the DDPM  sampling strategy leading to unnatural motion and inconsistency of the objects in the different frames. On the contrary, in this work, we avoid the use of a constant warping operation derived from physics simulation flows in the latent space in order to incorporate complex motion dynamics.

\paragraph{Diffusion Based Video and Image Editing}
Recently, different methods exploit the prior of text-to-image diffusion models for video editing.
In particular, Tune-A-Video \cite{wu2023tune} finetunes a text-to-image diffusion model to edit a video. They start from the inverted frames in the latent space and use the text prompt as an editing tool.
Pix2Video \cite{ceylan2023pix2video} employs a self-attention injection mechanism to edit videos using a pretrained image diffusion model.

Other methods use the optical flow to edit reference images or videos. Motion Guidance \cite{geng2024motion} leverages a user defined optical flow that allow zero-shot image editing. It works by guiding the diffusion sampling process with the gradient from a pretrained optical flow network via a guidance loss.  LatentWarp \cite{bao2023latentwarp} and TokenFlow \cite{geyer2023tokenflow}, use an optical flow estimated from a reference video to warp the latent space of the diffusion model to achieve consistent editing. These methods leverage both diffusion models priors and other components such as ControlNet for structural control, and trained flow estimators such as RAFT \cite{teed2020raft}.   
Alternatively, we propose a zero-shot video generation method, using only vanilla SD. This means that \ourmethod does not require a reference video but it can animate an image, generated by the SD model or obtained by inverting a real one. Moreover, the physics simulations allow to generate different videos from the same starting image. 

\section{Method}
\label{sec:method}

This section describes \ourmethod, a zero-shot video generation method, where the meaning of \virg{zero-shot} is twofold: we do not train or finetune any component of the text-to-image diffusion model, nor we do not use  reference video or optical flow estimators as starting point. In the following, we used used Stable Diffusion as pretrained text-to-image model.

\subsection{Optical Flow is preserved in the Latent Space of Stable Diffusion}

\begin{figure}[t]
    \centering
    \includegraphics[width=0.98\textwidth]{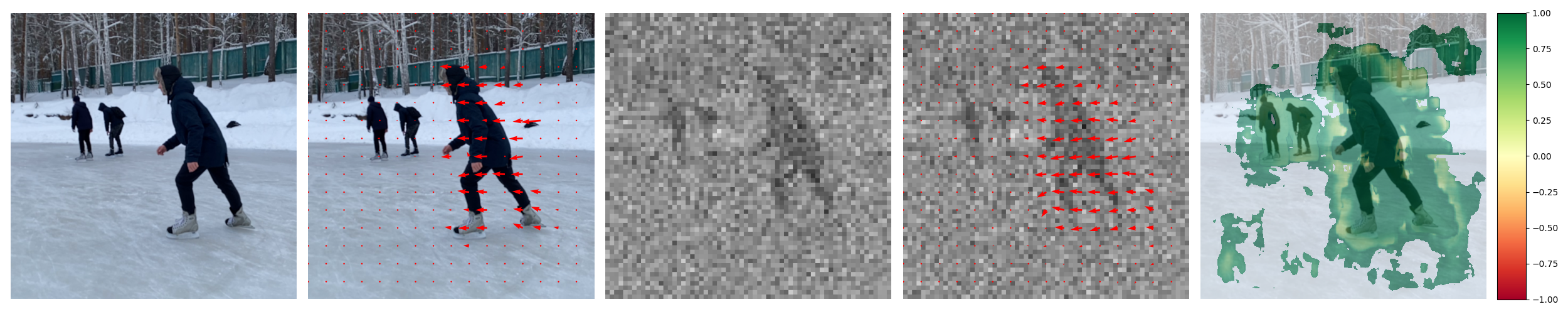}
    \caption{A qualitative example of the image and latent flows correlation. This figure shows, from left to right, (a) the first RGB frame, (b) the second RGB frame superimposed with the estimated flow in the RGB domain, (c) the first latent frame, (d) the second latent frame superimposed with the estimated flow in the latent domain and (e) the correlation map of the two non-zero flows.}
    \label{fig:correlation_qualitative}
\end{figure}

Our proposed method stems from a key observation: the optical flow estimated between two frames in the pixel space is correlated with the flow estimated between the corresponding noise latent representations of SD. We conjecture that this is related to the specific design of the SD variational auto-encoder and denoiser architectures. In fact, by largely using convolution operations, they enforce a locality prior which preserves spatial information to some extent.

In order to empirically investigate this phenomenon, we conducted a quantitative experiment using the MSU Video Frame Interpolation Benchmark dataset \cite{MSUDataset}, considering only real videos. For each pair of consecutive video frames, the following steps have been taken. We first estimate the optical flow in the RGB space by using a well-established method, based on the Gunnar Farneback's algorithm, provided by OpenCV \cite{itseez2015opencv}. Then, we compute the noise latent representations of the two frames, first encoding the image in the variational autoencoder (VAE) of SD  at timestep $\tau=0$, followed by DDIM inversion \cite{song2020denoising} up to timestep $\tau=400$ (same value for all experiments in this work, empirically determined). Finally, a correlation coefficient based on cosine similarity is computed between the optical flows estimated in the RGB and noise latent spaces.
The resulting correlations are then averaged across all pairs of consecutive frames in the dataset, obtaining an average value of 0.727, which indicates a strong correlation between the optical field in the RGB and noise latent domains. An example of this experiment is presented in Fig. \ref{fig:correlation_qualitative}, showcasing the two estimated flows in the image and latent space and their correlation.

\subsection{Physics-based zero shot video generation}

\begin{figure}
    \centering
    \includegraphics[width=0.8\textwidth]{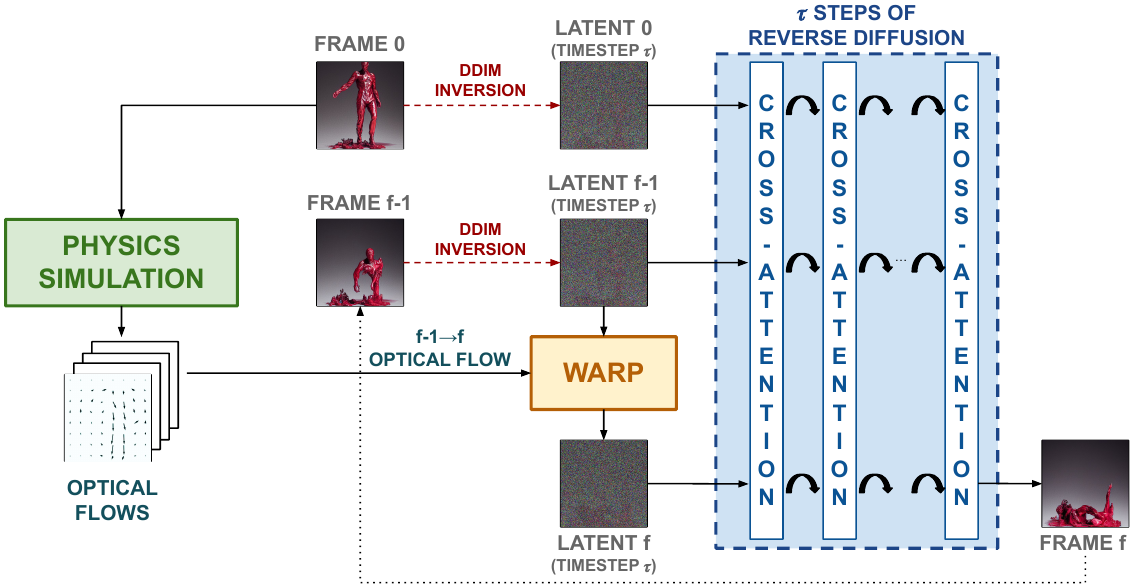}
    \caption{\ourmethod overview. A video is generated from a starting image using a pretrained still image generative model by warping noise latents according to an optical flow description of the motion to be synthesised.}
    \label{fig:method_arch}
\end{figure}

\begin{algorithm}[t]
\caption{Pseudocode of \ourmethod}
\label{alg:H}
\SetAlgoLined
\SetKwInOut{Input}{Input}
\SetKwInOut{Output}{Output}

\Input{$I^0, \mathcal{W}, \eta, \mathcal{P}, \mathcal{P}_{\emptyset}$}
\Output{$I^0, \dots, I^{N-1}$}

\For{$f=1$ \KwTo $N-1$}{
    \textcolor{CadetBlue}{
    $z^{f-1}_0 = \mathcal{E}(I^{f-1})$ \tcp*{Encode the frame}
    \For(\tcp*[f]{Inversion loop}){$t = 0$ \KwTo $\tau-1$}{
        $\hat{\epsilon} \gets \epsilon_t(z_{t}^{f-1}, \mathcal{P}; \{z_{t}^{f-1}\})$ \tcp*{Self-Attention, No MCFA}
        $z_{t+1}^{f-1} \gets \text{DDIMInversion}_{t \to t+1}(z_{t}^{f-1}, \hat{\epsilon}, 0)$ \tcp*{$\eta=0 \iff$ DDIM}
    }}
    
    \textcolor{DarkOrchid}{
    $\zeta_{\tau}^{f} = \mathcal{W}^{f-1 \to f}(z_{\tau}^{f-1})$ \tcp*{Warp the latent}
    }

    \textcolor{Sepia}{
    \For(\tcp*[f]{Generation loop}){$t=\tau-1$ \KwTo $0$}{
        $\hat{\epsilon}_{\mathcal{P}} \gets \epsilon_t(\zeta^{f}_{t+1}, \mathcal{P} ; \{z^0_t, z^{f-1}_t\})$ \tcp*{MCFA with $I^0$ and $I^{f-1}$}
        $\hat{\epsilon}_{\emptyset} \gets \epsilon_t(\zeta^{f}_{t+1}, \mathcal{P}_{\emptyset} ; \{z^0_t, z^{f-1}_t\})$ \tcp*{MCFA with $I^0$ and $I^{f-1}$}
        $\hat{\epsilon} \gets \hat{\epsilon}_{\emptyset} + \gamma (\hat{\epsilon}_{\mathcal{P}}-\hat{\epsilon}_{\emptyset})$ \tcp*{Classifier-free guidance}
        $\zeta^{f}_t \gets \textrm{DDIM}_{t+1 \to t}(\zeta^{f}_{t+1}, \hat{\epsilon}, \eta^{f})$ \tcp*{Perform Spatial-$\eta$}
    }
    $I^{f} \gets \mathcal{D}(\zeta^{f}_0)$ \tcp*{Decode the latent}
    }
}
\Return{$I^0, \dots, I^{N-1}$}
\end{algorithm}

Based on the analysis presented in the previous section, we propose a novel zero-shot video generation method, named \ourmethod, where an image (real or generated), serving as a starting frame $I^0$, is animated according to a physical simulation, by means of a (possibly time-varying) optical-flow generator $\mathcal{W}$ in the noise latent space. The outcome is a video made of $N$ frames $I^0,\dots,I^{N-1}$ that follows the motion prescribed by the physical simulation and evolves  the content of the first frame coherently. Inspired by the previous observation, this animation is obtained by warping the noisy latent representation of an image in the latent diffusion space.  Regarding the physics simulation for the optical flow generation, we use different libraries to simulate different physics, as explained in the experimental section, such as fluid dynamics, rigid motion and multi-agent systems. It is also possible, albeit not shown in this paper, to use animation software to generate the required optical flows.

Fig. \ref{fig:method_arch} illustrates an overview of \ourmethod highlighting the autoregressive generation of the video. At each iteration $f \geq 1$, the frame $I^{f}$ is generated using only the information contained in the first frame $I^0$ and the previous frame $I^{f-1}$. Given this Markovian structure, \ourmethod is characterized by $\bigO{1}$ space complexity and $\bigO{N}$ time complexity with respect to the total number $N$ of frames to be generated. More in detail, first, the two RGB frames $I^0, I^{f-1}$ are encoded into the latent space and they are independently inverted with the reversed $\text{DDIM}$ sampling scheme up to a fixed diffusion timestep $\tau$, obtaining $z_{\tau}^0$, and $z_{\tau}^{f-1}$, respectively. Then, the optical flow warping operator $\mathcal{W}^{f-1 \to f}$ prescribed by the physical simulation is applied to $z_{\tau}^{f-1}$, obtaining $\zeta_{\tau}^{f}$. Finally, the next RGB frame $I^{f}$ is generated by performing $\tau$ steps of reverse diffusion using the DDIM sampling scheme with a novel cross-frame attention mechanism and a novel spatial noise map $\eta^{f}$ weighting technique, explained below. Furthermore, we exploit the classifier-free guidance (CFG) technique for generation proposed in \cite{ho2022classifier}, with $\mathcal{P}$ and $\mathcal{P}_{\emptyset}$ being the positive and negative prompt, respectively, and $\gamma > 1$ being the strength of the CFG. More details can be found in the Appendix \ref{sec:background}.

Algorithm \ref{alg:H} reports the pseudocode of \ourmethod. Lines $2-6$ include the DDIM inversion up to timestep $\tau$. Starting current frame $I^{f-1}$ that was previously generated, in line $2$ we embed it with the VAE encoder $\mathcal{E}$, obtaining $z^{f-1}_0$.
Then we apply DDIM inversion on $z^{f-1}_0$ for $\tau$ timesteps (line $3-6$). This involves the UNet  with the standard self-attention (note the repetition of the noisy latent $z^{f-1}_t$) and the positive prompt $\mathcal{P}$.
As briefly reported in \cite{mokady2023null}, we have also experienced that DDIM inversion is not compatible with CFG; hence, during the inversion, we do not use the negative prompt $\mathcal{P}_{\emptyset}$. The resulting estimated noise is used in line $5$ for applying the DDIM inversion step (note that the $\eta=0$, so pure DDIM is performed). Upon completion of the DDIM inversion process, we obtain $z_{\tau}^{f-1}$, the noisy latent corresponding to the frame $I^{f-1}$.

In line $7$, the optical flow warping operator $\mathcal{W}^{f-1 \to f}$ is applied to the noisy latent of the current frame $z_{\tau}^{f-1}$ to obtain a new noisy latent $\zeta_{\tau}^f$ that will generate the successive frame. 
Finally, in lines $8-14$, the frame is generated. During this generation phase we use CFG to increase the quality of the generated images, hence also the negative prompt $\mathcal{P}_{\emptyset}$ is used. To create new content while preserving the original image, we propose two direct generalization of two known techniques: the multiple cross-frame attention (MCFA) mechanism and a spatial noise map weighting (Spatial-$\eta$).

The MCFA technique generalizes the Cross Frame Attention (CFA) \cite{khachatryan2023text2video}, as it enables the to-be-generated frame to attend to an arbitrary number of frames. We choose to attend to the first frame and the previous frame (as shown in lines $9-10$ of Alg. \ref{alg:H} and Fig. \ref{fig:method_arch}) to ensure long-range and short-range temporal consistency, respectively. MCFA intervenes in all the self-attention blocks of the SD UNet, by replacing the keys and values, that are originally computed from projections of the generating frame features, with the ones computed from the attended frames.

We also propose Spatial-$\eta$ (line $12$), that is a novel technique that enables to choose, on a pixel-by-pixel basis, whether to use DDIM or DDPM as a sampling scheme. This enables the usage of DDPM in regions of the images where novel content should be created (for example, when a new part of an object is entering the scene), while using DDIM in the other regions to ensure consistency and determinism where the already-present content is just moving. Note that this spatial map $\eta^{f}$ can be obtained in multiple ways from the physical simulation. For example, $\eta^{f}$ can be set to $1$ in regions of the image where the flow is not well-defined (pointing outside of the image boundaries) or in regions where the optical flow field has discontinuities.

\section{Experimental results}
\label{sec:results}

\subsection{Experimental setting}
\label{subsec:experimental}

In this section, we show examples of video generation based on different physics simulations: rigid body motion, fluid dynamics and multi-agent systems. Given an optical flow, we apply it on the SD latent space using \ourmethod (code is available at \url{https://mezzelfo.github.io/MotionCraft/}). Then, we compare our method to Text2Video-Zero \cite{khachatryan2023text2video} that, to the best of our knowledge, is the only diffusion-based zero-shot method for video generation.

\begin{figure*}[t]
\centering
\begin{subfigure}[t]{\textwidth}
    \centering
    \includegraphics[width=0.16\textwidth]{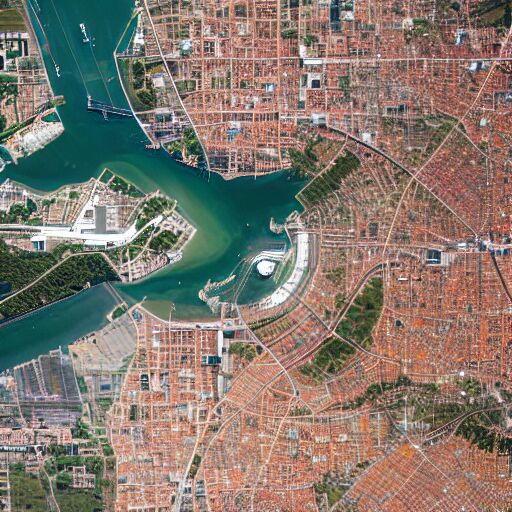}
    \includegraphics[width=0.16\textwidth]{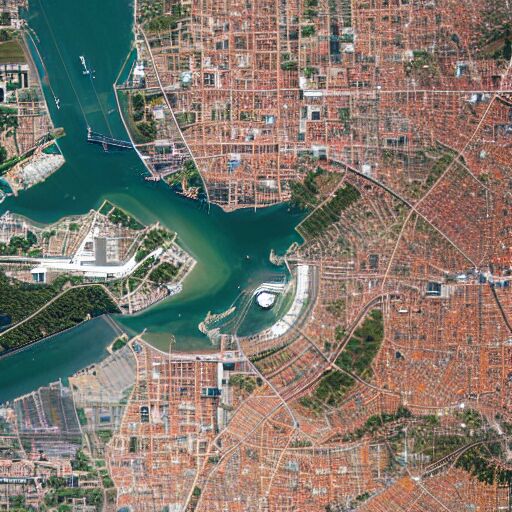}
    \includegraphics[width=0.16\textwidth]{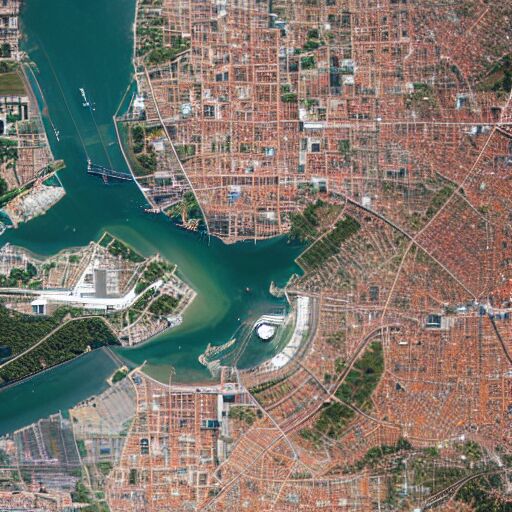}
    \includegraphics[width=0.16\textwidth]{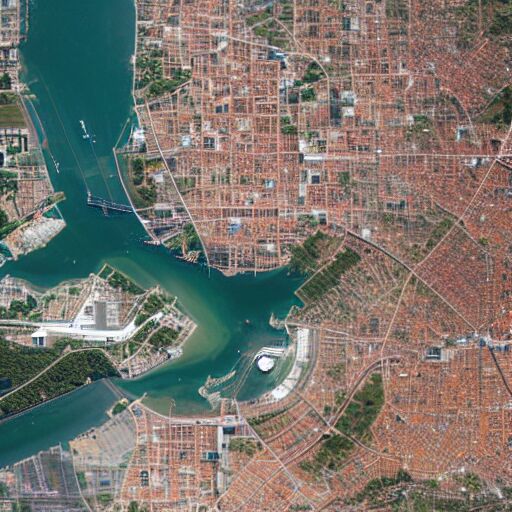}
    \includegraphics[width=0.16\textwidth]{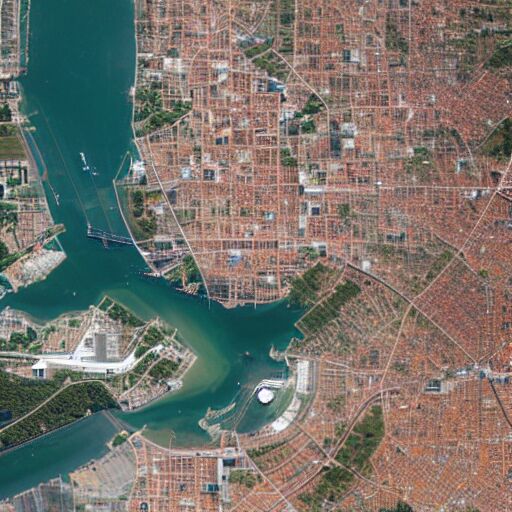}
    \includegraphics[width=0.16\textwidth]{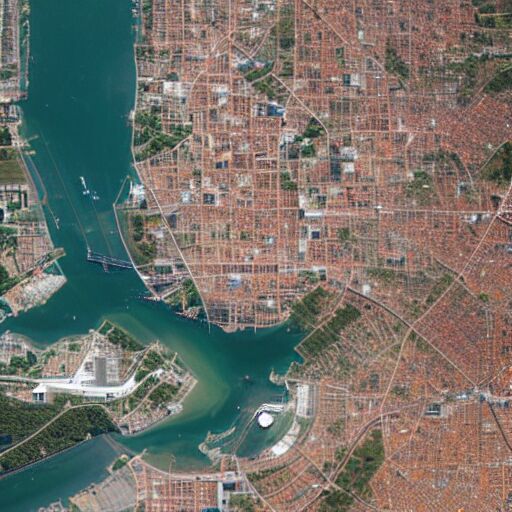}
\end{subfigure}\hfill
\begin{subfigure}[t]{\textwidth}
    \centering
    \includegraphics[width=0.16\textwidth]{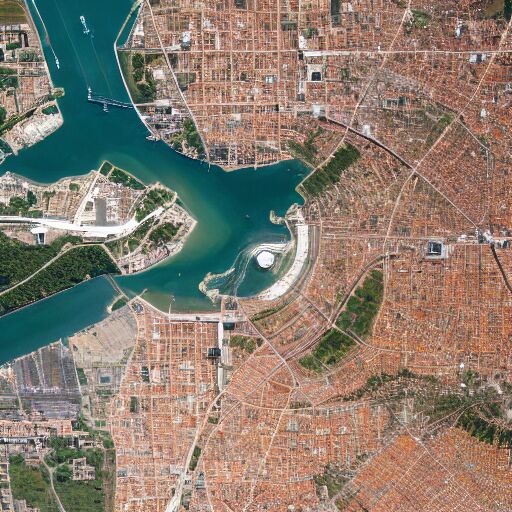}
    \includegraphics[width=0.16\textwidth]{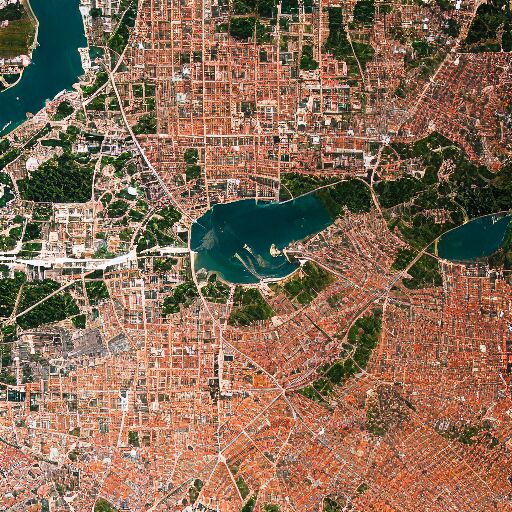}
    \includegraphics[width=0.16\textwidth]{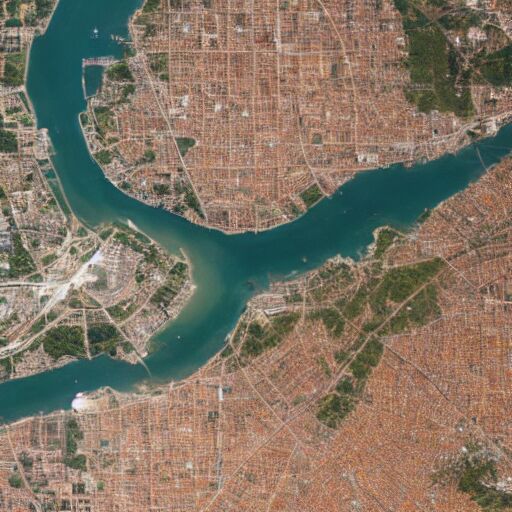}
    \includegraphics[width=0.16\textwidth]{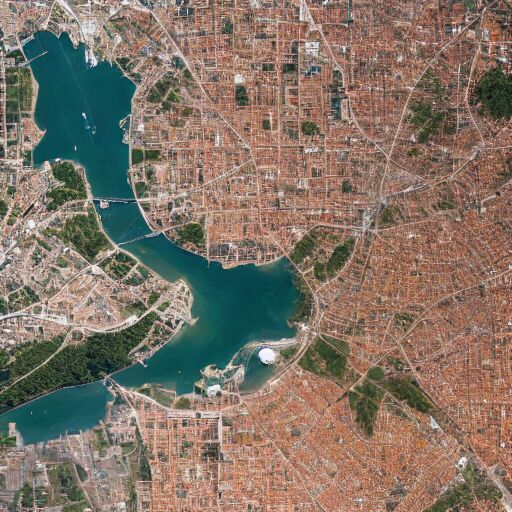}
    \includegraphics[width=0.16\textwidth]{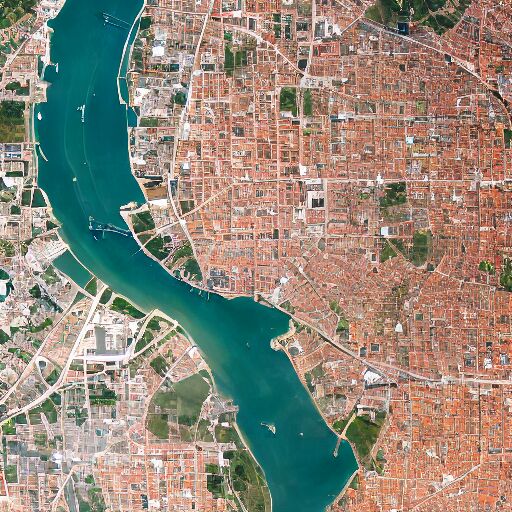}
    \includegraphics[width=0.16\textwidth]{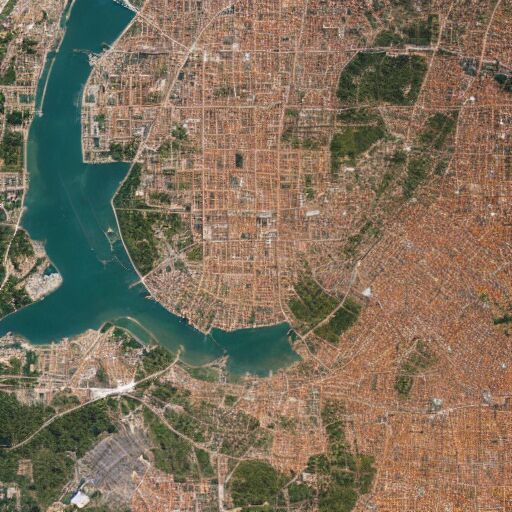}
\end{subfigure}
\caption{Rigid motion simulation: satellite orbit. Top: \ourmethod; Bottom: T2V0 \cite{khachatryan2023text2video}.}
\label{fig:satellite}
\end{figure*}

\begin{figure*}[t]
\centering
\begin{subfigure}[t]{\textwidth}
    \centering
    \includegraphics[width=0.16\textwidth]{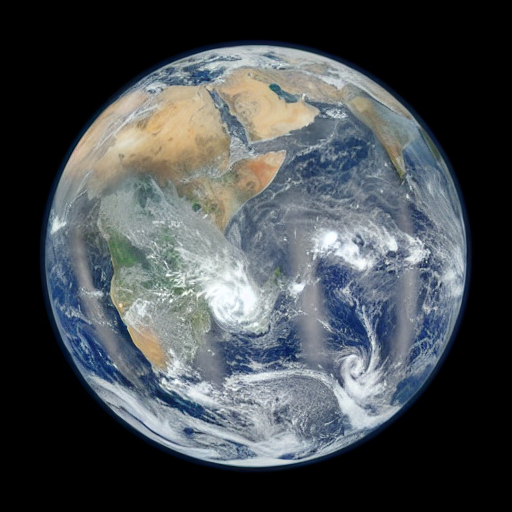}
    \includegraphics[width=0.16\textwidth]{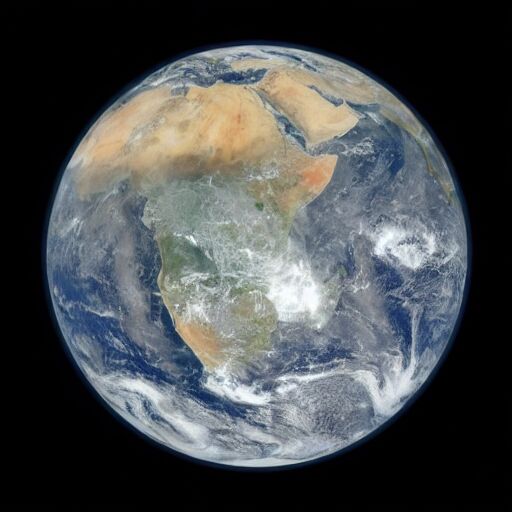}
    \includegraphics[width=0.16\textwidth]{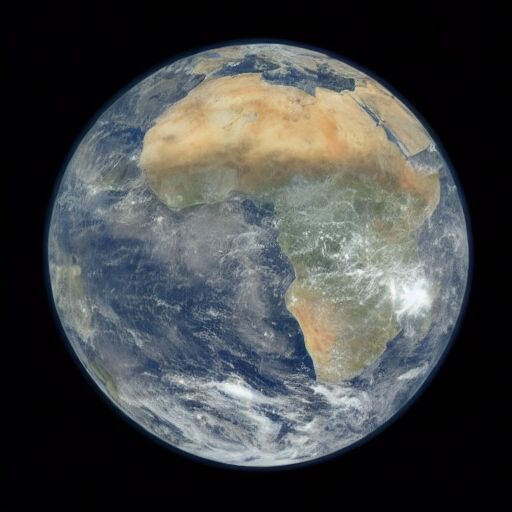}
    \includegraphics[width=0.16\textwidth]{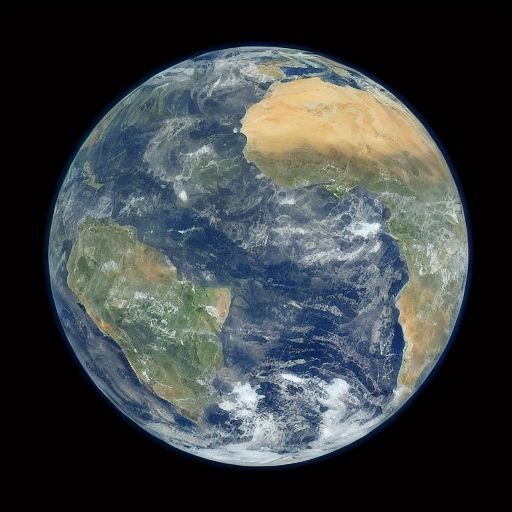}
    \includegraphics[width=0.16\textwidth]{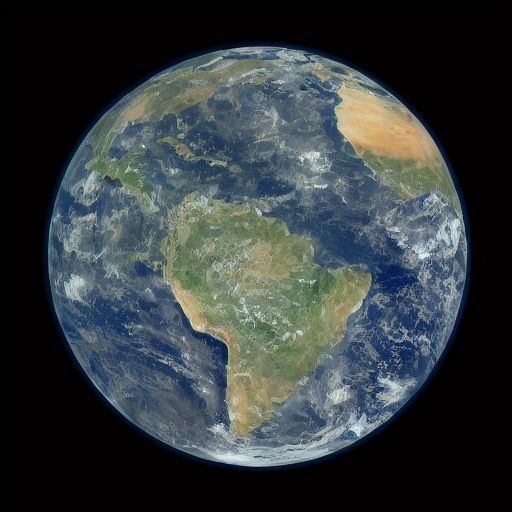}
    \includegraphics[width=0.16\textwidth]{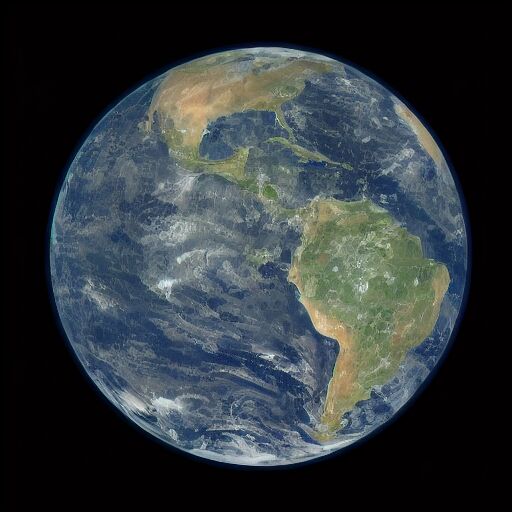}
\end{subfigure}\hfill
\begin{subfigure}[t]{\textwidth}
    \centering
    \includegraphics[width=0.16\textwidth]{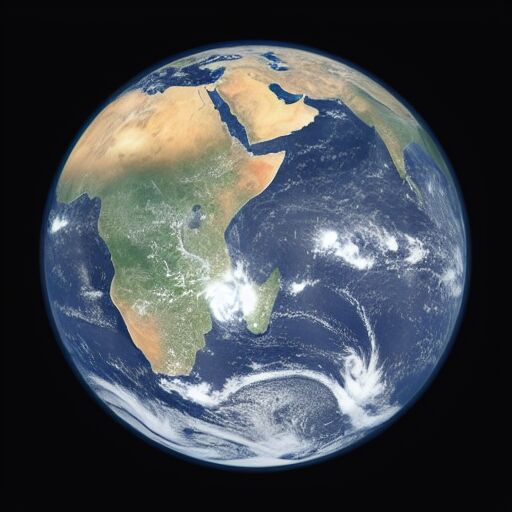}
    \includegraphics[width=0.16\textwidth]{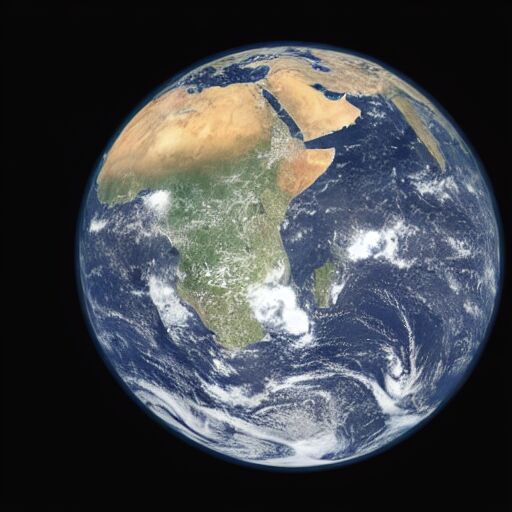}
    \includegraphics[width=0.16\textwidth]{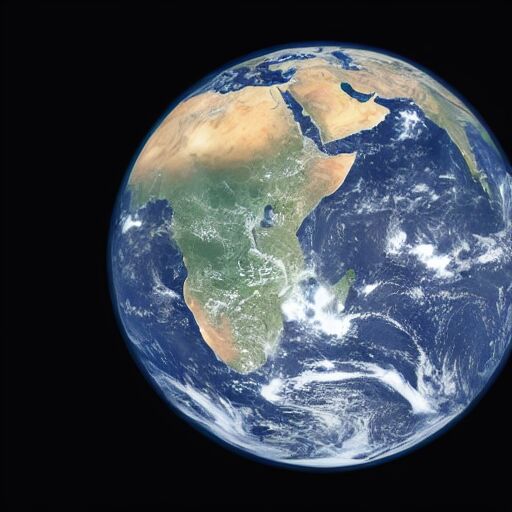}
    \includegraphics[width=0.16\textwidth]{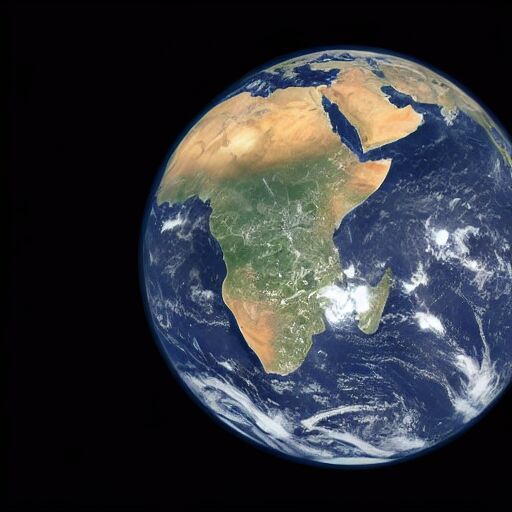}
    \includegraphics[width=0.16\textwidth]{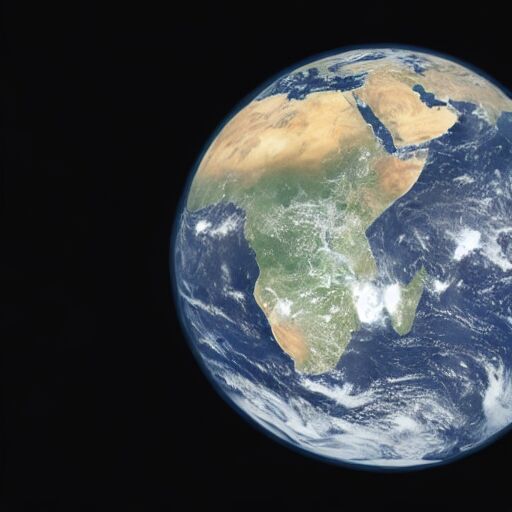}
    \includegraphics[width=0.16\textwidth]{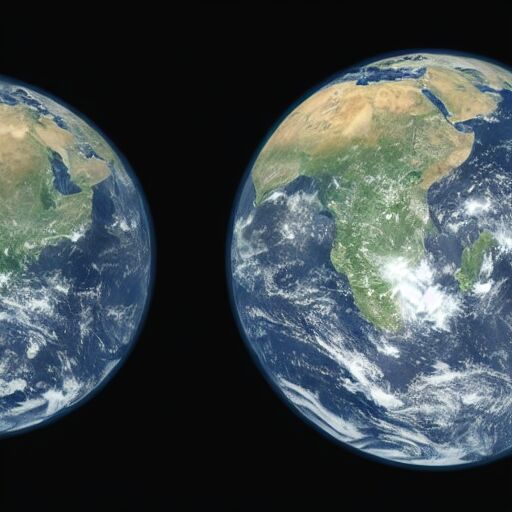}
\end{subfigure}
\caption{Rigid motion simulation: revolving Earth. Top: \ourmethod; Bottom: T2V0 \cite{khachatryan2023text2video}.}
\label{fig:earth}
\end{figure*}

\begin{figure*}[t]
\centering
\begin{subfigure}[t]{\textwidth}
    \centering
    \includegraphics[width=0.16\textwidth]{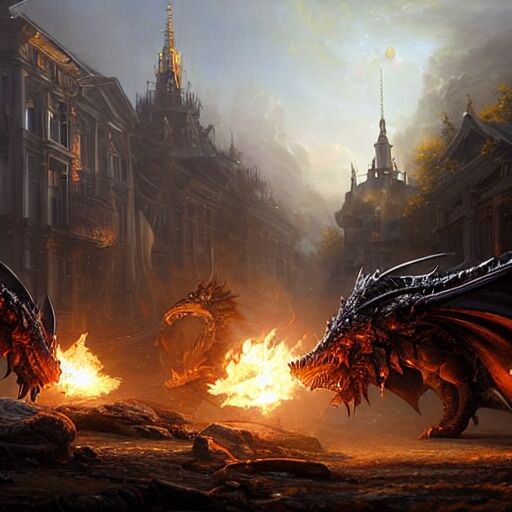}
    \includegraphics[width=0.16\textwidth]{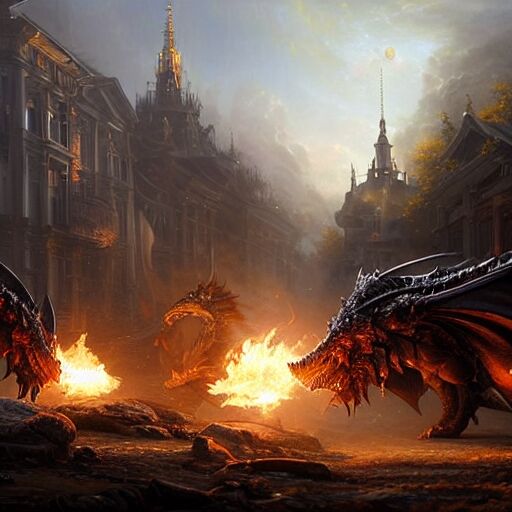}
    \includegraphics[width=0.16\textwidth]{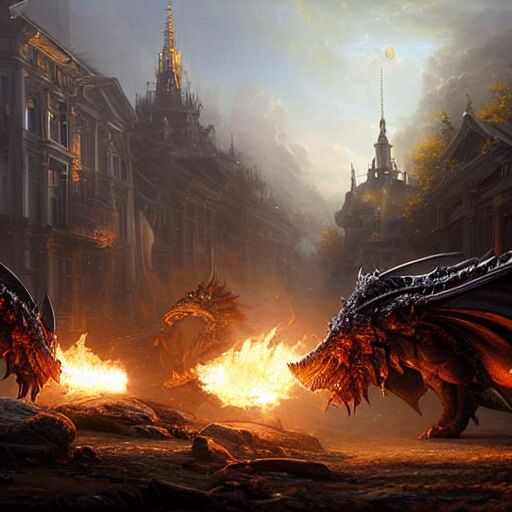}
    \includegraphics[width=0.16\textwidth]{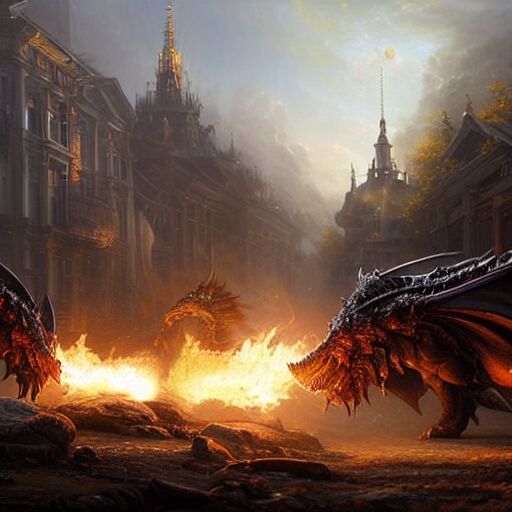}
    \includegraphics[width=0.16\textwidth]{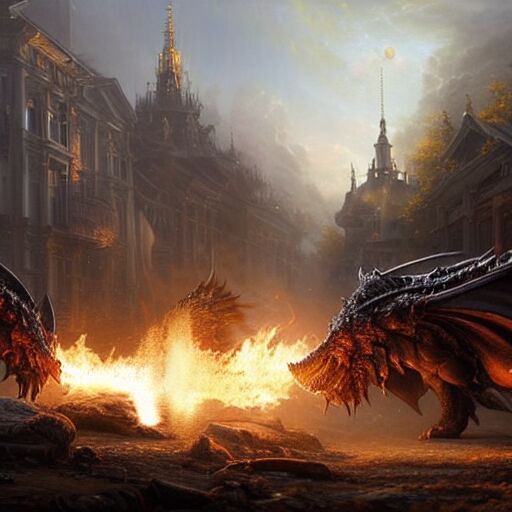}
    \includegraphics[width=0.16\textwidth]{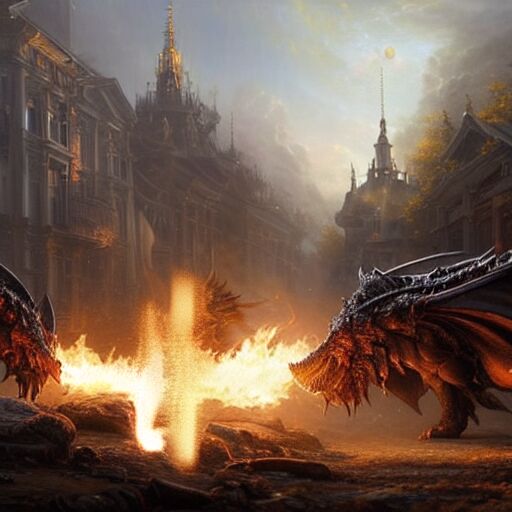}
\end{subfigure}\hfill
\begin{subfigure}[t]{\textwidth}
    \centering
    \includegraphics[width=0.16\textwidth]{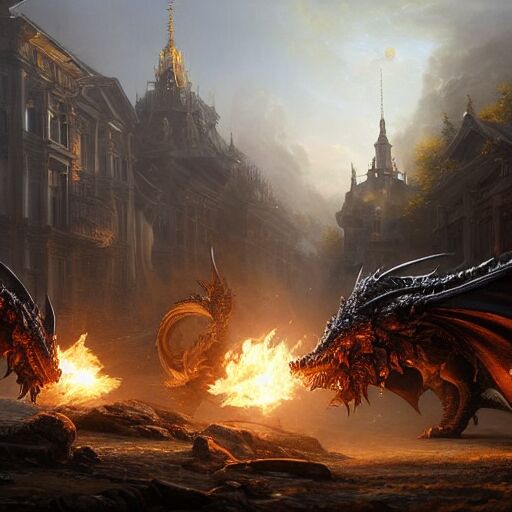}
    \includegraphics[width=0.16\textwidth]{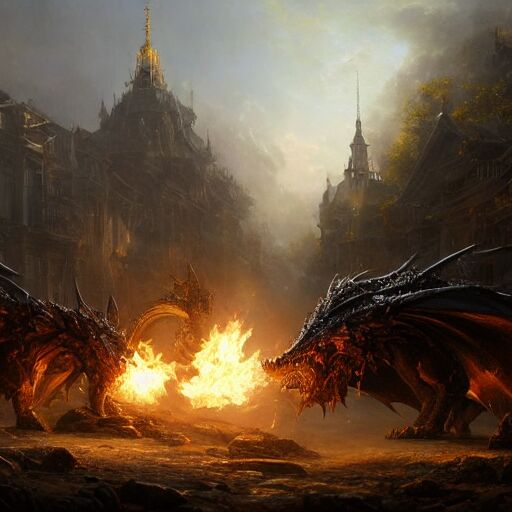}
    \includegraphics[width=0.16\textwidth]{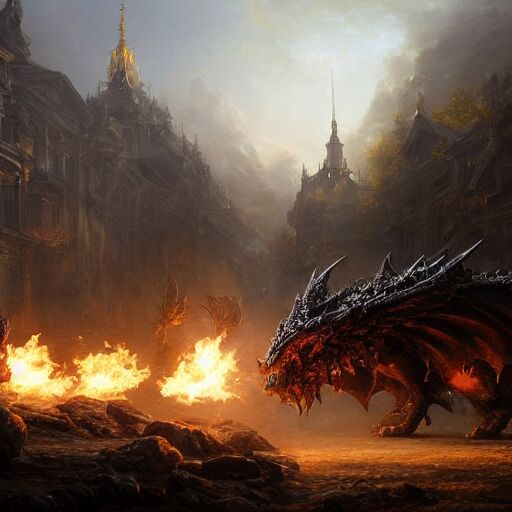}
    \includegraphics[width=0.16\textwidth]{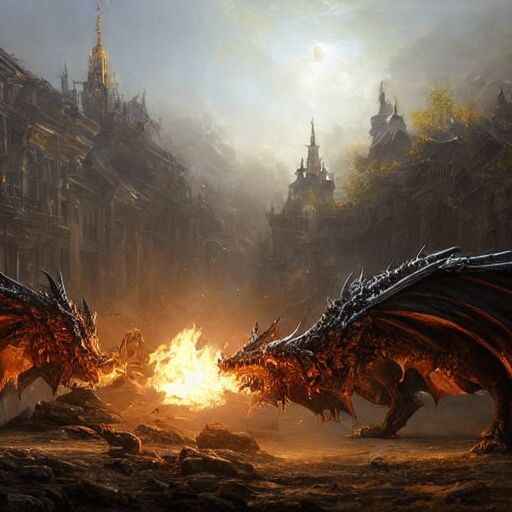}
    \includegraphics[width=0.16\textwidth]{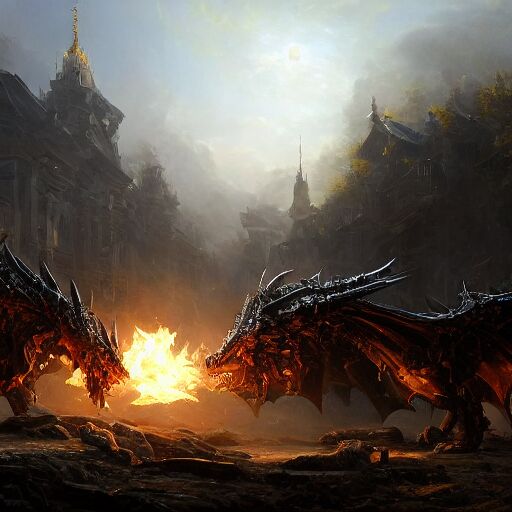}
    \includegraphics[width=0.16\textwidth]{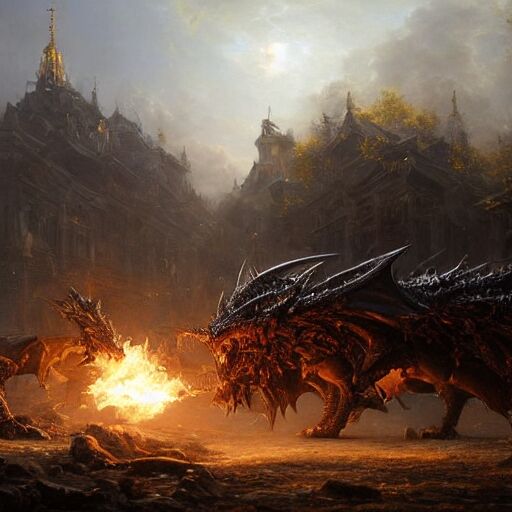}
\end{subfigure}
\caption{Fluid simulation: dragon fire. Top: \ourmethod; Bottom: T2V0 \cite{khachatryan2023text2video}.}
\label{fig:dragons}
\end{figure*}

\begin{figure*}[t]
\centering
\begin{subfigure}[t]{\textwidth}
    \centering
    \includegraphics[width=0.16\textwidth]{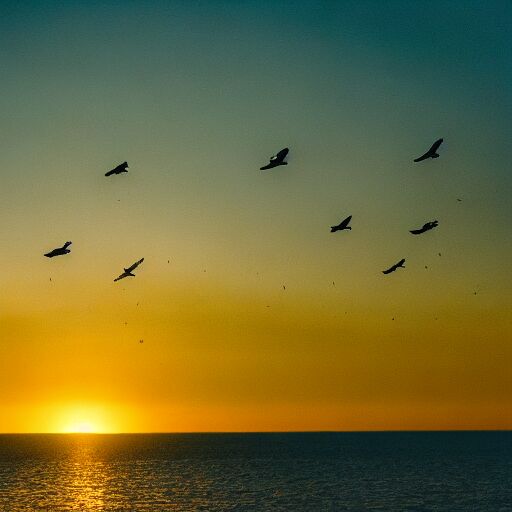}
    \includegraphics[width=0.16\textwidth]{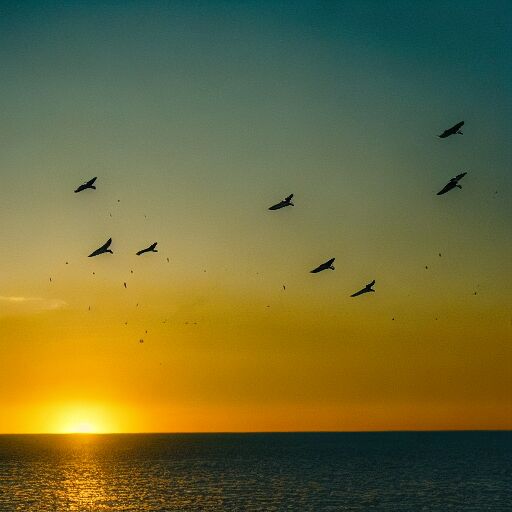}
    \includegraphics[width=0.16\textwidth]{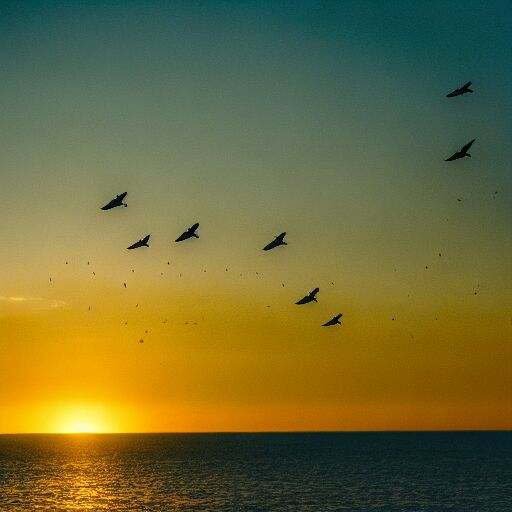}
    \includegraphics[width=0.16\textwidth]{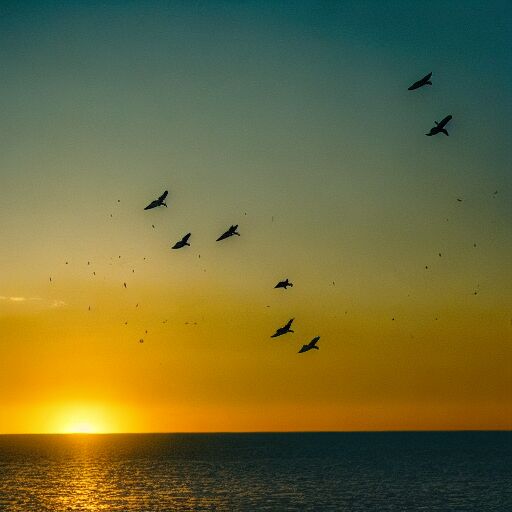}
    \includegraphics[width=0.16\textwidth]{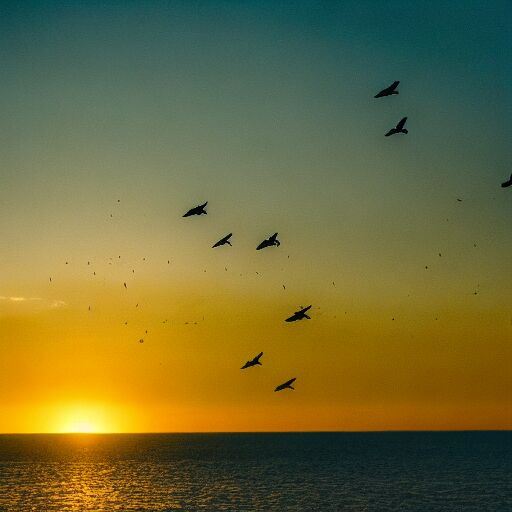}
    \includegraphics[width=0.16\textwidth]{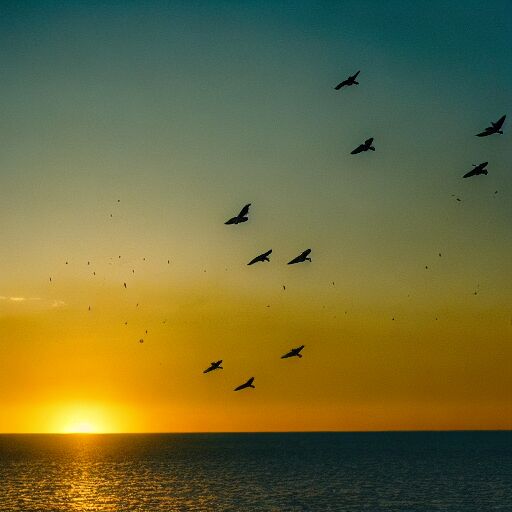}
\end{subfigure}\hfill
\begin{subfigure}[t]{\textwidth}
    \centering
    \includegraphics[width=0.16\textwidth]{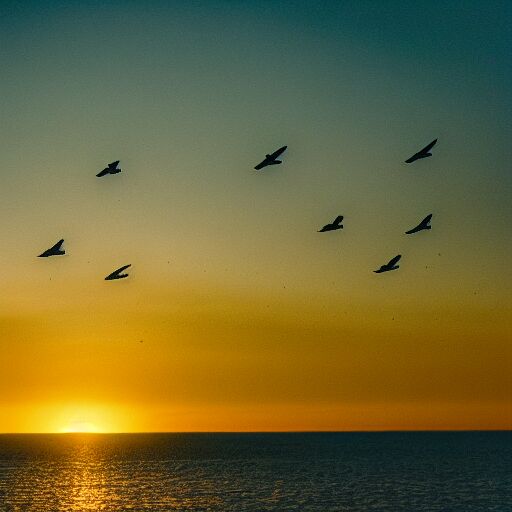}
    \includegraphics[width=0.16\textwidth]{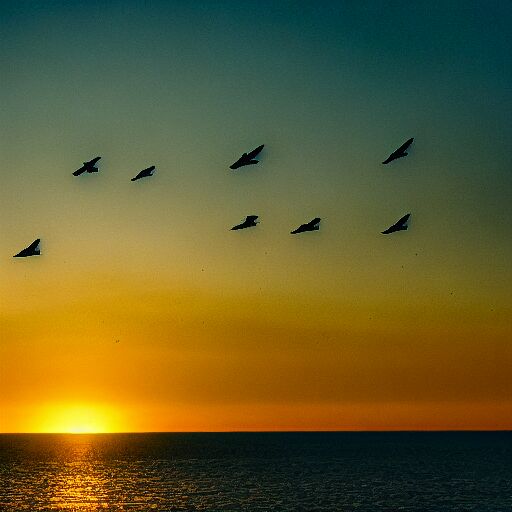}
    \includegraphics[width=0.16\textwidth]{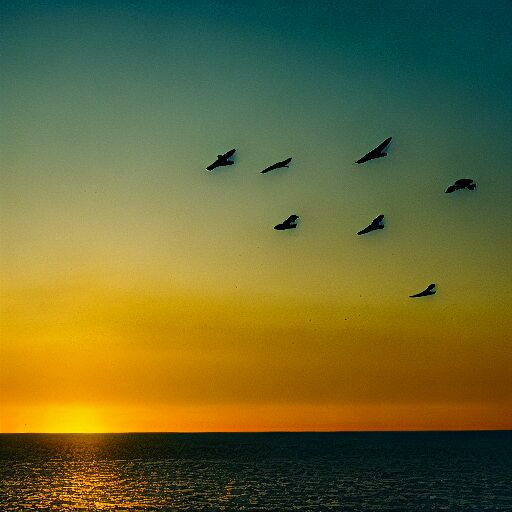}
    \includegraphics[width=0.16\textwidth]{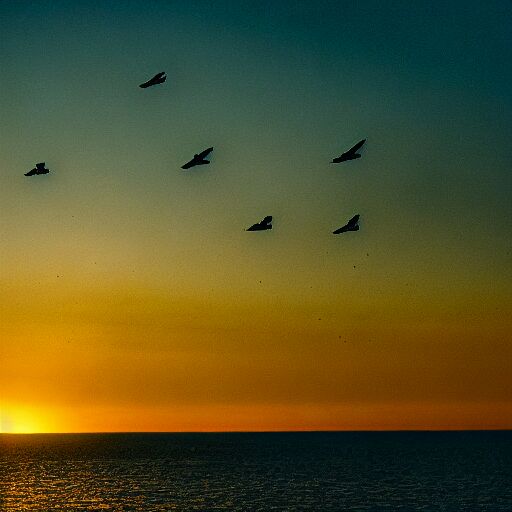}
    \includegraphics[width=0.16\textwidth]{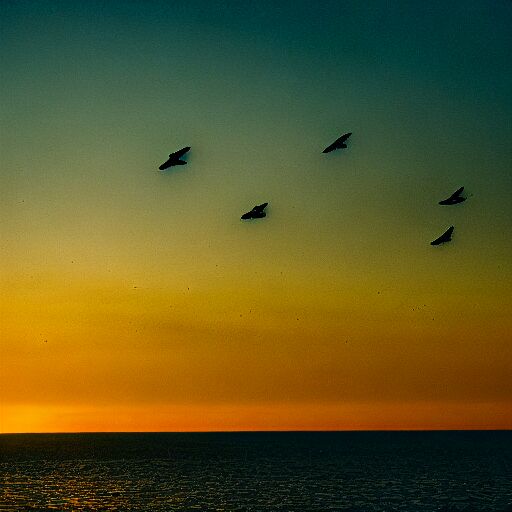}
    \includegraphics[width=0.16\textwidth]{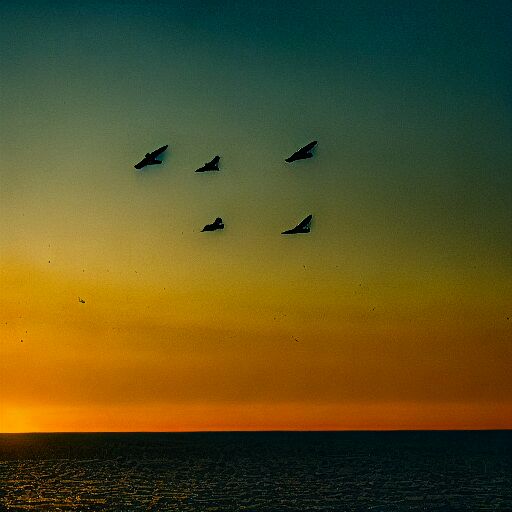}
\end{subfigure}
\caption{Multi-agent system simulation: bird flock. Top: \ourmethod; Bottom: T2V0 \cite{khachatryan2023text2video}.}
\label{fig:birds}
\end{figure*}

We show qualitative results in Figs. \ref{fig:meltingman}, \ref{fig:satellite}, \ref{fig:earth}, \ref{fig:dragons}, \ref{fig:birds}, which we separately describe in the following sections. Table \ref{tab:metrics} reports two metrics to evaluate the quality of the generated videos. As done in previous works, we use the \textit{Frame Consistency} metric, defined as the average cosine similarity of the CLIP embeddings of consecutive frames. However, this metric presents some limitations, as CLIP focuses on high-level semantic features and not on low-level details, resulting in high correlations even if the content changes but its semantics do not (as an example, see the video generated by T2V0 of the dragon in Fig. \ref{fig:dragons}, which has a \textit{Frame Consistency} of 0.97 even if the dragons are not the same dragons in each frame).
To overcome some of these limitations, we propose a novel metric, named \textit{Motion Consistency}, that measures how similar two frames are while accounting for the motion between them.
We start from the observation that, if an object moves through the scene, its textures should remain almost the same, and, if we know its flow, we can bring back that object to overlap with its starting position. Then we can apply a similarity distance between the initial image and the next frame brought back by the reversed flow.
Given two consecutive frames, we use a high-quality flow estimator (RAFT \cite{teed2020raft}) to estimate the optical flow between them and apply it on the second frame to reverse the motion. Then we compute the SSIM metric \cite{wang2004image} on the first frame and the registered one.

\begin{table}[t]
\centering
\setlength{\tabcolsep}{2pt}
{\renewcommand{\arraystretch}{1.2}
\caption{Quantitative results.}
\label{tab:metrics}

\begin{tabular}{rrcccc}
\hline
\multicolumn{1}{l}{}        & \multicolumn{1}{l}{} & \multicolumn{2}{c}{\textbf{Frame Consistency}} & \multicolumn{2}{c}{\textbf{Motion Consistency}} \\ \hline
\multicolumn{1}{l}{} & \multicolumn{1}{l}{} & T2V0 \cite{khachatryan2023text2video}            & \textbf{\ourmethod}            & T2V0 \cite{khachatryan2023text2video}   & \textbf{\ourmethod}            \\ \hline \hline
Fluid                & Dragons              & 0.9664          & \textbf{0.9991} & 0.6846 & \textbf{0.9637} \\ \hline
                     & Melting Man          & 0.9463          & \textbf{0.9566} & 0.7817 & \textbf{0.8252} \\ \hline
\multirow{2}{*}{Rigid Body} & Satellite Scan       & 0.9588            & \textbf{0.9875}            & 0.2852             & \textbf{0.9219}            \\ \cline{2-6} 
                     & Revolving Earth      & \textbf{0.9812} & 0.9696          & \textbf{0.7213} & 0.6783 \\ \hline
Agents               & Birds                & 0.9765          & \textbf{0.9968} & 0.8973 & \textbf{0.9385} \\ \hline \hline
\multicolumn{2}{c}{Average} & 0.9658 $\pm$ 0.01  & \textbf{0.9819} $\pm$ 0.02 & 0.6740 $\pm$ 0.23 & \textbf{0.8655} $\pm$ 0.12 \\ \hline
\end{tabular}
}
\end{table}

\subsection{Rigid Body flows}

Fig. \ref{fig:satellite} shows a pivotal example where \ourmethod can be directly compared to the state-of-the-art T2V0, as in this case we use an optical flow equivalent to a their proposed shift along the vertical axis. This example shows a video generated starting from a satellite view of a city, and, by simulating the rectilinear motion of the satellite, new portions of the city appear from the top of the image. While T2V0 struggles with keeping temporal consistency, even with large structural elements (e.g., the river), \ourmethod is able to coherently scroll down the already present part of the city, while also generating new plausible content in the top part of the frames.

A similar case study is the Earth rotation in Fig. \ref{fig:earth}. Here, the optical flow is obtained by simulating a rotating sphere that was fitted to the first frame while keeping track of the starting and ending position of each point. As the Earth rotates, a slice disappears from one side and a new one needs to be generated on the opposite side. Thanks to the powerful natural image prior of SD, \ourmethod is able to autonomously generate other continents in the correct position, even if the text prompt contains no reference about them (see Appendix \ref{sec:prompt} for all the text prompts used in this paper). On the contrary, T2V0 is not able to rotate the Earth consistently while creating new content, as visible in the same Fig. \ref{fig:earth}.

\subsection{Fluids dynamics}

In this set of experiments, we use the $\Phi$-flow \cite{holl2020learning} library to simulate fluid dynamics (by numerically solving Navier-Stokes equations) with the shape and position provided by the first frame $I^0$.
Moreover, we can set up the simulation in different ways, depending on the numerical solvers, i.e. \textit{Eulerian} (particle-based) or \textit{Lagrangian} (grid-based), we can add rigid obstacles to the fluid or we can define a initial velocity and force fields. All these different options result in videos that can have the same starting frame but differ in their evolution according to the simulation constraints. We extract the velocity field of the simulation as a proxy for the optical flow. Examples of the velocity field can be seen in Appendix \ref{sec:extraphysics}.

Fig. \ref{fig:dragons} shows a fluid simulation of two dragons breathing fire. We can approximate the two initial fire breaths with two centered smoke balls, obtaining a binary mask that will be fed to the simulation. At this point, we run the simulation, solving the Navier-Stokes equations by sequentially evaluating advection, diffusion and pressure. The vorticity and the expansion of the smoke is due to the buoyancy force set in the desired direction, that in this case is such that the two balls cross near the middle of the image.

The figure shows that \ourmethod produces a consistent scene with a realistic animation of the fire breaths. Moreover, the global scene illumination seems to change accordingly, and a realistic occlusion of a dragon due to smoke gradually appears. This is mainly due to the MCFA mechanism, as we ablate in Sec. \ref{sec:ablations}. In T2V0, the scene is not temporally consistent and shows increasingly more artefacts, such as color shifts or the fact that the right dragon changes with time, while the left one even disappears.

A similar analysis can be conducted for Fig. \ref{fig:meltingman}, where a simulation of a melting statue is shown. We can see that the generated video includes bouncing of parts on the ground before the fluid settles. 

\subsection{Multi-agent systems}
\label{subsec:multiag}
Multi-agent systems are another interesting family of simulated dynamics. A simple agents model is the \textit{Boids} model \citep{reynolds1987flocks}, consisting of a set of point-like agents (named boids) that move according to three steering behaviour rules: separation, as boids avoid collisions with nearby agents by steering away from them, alignment, as boids align their direction with that of nearby agents, and cohesion, as boids move towards the average position of nearby agents to stay together as a group. To simulate this system we used the agentpy \citep{foramitti2021agentpy} library, in which the number of agents, the simulation time-steps and different physical parameters related to the steering rules can be chosen.

An example is shown in Fig. \ref{fig:birds}, generating the temporal evolution of a flock of birds.
As SD is not able to generate images with a controllable number of agents in specified positions, we start from an image where there is a single agent (a bird in the example). Then, we extract the corresponding latent vector patch with the attention map \citep{epstein2023diffusion} related to the CLIP token containing the word "bird", and clone it to the simulated positions of the other agents. At this point, we evolve the frames according to the optical flow derived from the simulation velocity field.

While \ourmethod produces a realistic flock motion, T2V0 motion is not consistent and the number of birds changes in each frame.

\begin{figure*}[t]
\centering
\begin{subfigure}[t]{\textwidth}
    \centering
    \includegraphics[width=0.16\textwidth]{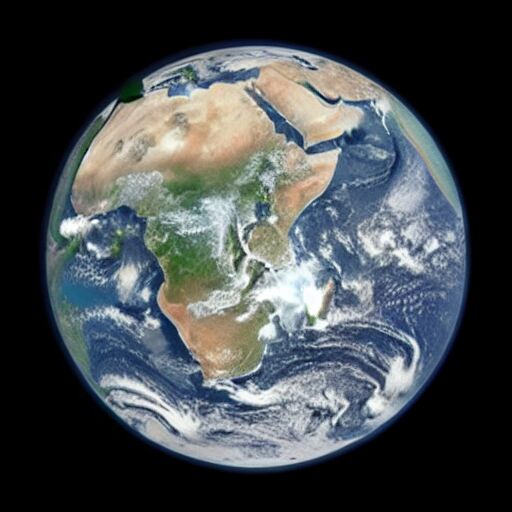}
    \includegraphics[width=0.16\textwidth]{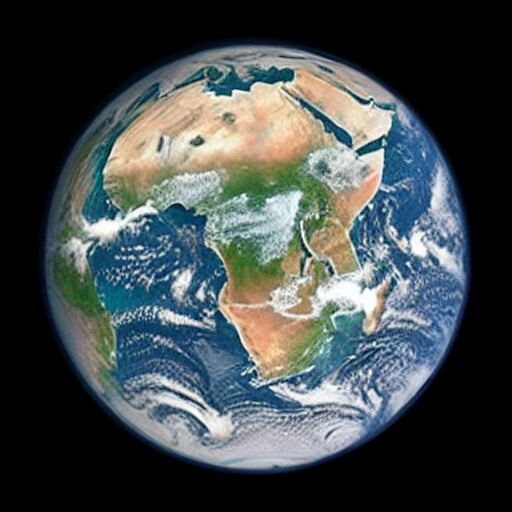}
    \includegraphics[width=0.16\textwidth]{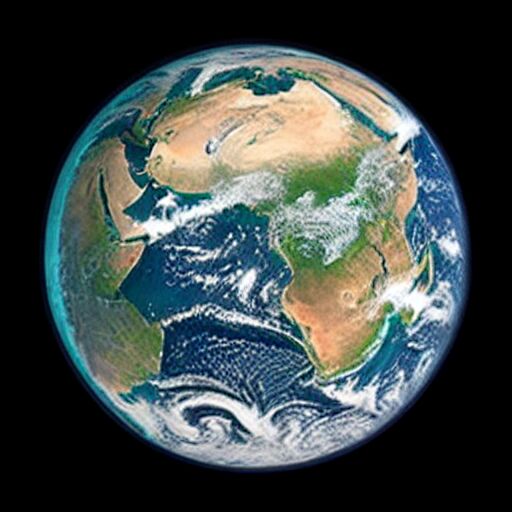}
    \includegraphics[width=0.16\textwidth]{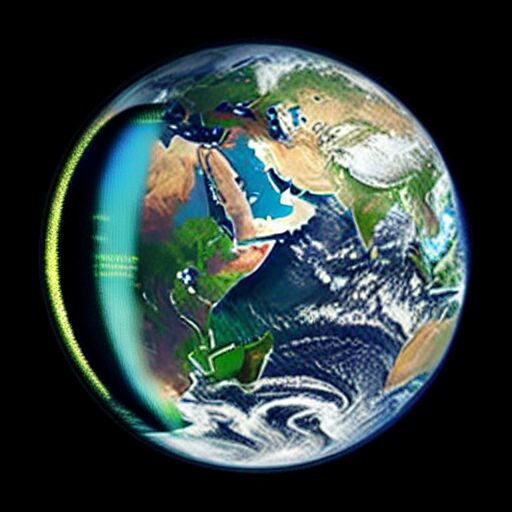}
    \includegraphics[width=0.16\textwidth]{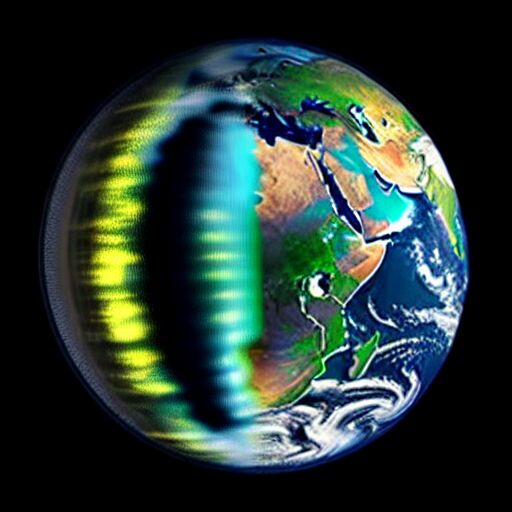}
    \includegraphics[width=0.16\textwidth]{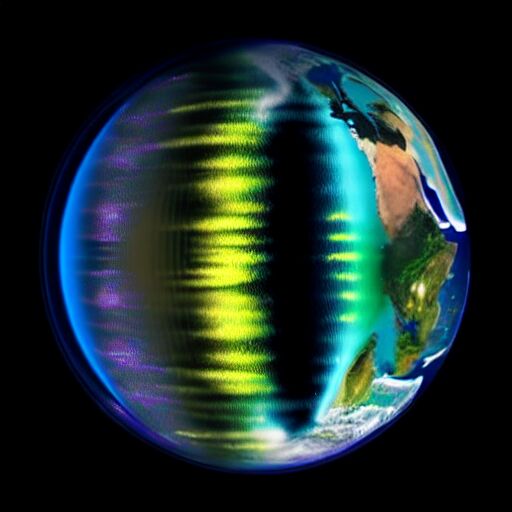}
\end{subfigure}\hfill
\begin{subfigure}[t]{\textwidth}
    \centering
    \includegraphics[width=0.16\textwidth]{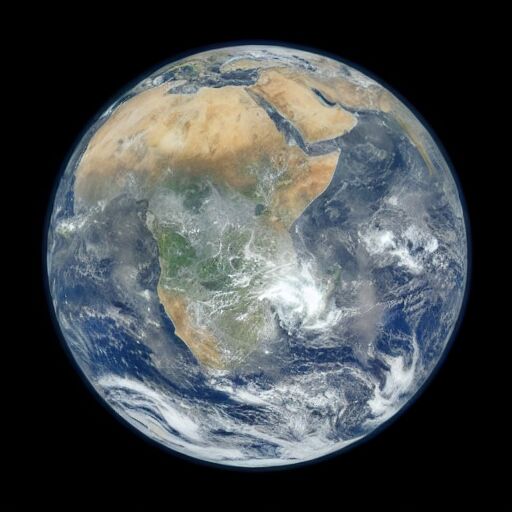}
    \includegraphics[width=0.16\textwidth]{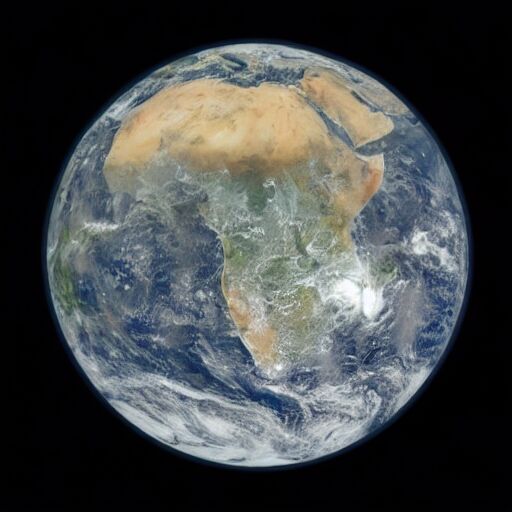}
    \includegraphics[width=0.16\textwidth]{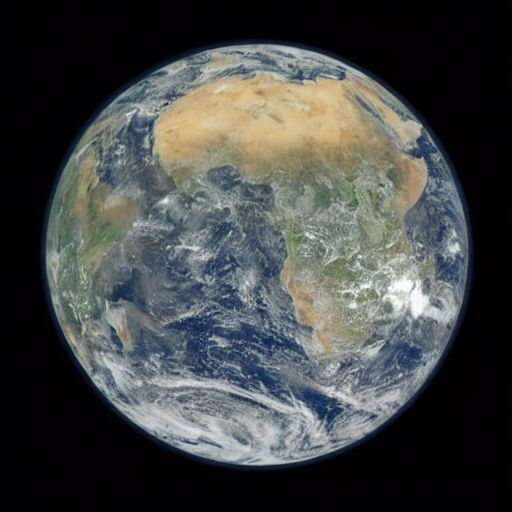}
    \includegraphics[width=0.16\textwidth]{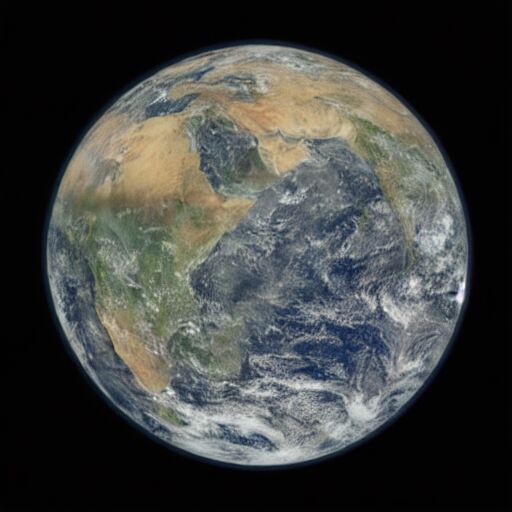}
    \includegraphics[width=0.16\textwidth]{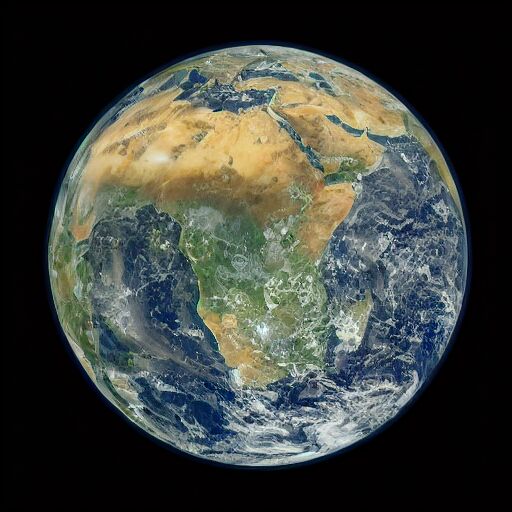}
    \includegraphics[width=0.16\textwidth]{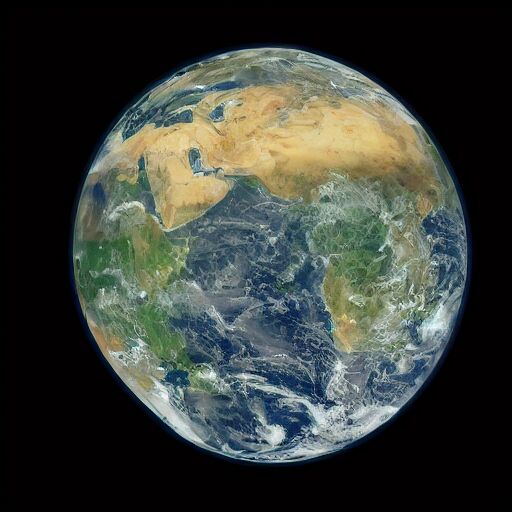}
\end{subfigure}\hfill
\begin{subfigure}[t]{\textwidth}
    \centering
    \includegraphics[width=0.16\textwidth]{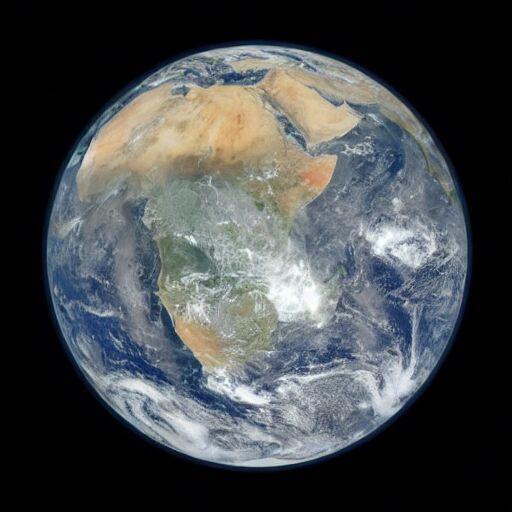}
    \includegraphics[width=0.16\textwidth]{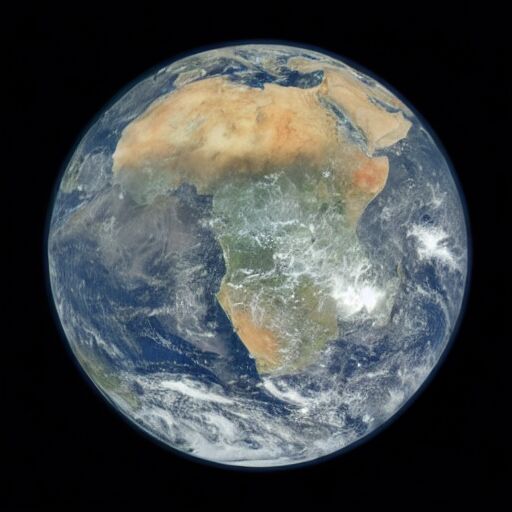}
    \includegraphics[width=0.16\textwidth]{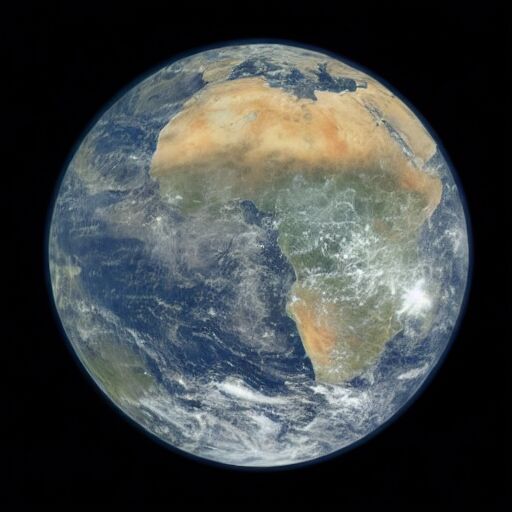}
    \includegraphics[width=0.16\textwidth]{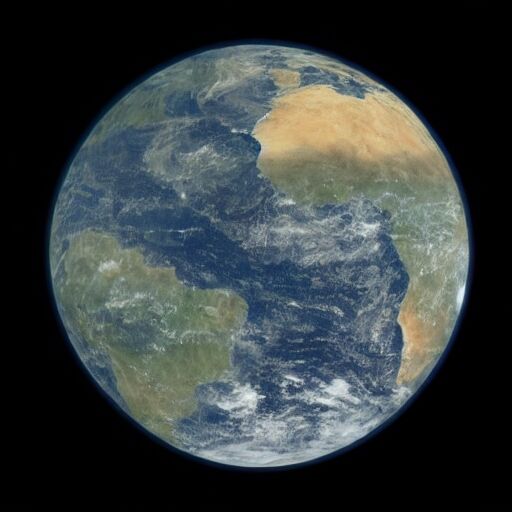}
    \includegraphics[width=0.16\textwidth]{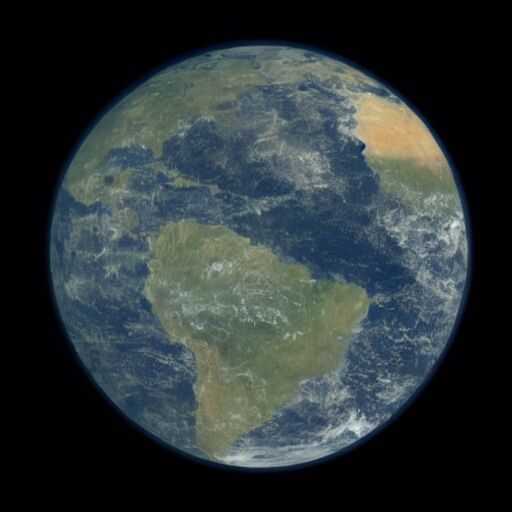}
    \includegraphics[width=0.16\textwidth]{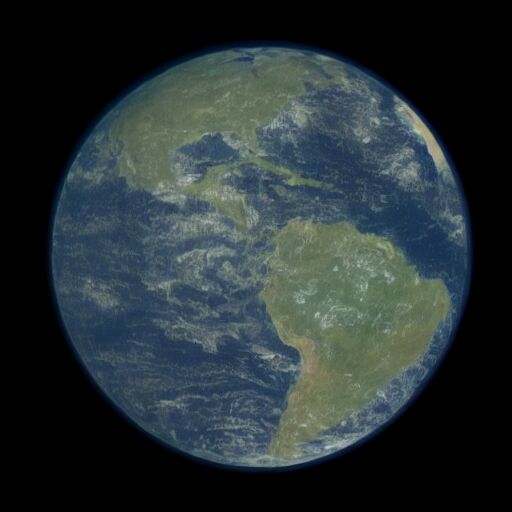}
\end{subfigure}\hfill
\begin{subfigure}[t]{\textwidth}
    \centering
    \includegraphics[width=0.16\textwidth]{figures/final/earth/frame_000_2}
    \includegraphics[width=0.16\textwidth]{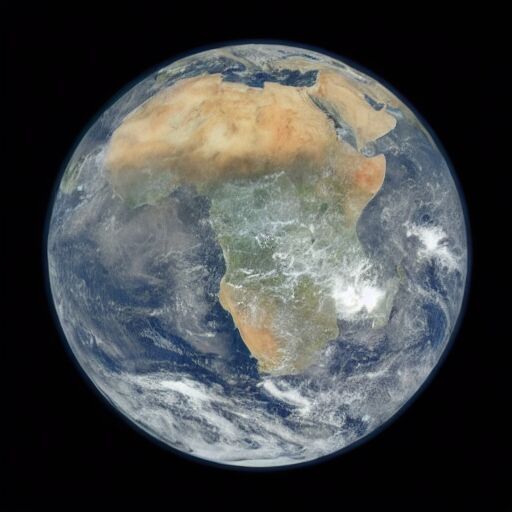}
    \includegraphics[width=0.16\textwidth]{figures/final/earth/frame_002_2}
    \includegraphics[width=0.16\textwidth]{figures/final/earth/frame_004_2}
    \includegraphics[width=0.16\textwidth]{figures/final/earth/frame_006_2}
    \includegraphics[width=0.16\textwidth]{figures/final/earth/frame_008_2}
    \end{subfigure}
    \caption{Ablation - Cross-Frame attention. First row: no cross frame attention; Second Row: Attend only to the initial frame; Third Row: Attend only to the previous frame; Fourth Row: Attend to the initial and preceding frame (ours).}
    \label{fig:earth_abl_cross}
\end{figure*}

\begin{figure*}[!ht]
    \centering
    \begin{subfigure}[t]{\textwidth}
        \centering
        \includegraphics[width=0.16\textwidth]{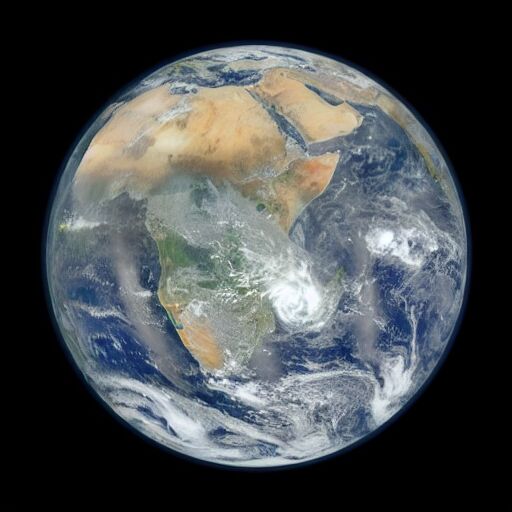}
        \includegraphics[width=0.16\textwidth]{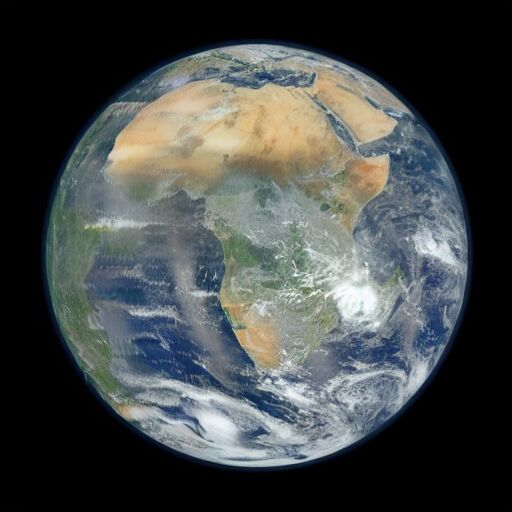}
        \includegraphics[width=0.16\textwidth]{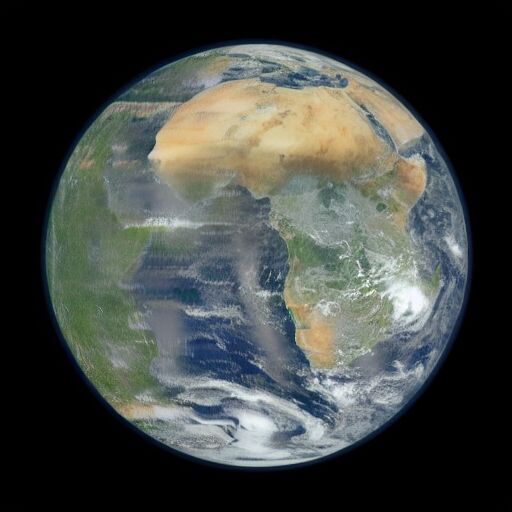}
        \includegraphics[width=0.16\textwidth]{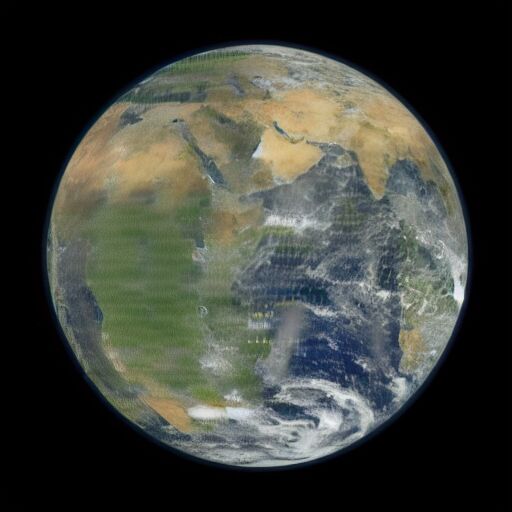}
        \includegraphics[width=0.16\textwidth]{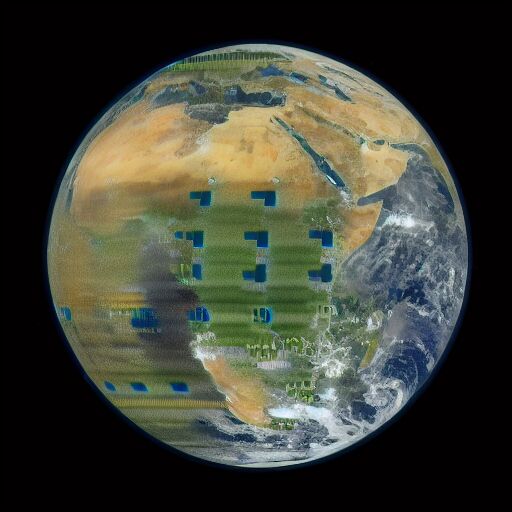}
        \includegraphics[width=0.16\textwidth]{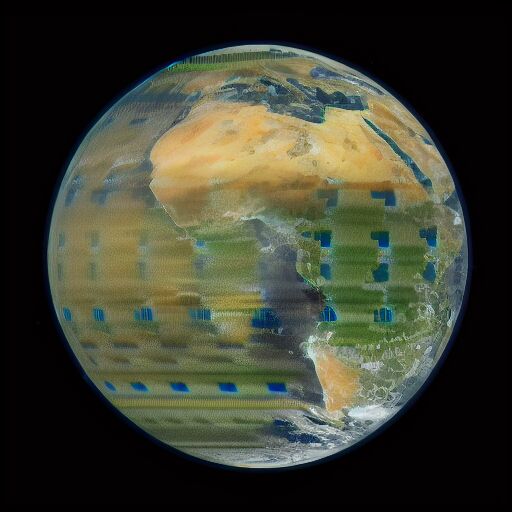}
    \end{subfigure}\hfill
    \begin{subfigure}[t]{\textwidth}
        \centering
        \includegraphics[width=0.16\textwidth]{figures/final/earth/frame_000_2}
        \includegraphics[width=0.16\textwidth]{figures/final/earth/frame_001_2}
        \includegraphics[width=0.16\textwidth]{figures/final/earth/frame_002_2}
        \includegraphics[width=0.16\textwidth]{figures/final/earth/frame_004_2}
        \includegraphics[width=0.16\textwidth]{figures/final/earth/frame_006_2}
        \includegraphics[width=0.16\textwidth]{figures/final/earth/frame_008_2}
    \end{subfigure}
    \caption{Ablation - Spatial-$\eta$. First Row: $\eta = 0$; Second Row: Spatial-$\eta$ on.}
    \label{fig:earth_abl_spatialeta}
\end{figure*}

\subsection{Ablations}\label{sec:ablations}
In this section, we ablate the contribution of the most important components/hyperparameters in the proposed pipeline. First, we start from investigating the impact of the cross-attention mechanism by comparing four different variants: i) each frame attends to itself (no MCFA); ii) each frame attends to the previous frame; iii) each frame attends to the first frame; iv) each frame attends to both the previous frame and the first frame (proposed MCFA).
Visual results are shown in Fig. \ref{fig:earth_abl_cross}. As can be seen, the MCFA mechanism is necessary to generate plausible frames; moreover, attending only to the first frame reduces the overall motion, (e.g., always showing Africa as in the first frame), while only attending to the previous frame reduces color consistency. Overall, we demonstrate that the proposed MCFA, attending to both the first and the previous frame, represents the optimal solution to keep global consistency with the initial image and local consistency with the preceding frame.

Fig. \ref{fig:earth_abl_spatialeta} shows the ablation of the Spatial-$\eta$ weighting technique. As shown, being able to sample with DDPM in some parts of the image is crucial in order to generate novel plausible content. Indeed, DDPM adds, during each reverse diffusion step, random white noise to the latent. We suppose that this allows to better sample from the real distribution, avoiding artefacts other components of the method, such as the warping operator or the MCFA, would otherwise introduce.

Finally, we ablated the partial inversion process, i.e., lines 2-6 in Alg. \ref{alg:H}. Without the DDIM inversion, textures and details generated by SD cannot be brought into the next frame, resulting in corrupted videos. Visual results can be found in the Appendix \ref{sec:abl_inv}.

\newpage
\subsection{Additional Qualitative Results}
\begin{figure}[t]
\centering
\includegraphics[width=0.16\textwidth]{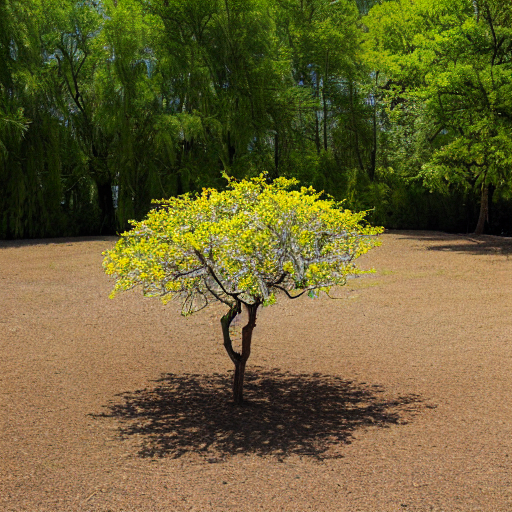}
\includegraphics[width=0.16\textwidth]{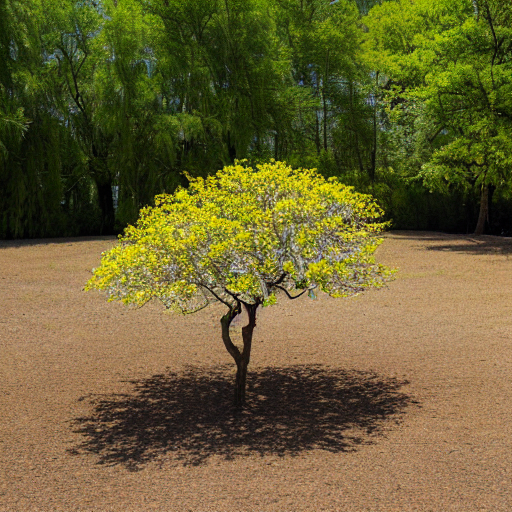}
\includegraphics[width=0.16\textwidth]{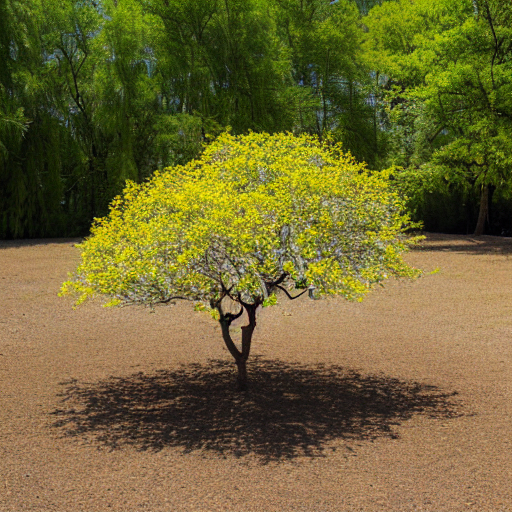}
\includegraphics[width=0.16\textwidth]{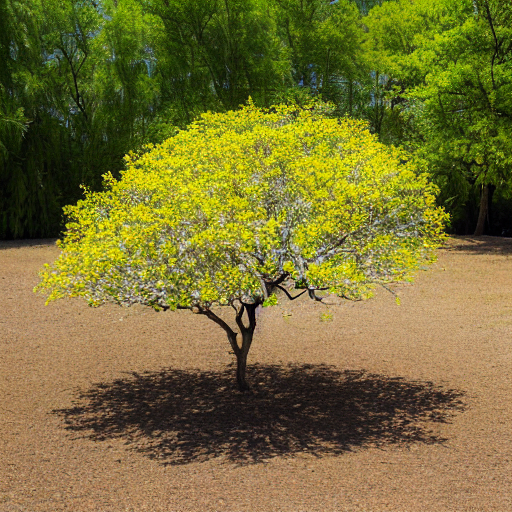}
\includegraphics[width=0.16\textwidth]{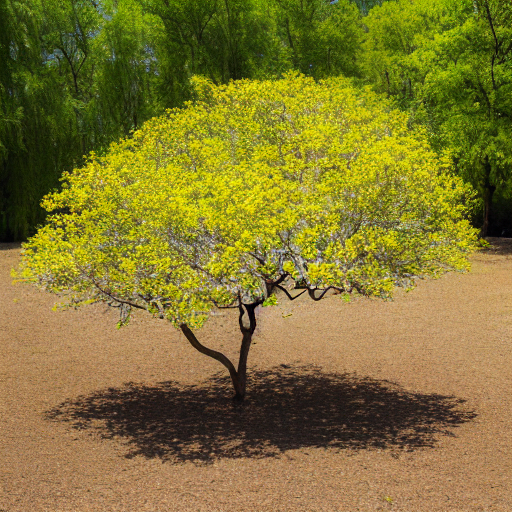}
\includegraphics[width=0.16\textwidth]{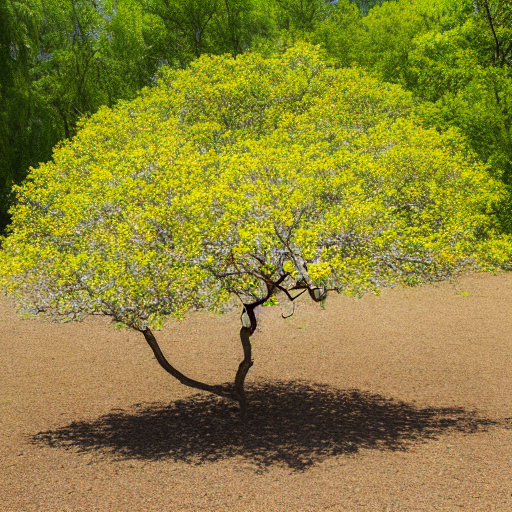}
\includegraphics[width=0.16\textwidth]{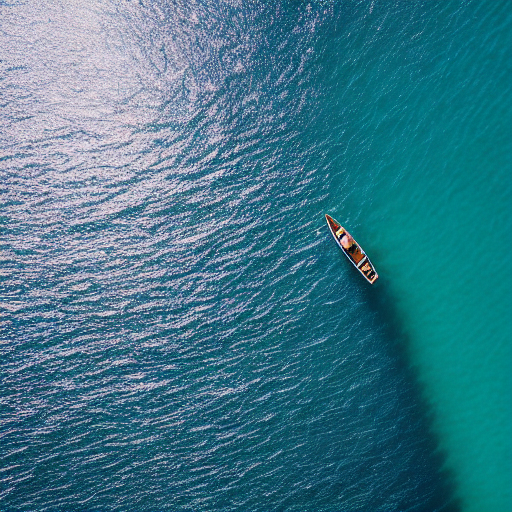}
\includegraphics[width=0.16\textwidth]{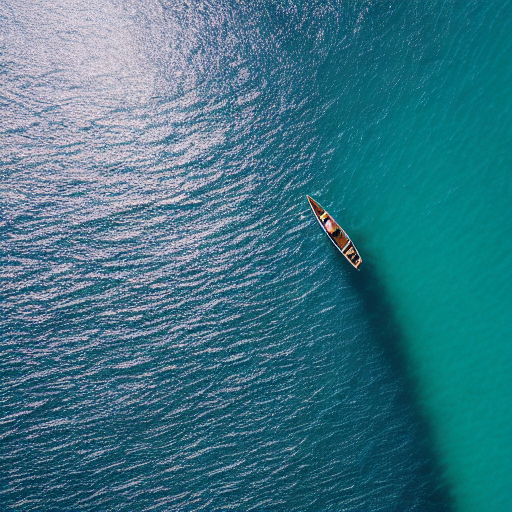}
\includegraphics[width=0.16\textwidth]{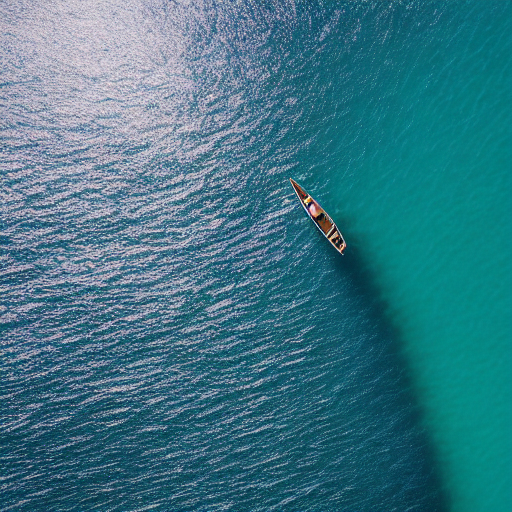}
\includegraphics[width=0.16\textwidth]{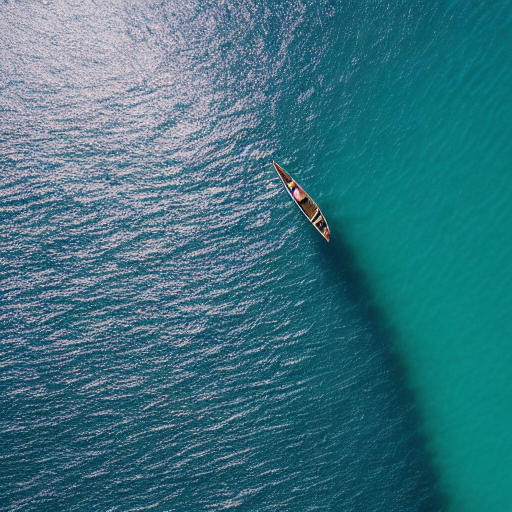}
\includegraphics[width=0.16\textwidth]{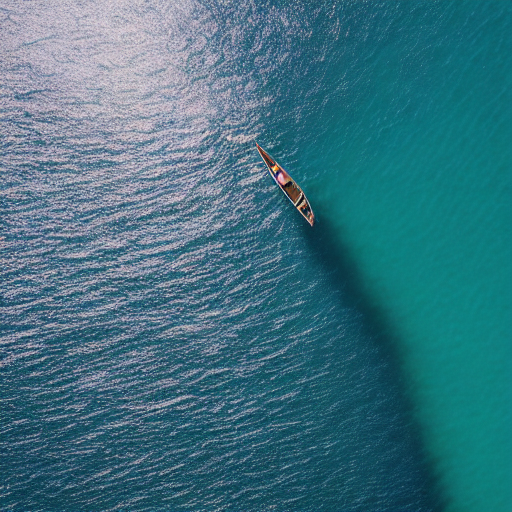}
\includegraphics[width=0.16\textwidth]{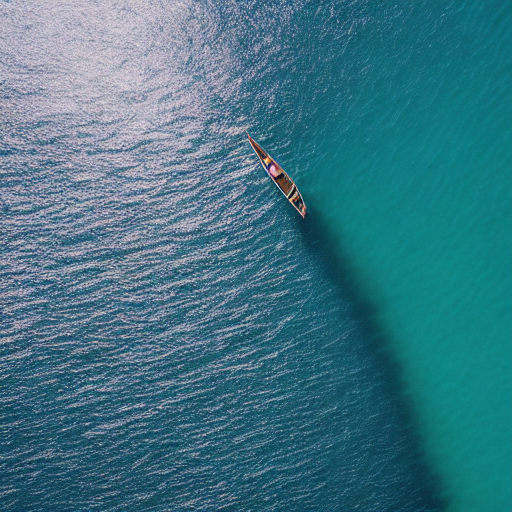}
\includegraphics[width=0.16\textwidth]{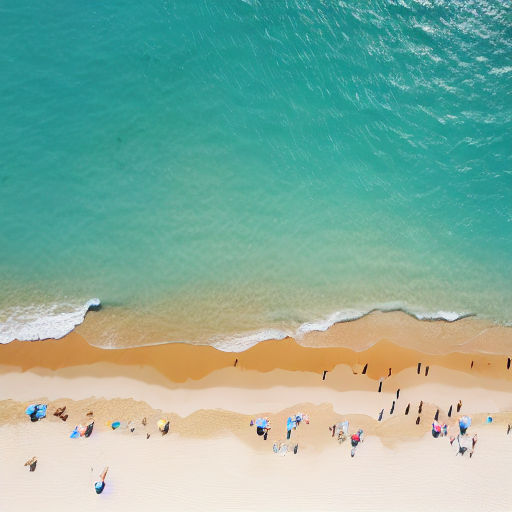}
\includegraphics[width=0.16\textwidth]{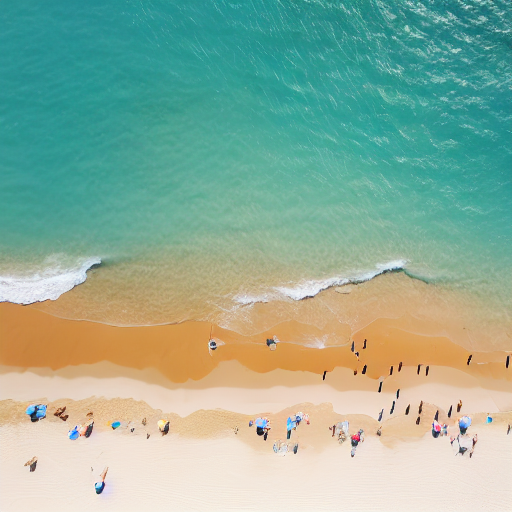}
\includegraphics[width=0.16\textwidth]{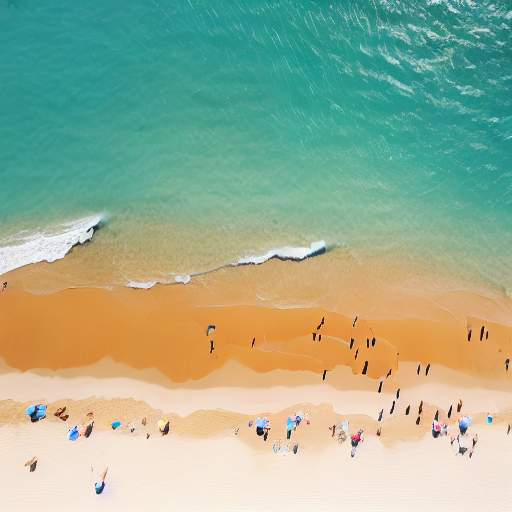}
\includegraphics[width=0.16\textwidth]{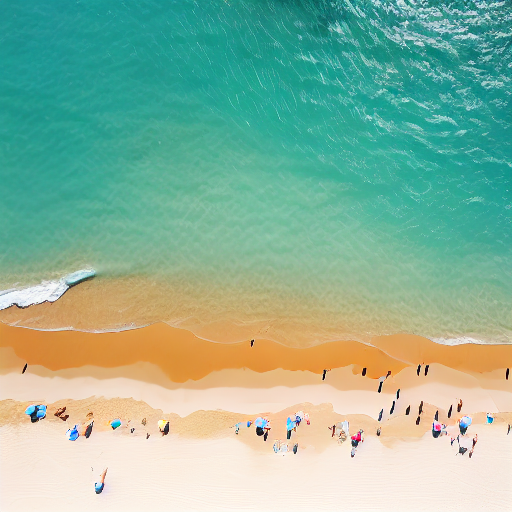}
\includegraphics[width=0.16\textwidth]{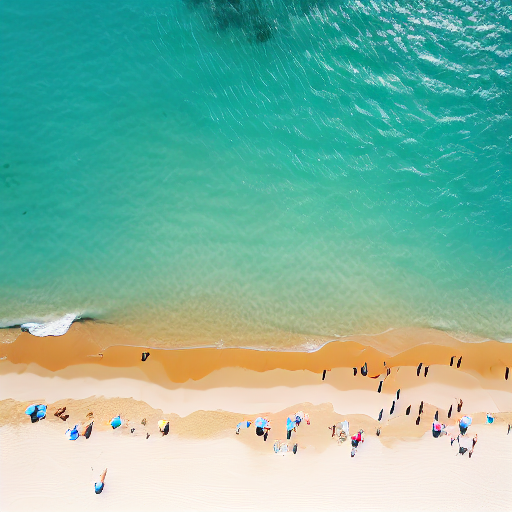}
\includegraphics[width=0.16\textwidth]{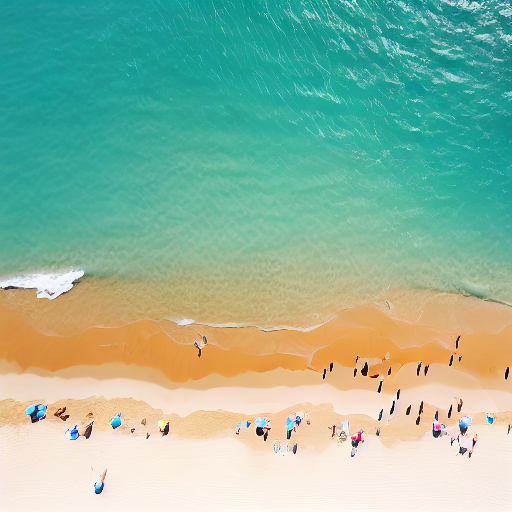}
\includegraphics[width=0.16\textwidth]{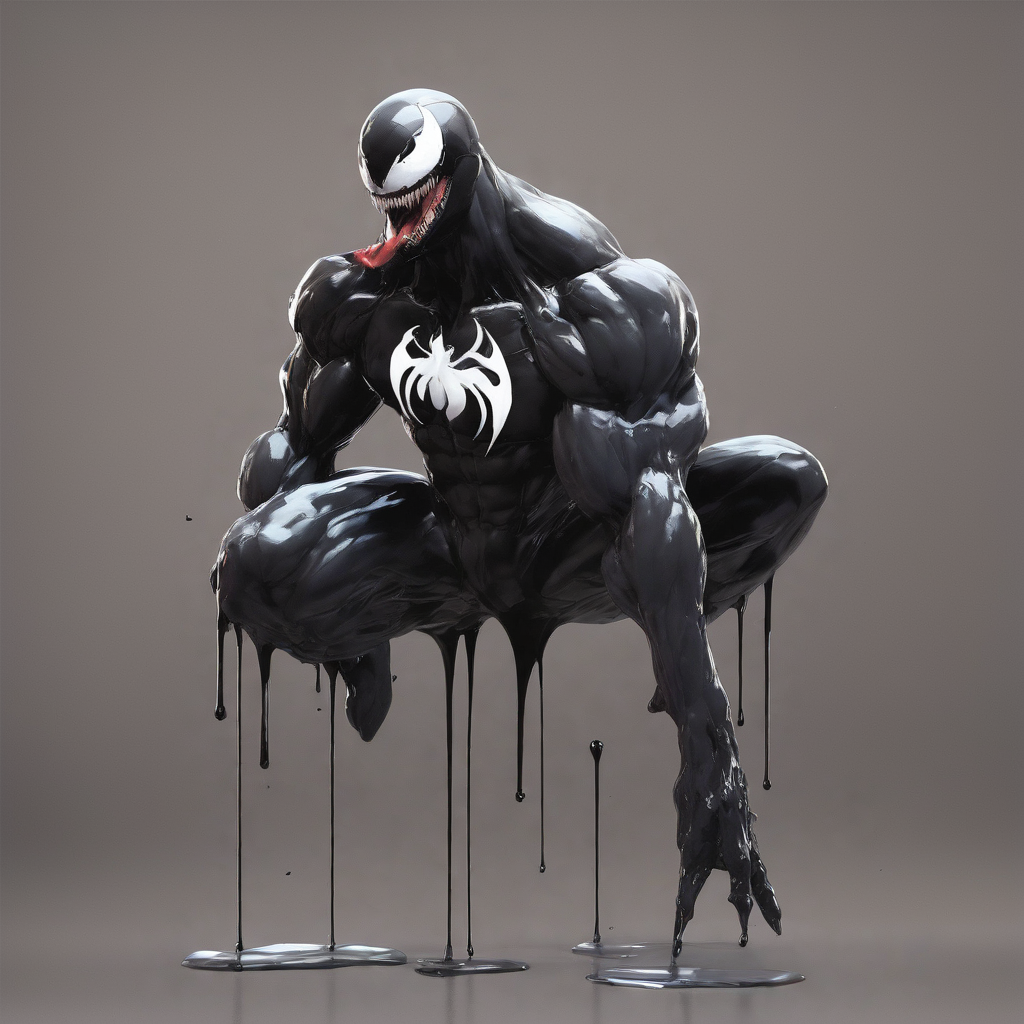}
\includegraphics[width=0.16\textwidth]{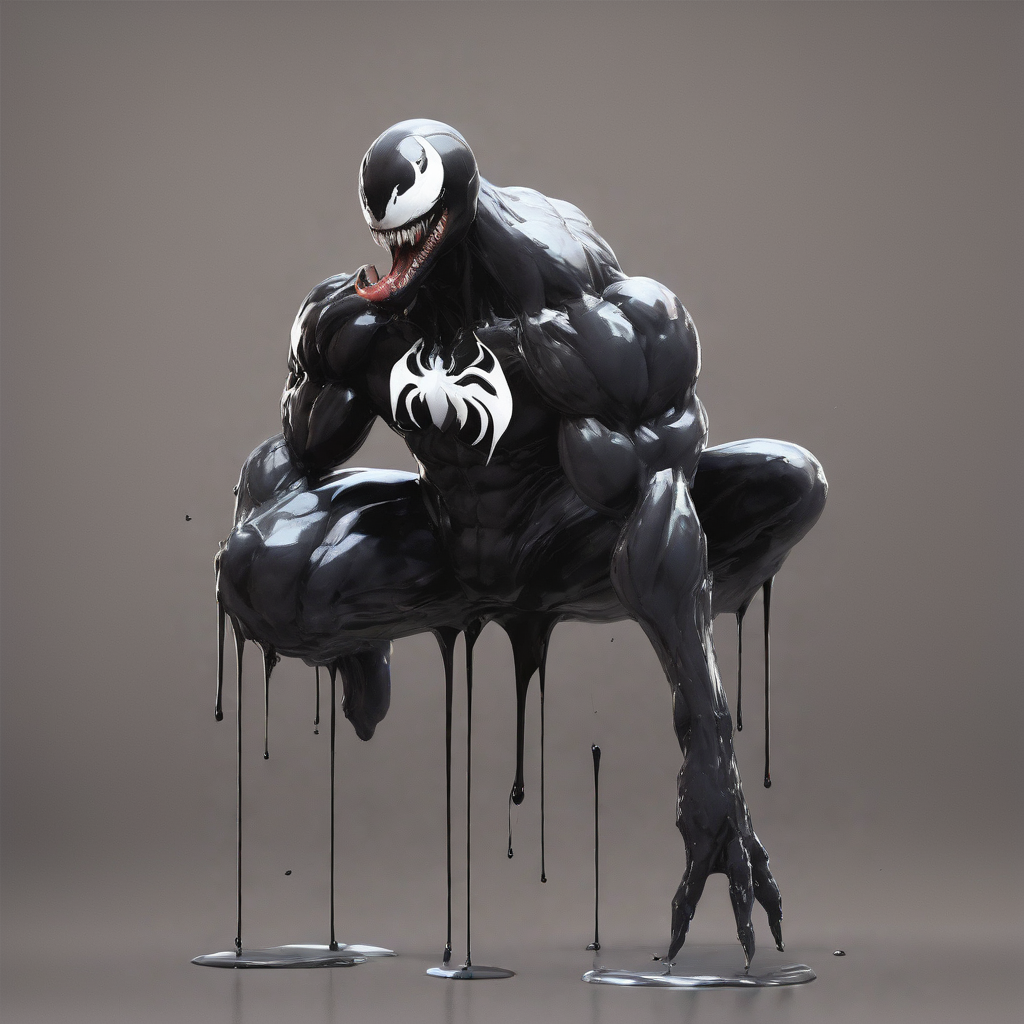}
\includegraphics[width=0.16\textwidth]{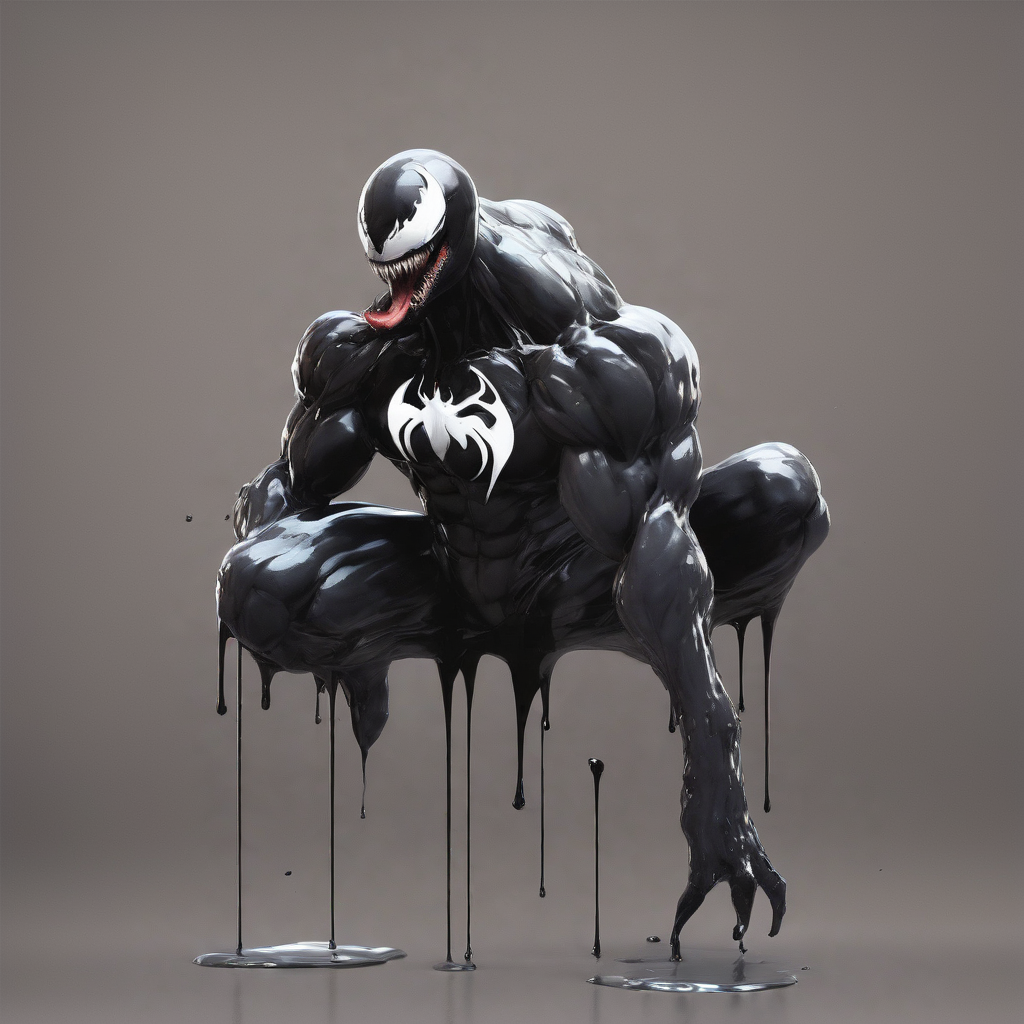}
\includegraphics[width=0.16\textwidth]{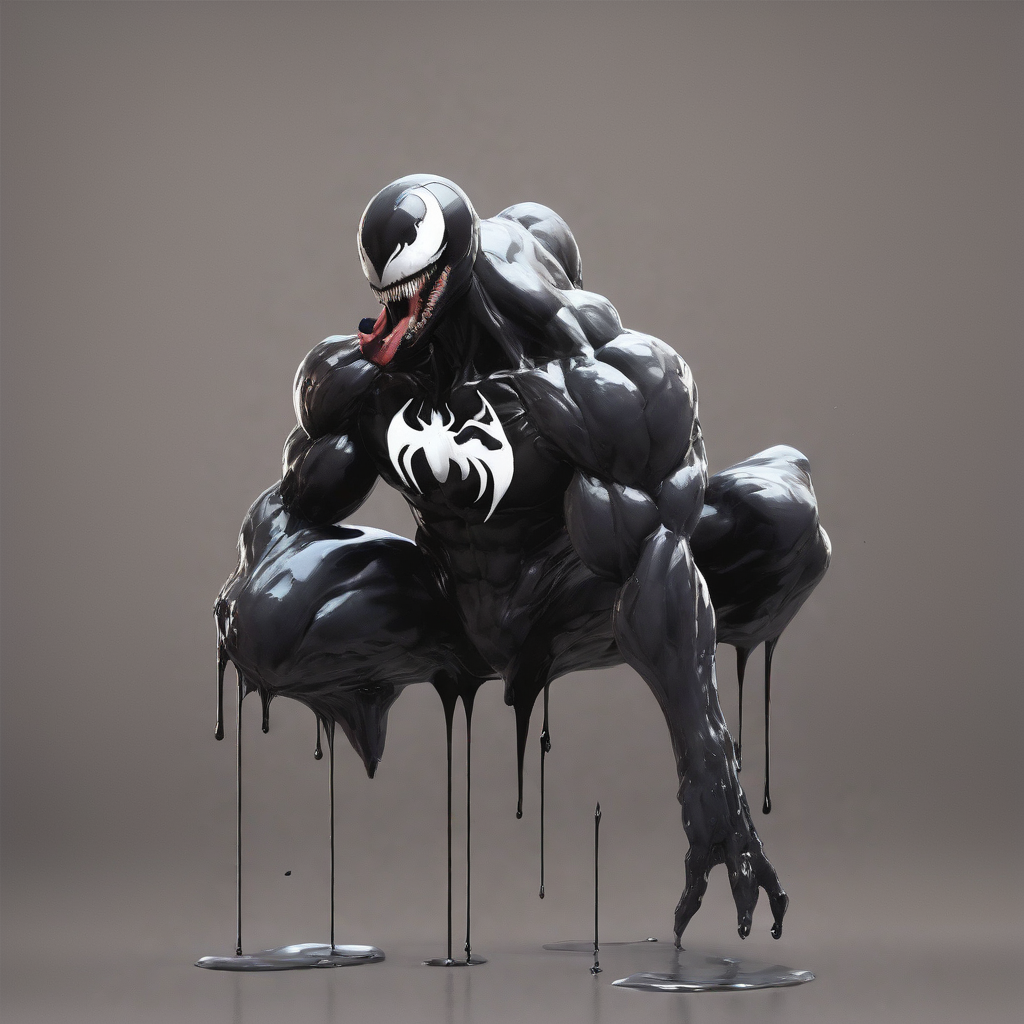}
\includegraphics[width=0.16\textwidth]{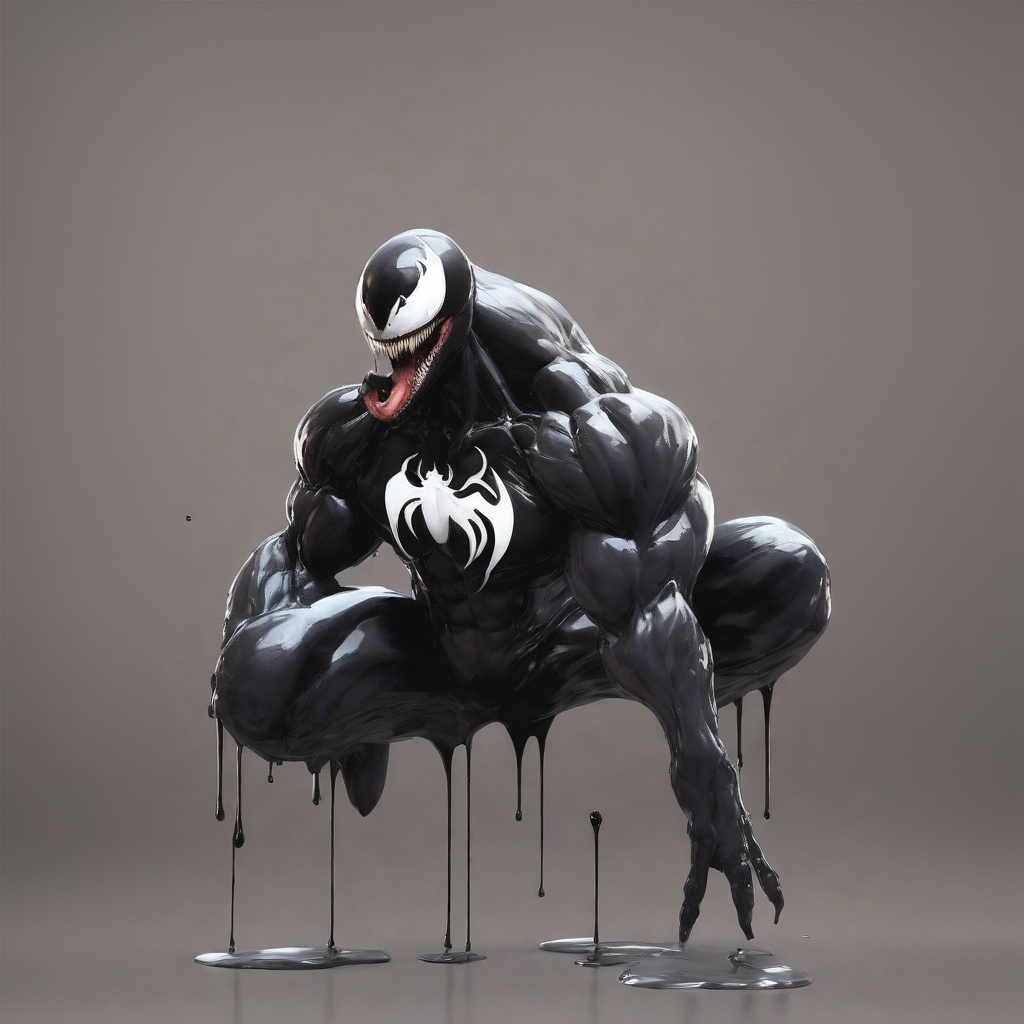}
\includegraphics[width=0.16\textwidth]{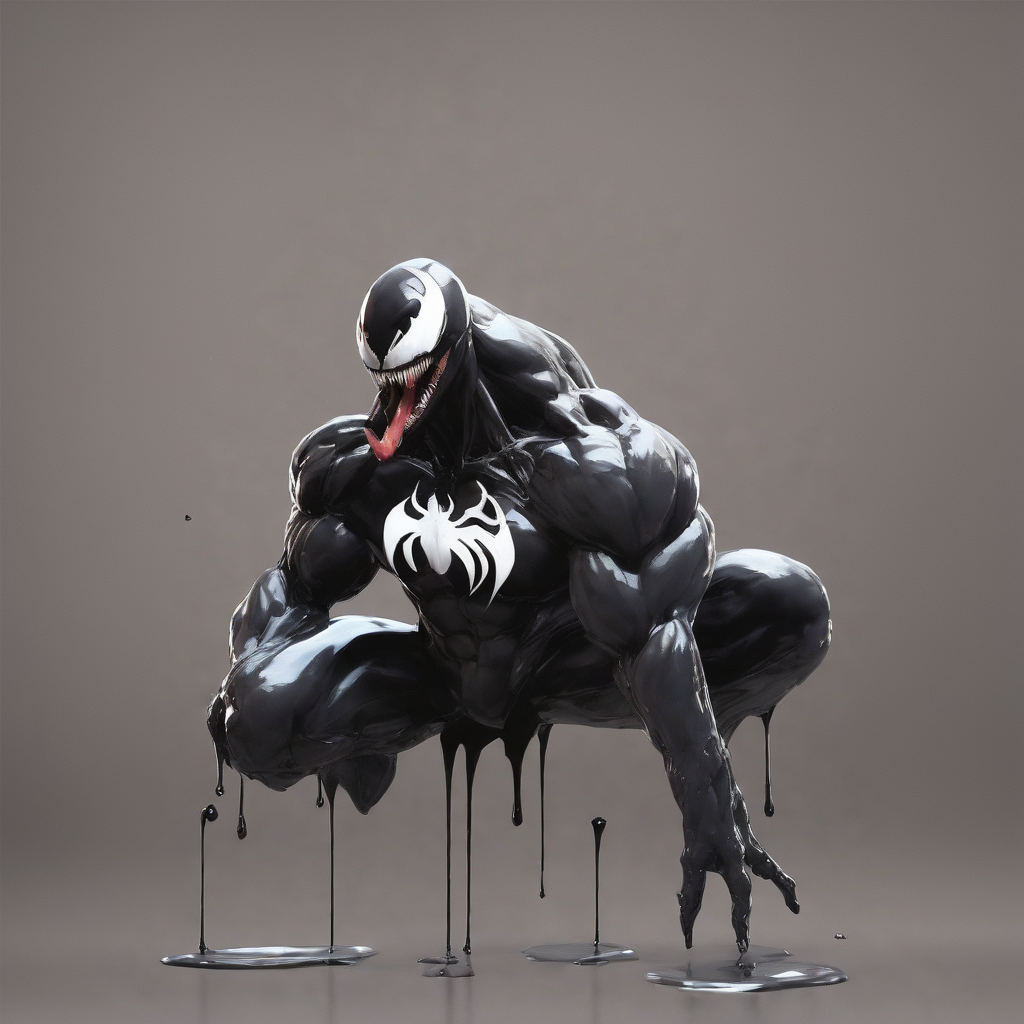}
\caption{Additional videos from \ourmethod. The last row is obtained by applying \ourmethod to SDXL \cite{podell2023sdxl}}
\label{fig:extravideos}
\end{figure}

Fig. \ref{fig:extravideos} shows some additional results of \ourmethod. The first row shows a tree growing. This video was obtained using a simple constant outward-facing radial optical flow applied only on the foliage. Note that while the tree grows, its shadow evolution is coherent. As the input flow is zero in this part of the image, the shadow consistency is recovered only by Stable Diffusion. The last row show a video obtained by applying \ourmethod to SDXL. This shows the generalizability of \ourmethod to different diffusion models, with also different resolutions. Hence, \ourmethod is able to produce high-resolution videos with a high level of detail.

\section{Conclusions}
In this work, we have presented \ourmethod, a novel zero-shot approach for video generation. Our method allows to generate realistic videos with the image prior of Stable Diffusion and a physically-derived optical flow, without any additional training.
\ourmethod warps the noise latent space according to the prescribed flow, and with a modified sampling process exploiting multi-frame cross-attention and the spatial-$\eta$ variable sampling scheme generates novel plausible contents following the prescribed motion and tempoirally consistent. 
For the evaluations of the results, we relied on a standard metric and a proposed one, showing that our method is not only qualitatively but also quantitatively superior to the state-of-the-art of zero-shot video generation.

\newpage\section*{Acknowledgements}
This publication is part of the project PNRR-NGEU which has received funding from the MUR - DM 352/2022. This work was partially supported by the European Union under the Italian National Recovery and Resilience Plan (NRRP) of NextGenerationEU, partnership on \virg{Telecommunications of the Future} (PE00000001 - program \virg{RESTART}).

\bibliographystyle{plainnat}
\bibliography{biblio}

\begin{thebibliography}{36}
\providecommand{\natexlab}[1]{#1}
\providecommand{\url}[1]{\texttt{#1}}
\expandafter\ifx\csname urlstyle\endcsname\relax
  \providecommand{\doi}[1]{doi: #1}\else
  \providecommand{\doi}{doi: \begingroup \urlstyle{rm}\Url}\fi

\bibitem[Bao et~al.(2023)Bao, Qiu, Kang, Zhang, Jin, Wang, and
  Yan]{bao2023latentwarp}
Yuxiang Bao, Di~Qiu, Guoliang Kang, Baochang Zhang, Bo~Jin, Kaiye Wang, and
  Pengfei Yan.
\newblock Latentwarp: Consistent diffusion latents for zero-shot video-to-video
  translation.
\newblock \emph{arXiv preprint arXiv:2311.00353}, 2023.

\bibitem[Bhagwatkar et~al.(2020)Bhagwatkar, Bachu, Fitter, Kulkarni, and
  Chiddarwar]{video_survey}
Rishika Bhagwatkar, Saketh Bachu, Khurshed Fitter, Akshay Kulkarni, and Shital
  Chiddarwar.
\newblock A review of video generation approaches.
\newblock In \emph{2020 International Conference on Power, Instrumentation,
  Control and Computing (PICC)}, pages 1--5, 2020.
\newblock \doi{10.1109/PICC51425.2020.9362485}.

\bibitem[Blattmann et~al.(2023)Blattmann, Rombach, Ling, Dockhorn, Kim, Fidler,
  and Kreis]{blattmann2023align}
Andreas Blattmann, Robin Rombach, Huan Ling, Tim Dockhorn, Seung~Wook Kim,
  Sanja Fidler, and Karsten Kreis.
\newblock Align your latents: High-resolution video synthesis with latent
  diffusion models.
\newblock In \emph{Proceedings of the IEEE/CVF Conference on Computer Vision
  and Pattern Recognition}, pages 22563--22575, 2023.

\bibitem[Brooks et~al.(2023)Brooks, Holynski, and
  Efros]{brooks2023instructpix2pix}
Tim Brooks, Aleksander Holynski, and Alexei~A Efros.
\newblock Instructpix2pix: Learning to follow image editing instructions.
\newblock In \emph{Proceedings of the IEEE/CVF Conference on Computer Vision
  and Pattern Recognition}, pages 18392--18402, 2023.

\bibitem[Brooks et~al.(2024)Brooks, Peebles, Holmes, DePue, Guo, Jing, Schnurr,
  Taylor, Luhman, Luhman, Ng, Wang, and Ramesh]{videoworldsimulators2024}
Tim Brooks, Bill Peebles, Connor Holmes, Will DePue, Yufei Guo, Li~Jing, David
  Schnurr, Joe Taylor, Troy Luhman, Eric Luhman, Clarence Ng, Ricky Wang, and
  Aditya Ramesh.
\newblock Video generation models as world simulators.
\newblock 2024.
\newblock URL
  \url{https://openai.com/research/video-generation-models-as-world-simulators}.

\bibitem[Cai et~al.(2023)Cai, Ceylan, Gadelha, Huang, Wang, and
  Wetzstein]{cai2023generative}
Shengqu Cai, Duygu Ceylan, Matheus Gadelha, Chun-Hao~Paul Huang, Tuanfeng~Yang
  Wang, and Gordon Wetzstein.
\newblock Generative rendering: Controllable 4d-guided video generation with 2d
  diffusion models.
\newblock \emph{arXiv preprint arXiv:2312.01409}, 2023.

\bibitem[Ceylan et~al.(2023)Ceylan, Huang, and Mitra]{ceylan2023pix2video}
Duygu Ceylan, Chun-Hao~P Huang, and Niloy~J Mitra.
\newblock Pix2video: Video editing using image diffusion.
\newblock In \emph{Proceedings of the IEEE/CVF International Conference on
  Computer Vision}, pages 23206--23217, 2023.

\bibitem[Epstein et~al.(2023)Epstein, Jabri, Poole, Efros, and
  Holynski]{epstein2023diffusion}
Dave Epstein, Allan Jabri, Ben Poole, Alexei Efros, and Aleksander Holynski.
\newblock Diffusion self-guidance for controllable image generation.
\newblock \emph{Advances in Neural Information Processing Systems},
  36:\penalty0 16222--16239, 2023.

\bibitem[Foramitti(2021)]{foramitti2021agentpy}
Jo{\"e}l Foramitti.
\newblock Agentpy: A package for agent-based modeling in python.
\newblock \emph{Journal of Open Source Software}, 6\penalty0 (62):\penalty0
  3065, 2021.

\bibitem[Geng and Owens(2024)]{geng2024motion}
Daniel Geng and Andrew Owens.
\newblock Motion guidance: Diffusion-based image editing with differentiable
  motion estimators.
\newblock In \emph{The Twelfth International Conference on Learning
  Representations}, 2024.
\newblock URL \url{https://openreview.net/forum?id=WIAO4vbnNV}.

\bibitem[Geyer et~al.(2024)Geyer, Bar-Tal, Bagon, and
  Dekel]{geyer2023tokenflow}
Michal Geyer, Omer Bar-Tal, Shai Bagon, and Tali Dekel.
\newblock Tokenflow: Consistent diffusion features for consistent video
  editing.
\newblock In \emph{The Twelfth International Conference on Learning
  Representations}, 2024.
\newblock URL \url{https://openreview.net/forum?id=lKK50q2MtV}.

\bibitem[Graphics and Lab(2022)]{MSUDataset}
MSU Graphics and Media Lab.
\newblock Msu video frame interpolation benchmark dataset, 2022.
\newblock URL
  \url{https://videoprocessing.ai/benchmarks/video-frame-interpolation-dataset.html}.

\bibitem[Hertz et~al.(2023)Hertz, Mokady, Tenenbaum, Aberman, Pritch, and
  Cohen-or]{hertz2022prompt}
Amir Hertz, Ron Mokady, Jay Tenenbaum, Kfir Aberman, Yael Pritch, and Daniel
  Cohen-or.
\newblock Prompt-to-prompt image editing with cross-attention control.
\newblock In \emph{The Eleventh International Conference on Learning
  Representations}, 2023.
\newblock URL \url{https://openreview.net/forum?id=_CDixzkzeyb}.

\bibitem[Ho and Salimans(2021)]{ho2022classifier}
Jonathan Ho and Tim Salimans.
\newblock Classifier-free diffusion guidance.
\newblock In \emph{NeurIPS 2021 Workshop on Deep Generative Models and
  Downstream Applications}, 2021.
\newblock URL \url{https://openreview.net/forum?id=qw8AKxfYbI}.

\bibitem[Ho et~al.(2020)Ho, Jain, and Abbeel]{ho2020denoising}
Jonathan Ho, Ajay Jain, and Pieter Abbeel.
\newblock Denoising diffusion probabilistic models.
\newblock \emph{Advances in Neural Information Processing Systems},
  33:\penalty0 6840--6851, 2020.

\bibitem[Ho et~al.(2022{\natexlab{a}})Ho, Chan, Saharia, Whang, Gao, Gritsenko,
  Kingma, Poole, Norouzi, Fleet, et~al.]{ho2022imagen}
Jonathan Ho, William Chan, Chitwan Saharia, Jay Whang, Ruiqi Gao, Alexey
  Gritsenko, Diederik~P Kingma, Ben Poole, Mohammad Norouzi, David~J Fleet,
  et~al.
\newblock Imagen video: High definition video generation with diffusion models.
\newblock \emph{arXiv preprint arXiv:2210.02303}, 2022{\natexlab{a}}.

\bibitem[Ho et~al.(2022{\natexlab{b}})Ho, Salimans, Gritsenko, Chan, Norouzi,
  and Fleet]{ho2022video}
Jonathan Ho, Tim Salimans, Alexey Gritsenko, William Chan, Mohammad Norouzi,
  and David~J Fleet.
\newblock Video diffusion models.
\newblock \emph{Advances in Neural Information Processing Systems},
  35:\penalty0 8633--8646, 2022{\natexlab{b}}.

\bibitem[Holl et~al.(2020)Holl, Thuerey, and Koltun]{holl2020learning}
Philipp Holl, Nils Thuerey, and Vladlen Koltun.
\newblock Learning to control pdes with differentiable physics.
\newblock In \emph{International Conference on Learning Representations}, 2020.
\newblock URL \url{https://openreview.net/forum?id=HyeSin4FPB}.

\bibitem[Itseez(2015)]{itseez2015opencv}
Itseez.
\newblock Open source computer vision library.
\newblock \url{https://github.com/itseez/opencv}, 2015.

\bibitem[Khachatryan et~al.(2023)Khachatryan, Movsisyan, Tadevosyan, Henschel,
  Wang, Navasardyan, and Shi]{khachatryan2023text2video}
Levon Khachatryan, Andranik Movsisyan, Vahram Tadevosyan, Roberto Henschel,
  Zhangyang Wang, Shant Navasardyan, and Humphrey Shi.
\newblock Text2video-zero: Text-to-image diffusion models are zero-shot video
  generators.
\newblock In \emph{Proceedings of the IEEE/CVF International Conference on
  Computer Vision}, pages 15954--15964, 2023.

\bibitem[Mokady et~al.(2023)Mokady, Hertz, Aberman, Pritch, and
  Cohen-Or]{mokady2023null}
Ron Mokady, Amir Hertz, Kfir Aberman, Yael Pritch, and Daniel Cohen-Or.
\newblock Null-text inversion for editing real images using guided diffusion
  models.
\newblock In \emph{Proceedings of the IEEE/CVF Conference on Computer Vision
  and Pattern Recognition}, pages 6038--6047, 2023.

\bibitem[Ni et~al.(2023)Ni, Shi, Li, Huang, and Min]{ni2023conditional}
Haomiao Ni, Changhao Shi, Kai Li, Sharon~X Huang, and Martin~Renqiang Min.
\newblock Conditional image-to-video generation with latent flow diffusion
  models.
\newblock In \emph{Proceedings of the IEEE/CVF Conference on Computer Vision
  and Pattern Recognition}, pages 18444--18455, 2023.

\bibitem[Podell et~al.(2023)Podell, English, Lacey, Blattmann, Dockhorn,
  M{\"u}ller, Penna, and Rombach]{podell2023sdxl}
Dustin Podell, Zion English, Kyle Lacey, Andreas Blattmann, Tim Dockhorn, Jonas
  M{\"u}ller, Joe Penna, and Robin Rombach.
\newblock Sdxl: Improving latent diffusion models for high-resolution image
  synthesis.
\newblock \emph{arXiv preprint arXiv:2307.01952}, 2023.

\bibitem[Reynolds(1987)]{reynolds1987flocks}
Craig~W Reynolds.
\newblock Flocks, herds and schools: A distributed behavioral model.
\newblock In \emph{Proceedings of the 14th annual conference on Computer
  graphics and interactive techniques}, pages 25--34, 1987.

\bibitem[Rombach et~al.(2022)Rombach, Blattmann, Lorenz, Esser, and
  Ommer]{rombach2022high}
Robin Rombach, Andreas Blattmann, Dominik Lorenz, Patrick Esser, and Bj{\"o}rn
  Ommer.
\newblock High-resolution image synthesis with latent diffusion models.
\newblock In \emph{Proceedings of the IEEE/CVF Conference on Computer Vision
  and Pattern Recognition}, pages 10684--10695, 2022.

\bibitem[Ronneberger et~al.(2015)Ronneberger, Fischer, and
  Brox]{ronneberger2015u}
Olaf Ronneberger, Philipp Fischer, and Thomas Brox.
\newblock U-net: Convolutional networks for biomedical image segmentation.
\newblock In \emph{Medical image computing and computer-assisted
  intervention--MICCAI 2015: 18th international conference, Munich, Germany,
  October 5-9, 2015, proceedings, part III 18}, pages 234--241. Springer, 2015.

\bibitem[Singer et~al.(2023)Singer, Polyak, Hayes, Yin, An, Zhang, Hu, Yang,
  Ashual, Gafni, Parikh, Gupta, and Taigman]{singer2022make}
Uriel Singer, Adam Polyak, Thomas Hayes, Xi~Yin, Jie An, Songyang Zhang, Qiyuan
  Hu, Harry Yang, Oron Ashual, Oran Gafni, Devi Parikh, Sonal Gupta, and Yaniv
  Taigman.
\newblock Make-a-video: Text-to-video generation without text-video data.
\newblock In \emph{The Eleventh International Conference on Learning
  Representations}, 2023.
\newblock URL \url{https://openreview.net/forum?id=nJfylDvgzlq}.

\bibitem[Sohl-Dickstein et~al.(2015)Sohl-Dickstein, Weiss, Maheswaranathan, and
  Ganguli]{sohl2015deep}
Jascha Sohl-Dickstein, Eric Weiss, Niru Maheswaranathan, and Surya Ganguli.
\newblock Deep unsupervised learning using nonequilibrium thermodynamics.
\newblock In \emph{International Conference on Machine Learning}, pages
  2256--2265. PMLR, 2015.

\bibitem[Song et~al.(2020{\natexlab{a}})Song, Meng, and
  Ermon]{song2020denoising}
Jiaming Song, Chenlin Meng, and Stefano Ermon.
\newblock {Denoising Diffusion Implicit Models}.
\newblock In \emph{International Conference on Learning Representations},
  2020{\natexlab{a}}.

\bibitem[Song and Ermon(2019)]{song2019generative}
Yang Song and Stefano Ermon.
\newblock Generative modeling by estimating gradients of the data distribution.
\newblock \emph{Advances in Neural Information Processing Systems}, 32, 2019.

\bibitem[Song et~al.(2020{\natexlab{b}})Song, Sohl-Dickstein, Kingma, Kumar,
  Ermon, and Poole]{song2020score}
Yang Song, Jascha Sohl-Dickstein, Diederik~P Kingma, Abhishek Kumar, Stefano
  Ermon, and Ben Poole.
\newblock Score-based generative modeling through stochastic differential
  equations.
\newblock In \emph{International Conference on Learning Representations},
  2020{\natexlab{b}}.

\bibitem[Teed and Deng(2020)]{teed2020raft}
Zachary Teed and Jia Deng.
\newblock Raft: Recurrent all-pairs field transforms for optical flow.
\newblock In \emph{Computer Vision--ECCV 2020: 16th European Conference,
  Glasgow, UK, August 23--28, 2020, Proceedings, Part II 16}, pages 402--419.
  Springer, 2020.

\bibitem[Van Den~Oord et~al.(2017)Van Den~Oord, Vinyals, et~al.]{van2017neural}
Aaron Van Den~Oord, Oriol Vinyals, et~al.
\newblock Neural discrete representation learning.
\newblock \emph{Advances in neural information processing systems}, 30, 2017.

\bibitem[Wang et~al.(2004)Wang, Bovik, Sheikh, and Simoncelli]{wang2004image}
Zhou Wang, Alan~C Bovik, Hamid~R Sheikh, and Eero~P Simoncelli.
\newblock Image quality assessment: from error visibility to structural
  similarity.
\newblock \emph{IEEE transactions on image processing}, 13\penalty0
  (4):\penalty0 600--612, 2004.

\bibitem[Wu et~al.(2023)Wu, Ge, Wang, Lei, Gu, Shi, Hsu, Shan, Qie, and
  Shou]{wu2023tune}
Jay~Zhangjie Wu, Yixiao Ge, Xintao Wang, Stan~Weixian Lei, Yuchao Gu, Yufei
  Shi, Wynne Hsu, Ying Shan, Xiaohu Qie, and Mike~Zheng Shou.
\newblock Tune-a-video: One-shot tuning of image diffusion models for
  text-to-video generation.
\newblock In \emph{Proceedings of the IEEE/CVF International Conference on
  Computer Vision}, pages 7623--7633, 2023.

\bibitem[Zhang et~al.(2023)Zhang, Rao, and Agrawala]{zhang2023adding}
Lvmin Zhang, Anyi Rao, and Maneesh Agrawala.
\newblock Adding conditional control to text-to-image diffusion models.
\newblock In \emph{Proceedings of the IEEE/CVF International Conference on
  Computer Vision}, pages 3836--3847, 2023.

\end{thebibliography}

\newpage

\appendix
\section{Background}
\label{sec:background}

DDMs (Denoising Diffusion Models) \citep{sohl2015deep,song2019generative,ho2020denoising}  represents a generative modeling approach that leverage a noise diffusion process to model a data distribution starting from random noise. These models are based on a predefined Markovian forward noising chain that progressively adds Gaussian noise to the data $\vx_0$ in an iterative procedure of $T$ steps.
The reverse diffusion process traverses back the Markov Chain and can be written as:
\begin{align}
    p_\theta(\vx_{0:T}) = p(\vx_T) \prod\nolimits_{t=1}^T p_\theta(\vx_{t-1} \mid \vx_t) \qquad
    p_\theta(\vx_{t-1} \mid \vx_{t}) =\mathcal{N}(\vx_{t-1} \mid \mu_{\theta}({\vx}_{t}, t), \sigma_t^2\bm{I})~. 
\end{align}
The training phase optimizes the parameters of the reverse process $p_\theta$ maximising an evidence lower bound (ELBO) over  the target data.
The work of \cite{song2020denoising} shows that is possible to construct a non-Markovian process defining a faster sampler (DDIM) that is compatible with the pretrained model.
So starting from $p_\theta(\vx_{0:T})$, it is possible to sample $\vx_{t-1}$ using:
\vspace{0pt}
\begin{align}
    x_{t-1} &= \sqrt{\alpha_{t-1}} \left(\frac{x_t - \sqrt{1-\alpha_t}\hat{\epsilon}_t}{\sqrt{\alpha_t}}\right) + \sqrt{1-\alpha_{t-1}-\sigma_{t}(\eta)^2} \cdot \hat{\epsilon}_t+\sigma_{t}(\eta)\varepsilon_t
\end{align}
where $\sigma_{t}(\eta) = \eta \sqrt{\frac{1-\alpha_{t-1}}{1-\alpha_{t}}}\sqrt{\frac{1-\alpha_{t}}{\alpha_{t-1}}}$ and $\eta \in (0,1)$ is a parameter controlling the forward process, when $\eta = 0$, the sampling becomes deterministic, when $\eta = 1$, the process result in DDPM sampling. $\hat{\epsilon}_t$ is the estimated noise present in $x_t$, typically estimated with a UNet architecture \cite{ronneberger2015u}: $\epsilon_t(\cdot)$. Finally, $\varepsilon_t$ is an independent normal stochastic variable.
In this work we employ a Latent Diffusion Model \cite{rombach2022high} that perform the diffusion process over a compressed latent space, reducing the computational burden of training in pixel space, while keeping high perceptual quality. Before the diffusion process, a VQ-VAE \cite{van2017neural} is trained; the input image is then encoded by the VQ-VAE Encoder $\mathcal{E}$ that reduces the spatial dimension. The generated features are decoded back to the image space when generating images by means of th VQ-VAE Decoder $\mathcal{D}$.
The UNet architecture is tipically composed by convolutional layers followed by  spatial self-attention layers and cross-attention conditioning layers.
Recent works \citep{khachatryan2023text2video, hertz2022prompt, brooks2023instructpix2pix} propose to reprogram this mechanism to enhance consistency between frames by letting the currently generated frame to attend to the first frame by swapping the original attention keys (K) and values (V) with the keys and values of the first frame, leading to the Cross-Frame Attention (CFA) mechanism:
\begin{align}
\text{Cross-Frame-Attn(Q,K,V)} = \text{Softmax}\left( \frac{Q^{f} \cdot K^{1}}{\sqrt{d_k}} \right)V^{1}
\end{align}
where $V^{1}$ and $K^{1}$ represent the keys and values of the first frame, while $Q^{f}$ represents the queries of the current frame, and $d_k$ is the channel dimension of the keys.
In this work we will use the notation $\epsilon_t(z, \mathcal{P} ; \{a,b,c,\dots\})$, where $z$ is a latent, $\mathcal{P}$ is the prompt, and $\{a,b,c,\dots\}$ is a \textit{list} of latents to attend to, as MCFA enables to attends to a list of latents and not only to a single one.

Classifier-Free Guidance (CFG) \cite{ho2022classifier} is a widely used technique to guide conditional generation process using a linear combination of conditional and unconditonal estimated scores:
\begin{equation}
    \hat{\epsilon} = \epsilon_t(z, \mathcal{P}_{\emptyset},\{\dots\}) + \gamma \left[\epsilon_t(z, \mathcal{P}, \{\dots\}) - \epsilon_t(z, \mathcal{P}_{\emptyset},\{\dots\})\right]
\end{equation}
where $\gamma$ is the scaling factor, $\mathcal{P}_{\emptyset}$ represents the null condition and $\mathcal{P}$ is the target text prompt.

\section{Extendend Ablation Study}
\begin{figure*}[!ht]
    \centering
    \begin{subfigure}[t]{\textwidth}
        \centering
        \includegraphics[width=0.16\textwidth]{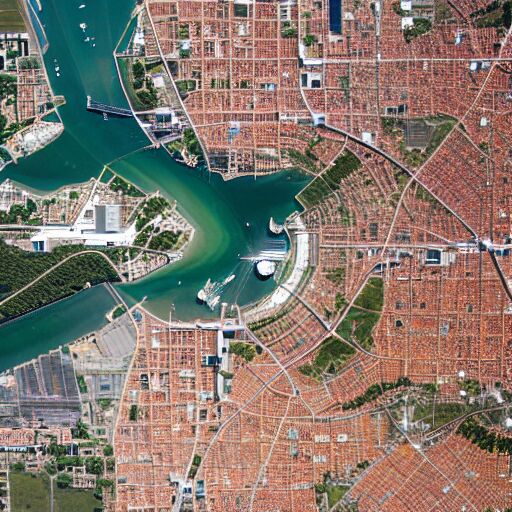}
        \includegraphics[width=0.16\textwidth]{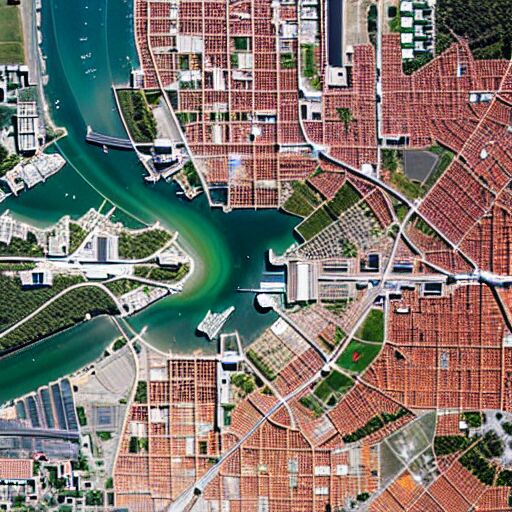}
        \includegraphics[width=0.16\textwidth]{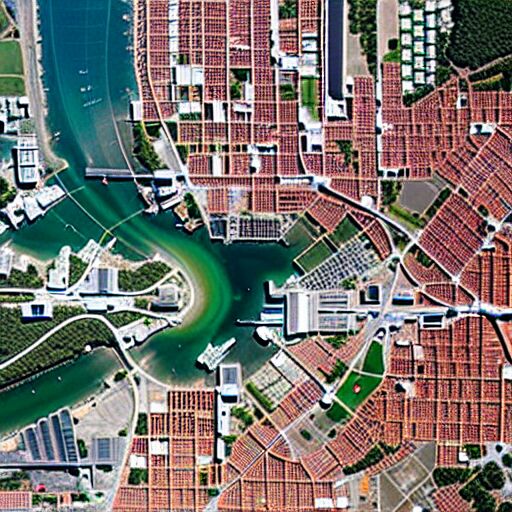}
        \includegraphics[width=0.16\textwidth]{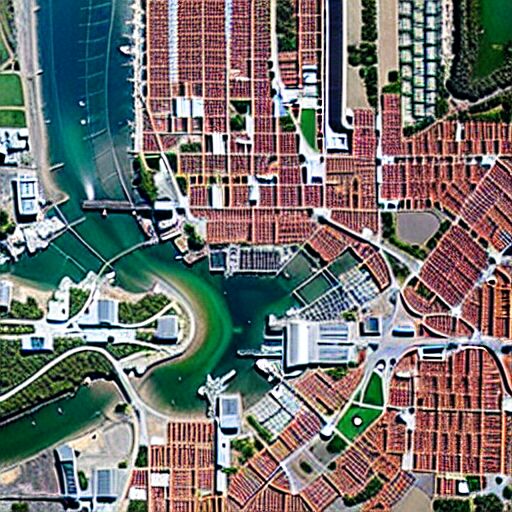}
        \includegraphics[width=0.16\textwidth]{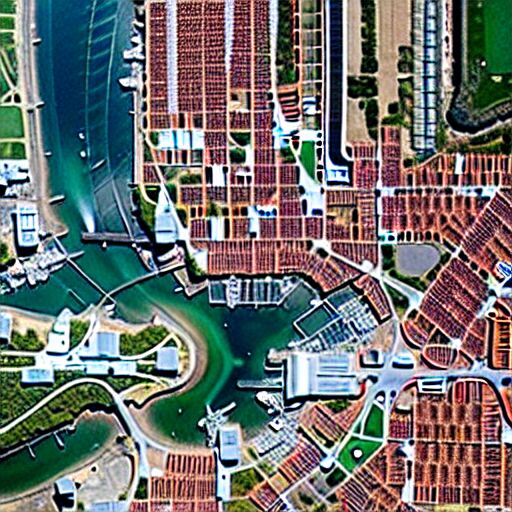}
        \includegraphics[width=0.16\textwidth]{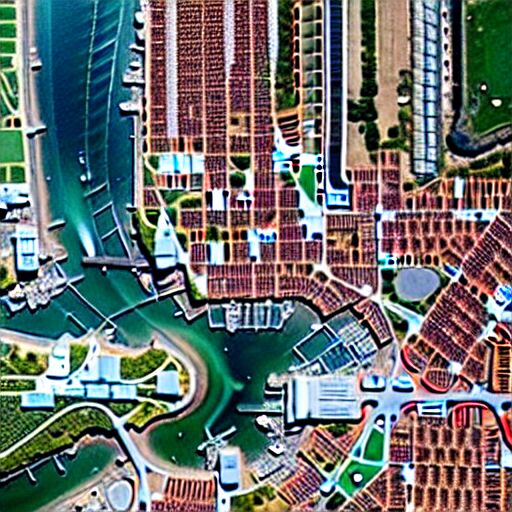}
    \end{subfigure}\hfill
    \begin{subfigure}[t]{\textwidth}
        \centering
        \includegraphics[width=0.16\textwidth]{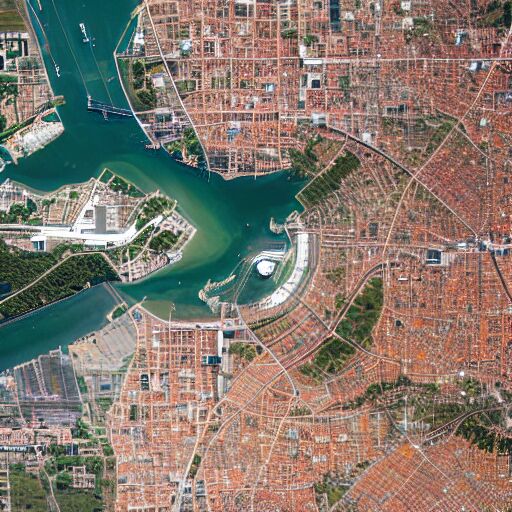}
        \includegraphics[width=0.16\textwidth]{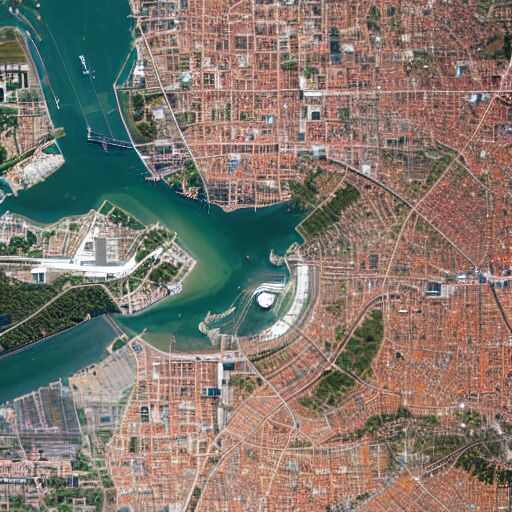}
        \includegraphics[width=0.16\textwidth]{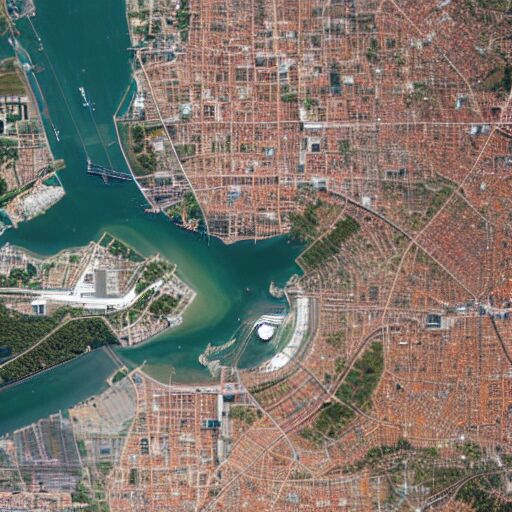}
        \includegraphics[width=0.16\textwidth]{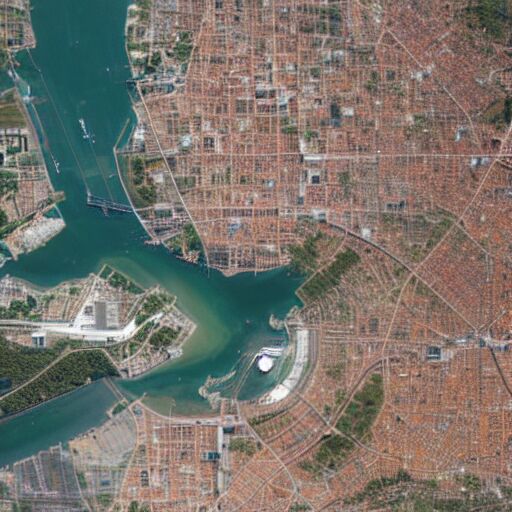}
        \includegraphics[width=0.16\textwidth]{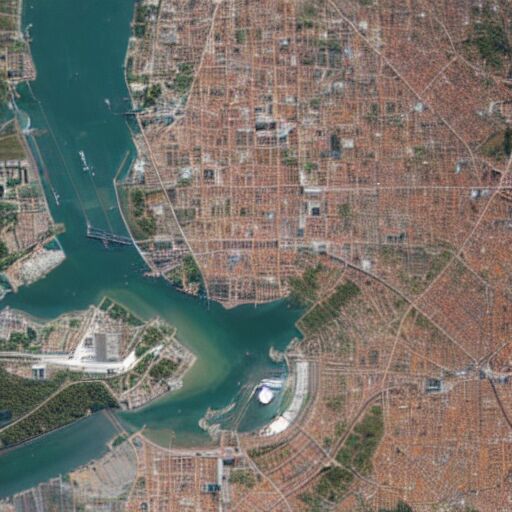}
        \includegraphics[width=0.16\textwidth]{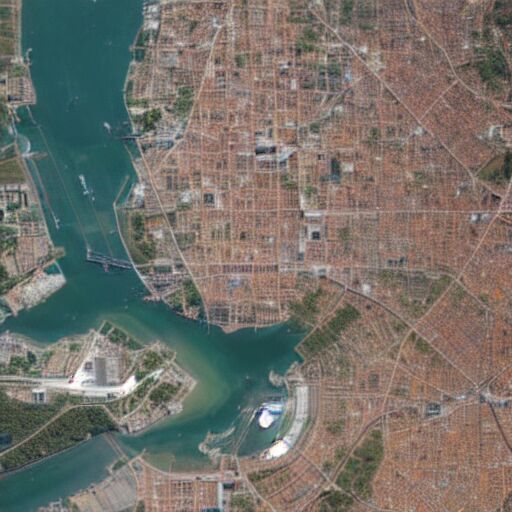}
    \end{subfigure}\hfill
    \begin{subfigure}[t]{\textwidth}
        \centering
        \includegraphics[width=0.16\textwidth]{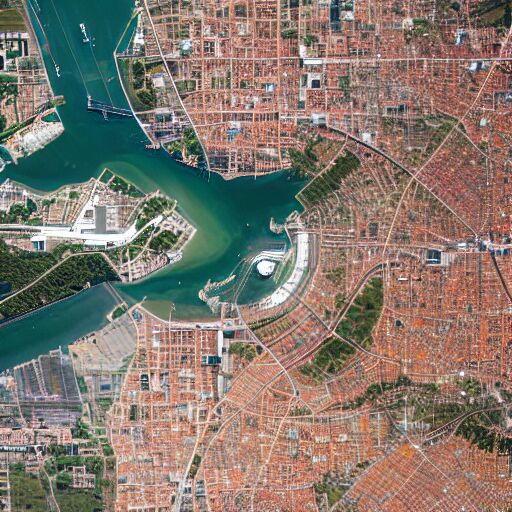}
        \includegraphics[width=0.16\textwidth]{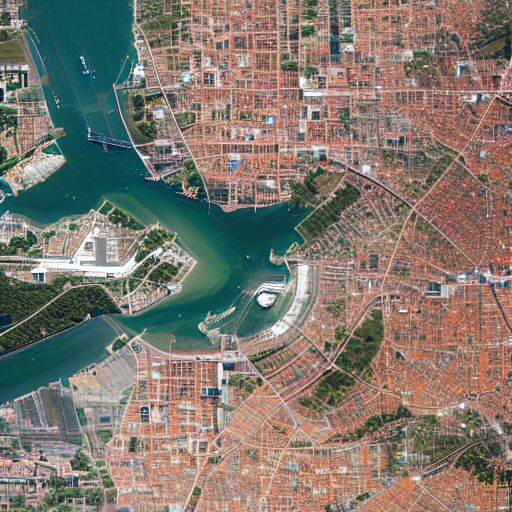}
        \includegraphics[width=0.16\textwidth]{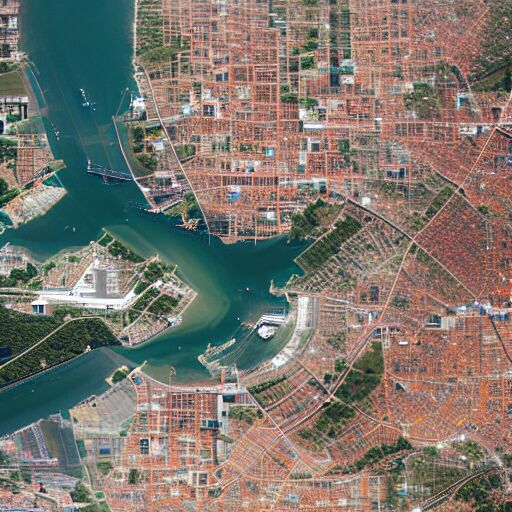}
        \includegraphics[width=0.16\textwidth]{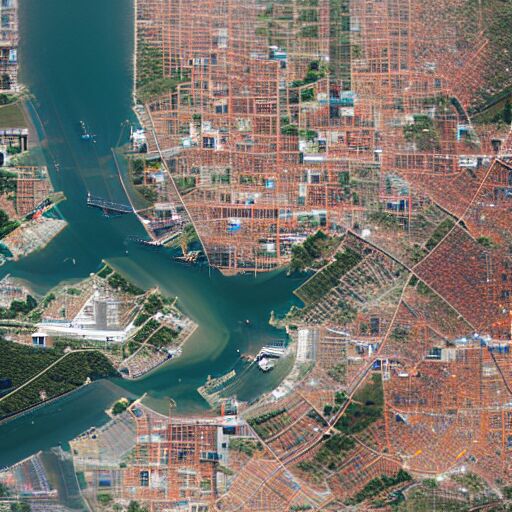}
        \includegraphics[width=0.16\textwidth]{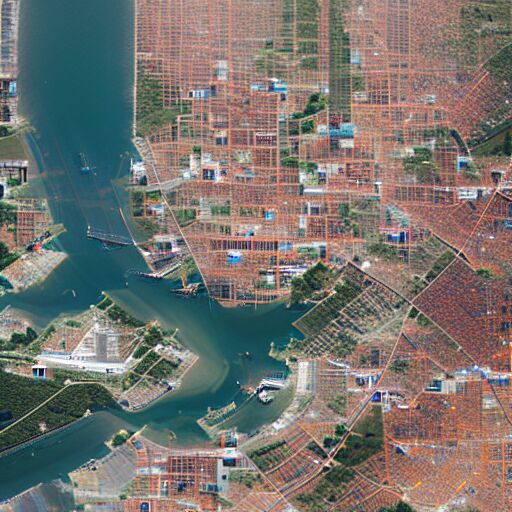}
        \includegraphics[width=0.16\textwidth]{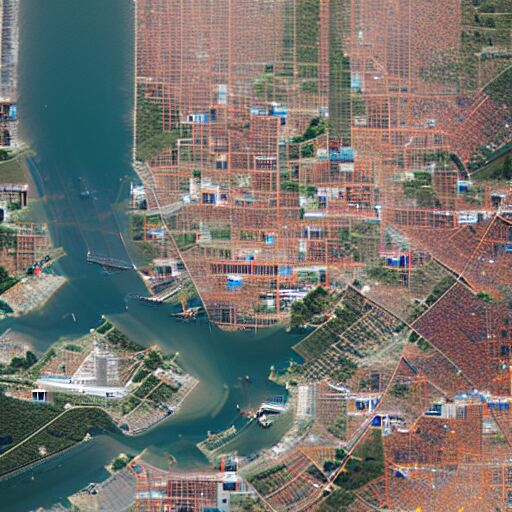}
    \end{subfigure}\hfill
    \begin{subfigure}[t]{\textwidth}
        \centering
        \includegraphics[width=0.16\textwidth]{figures/final/satellite/frame_000_2}
        \includegraphics[width=0.16\textwidth]{figures/final/satellite/frame_004_2}
        \includegraphics[width=0.16\textwidth]{figures/final/satellite/frame_008_2}
        \includegraphics[width=0.16\textwidth]{figures/final/satellite/frame_012_2}
        \includegraphics[width=0.16\textwidth]{figures/final/satellite/frame_016_2}
        \includegraphics[width=0.16\textwidth]{figures/final/satellite/frame_019_2}
    \end{subfigure}
    \caption{Ablation - Cross-Frame attention. First row: no cross frame attention; Second Row: Attend only to the initial frame; Third Row: Attend only to the previous frame; Fourth Row: Attend to the initial and preceding frame (ours).}
    \label{fig:satellite_abl_cross}
\end{figure*}

\begin{figure*}[!ht]
    \centering
    \begin{subfigure}[t]{\textwidth}
        \centering
        \includegraphics[width=0.16\textwidth]{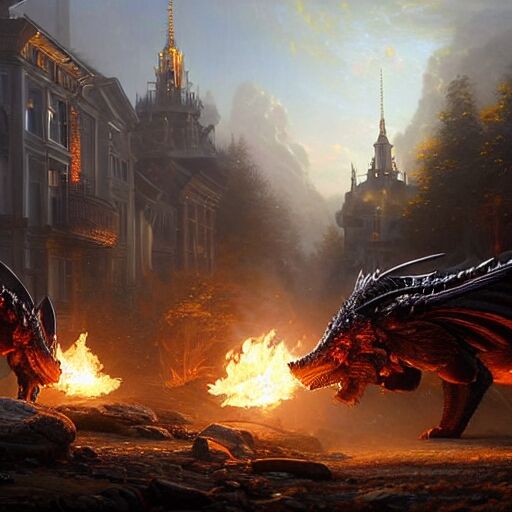}
        \includegraphics[width=0.16\textwidth]{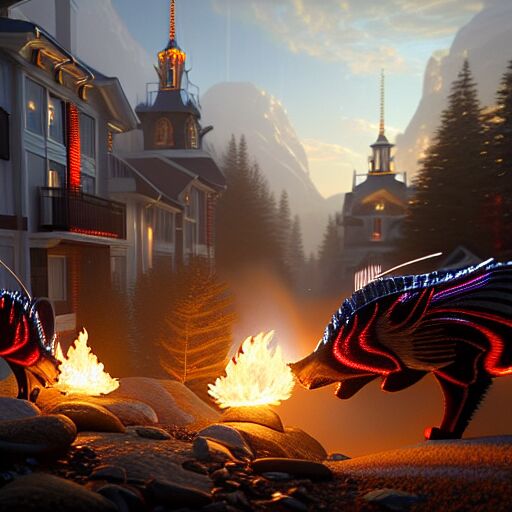}
        \includegraphics[width=0.16\textwidth]{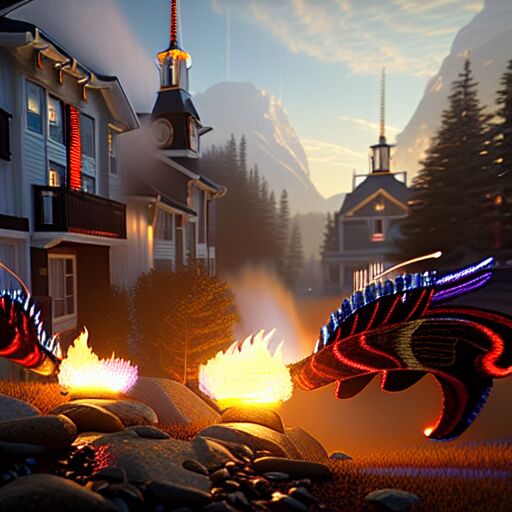}
        \includegraphics[width=0.16\textwidth]{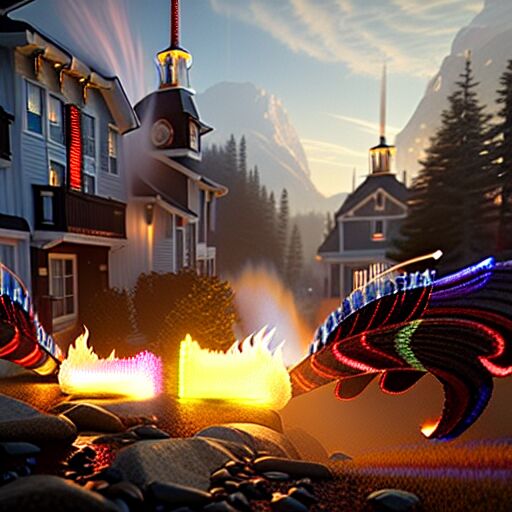}
        \includegraphics[width=0.16\textwidth]{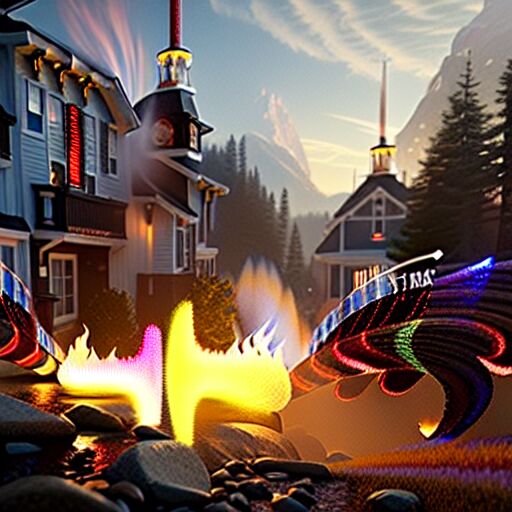}
        \includegraphics[width=0.16\textwidth]{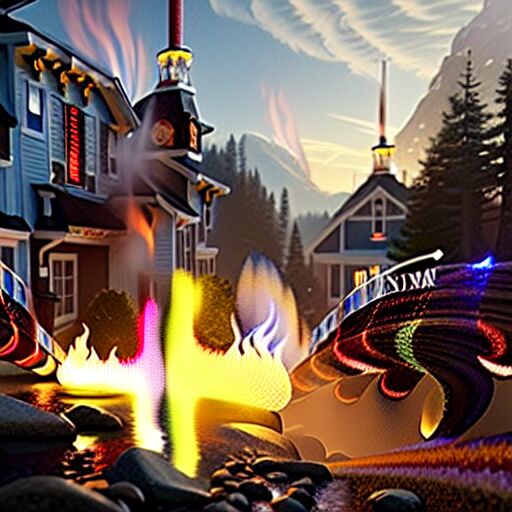}
    \end{subfigure}\hfill
    \begin{subfigure}[t]{\textwidth}
        \centering
        \includegraphics[width=0.16\textwidth]{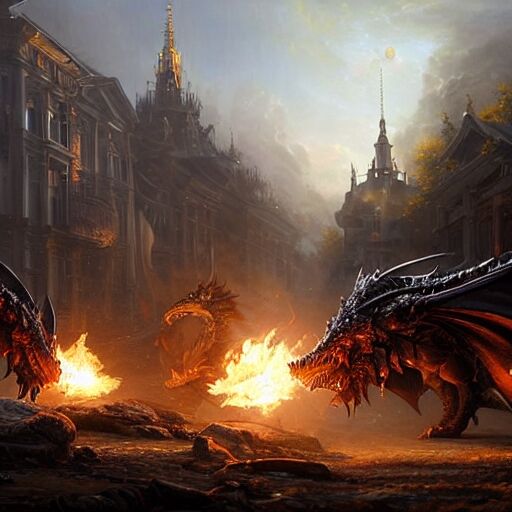}
        \includegraphics[width=0.16\textwidth]{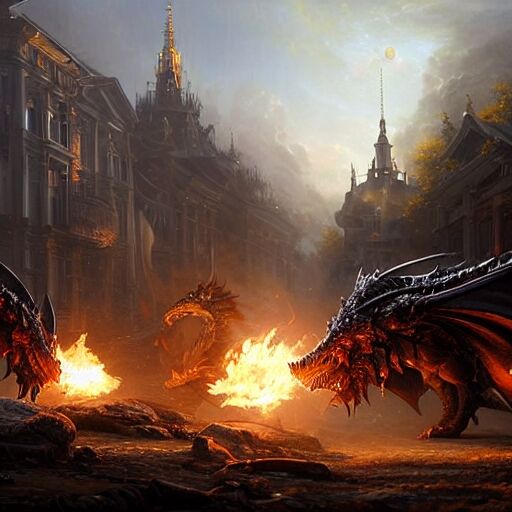}
        \includegraphics[width=0.16\textwidth]{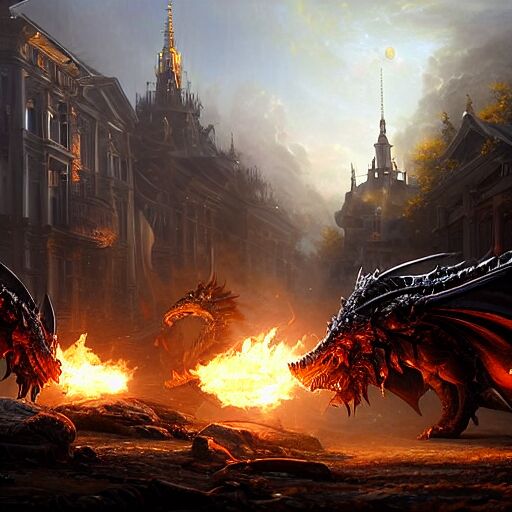}
        \includegraphics[width=0.16\textwidth]{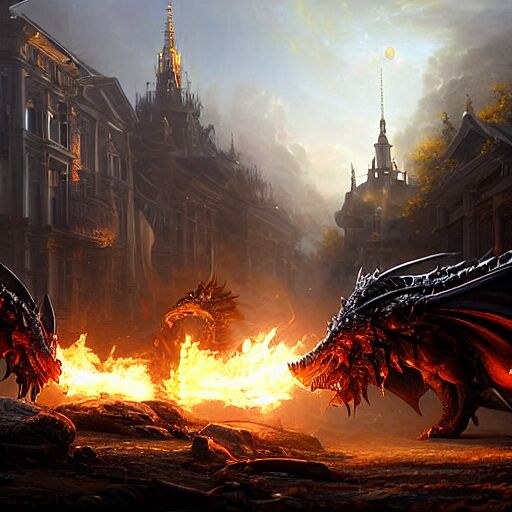}
        \includegraphics[width=0.16\textwidth]{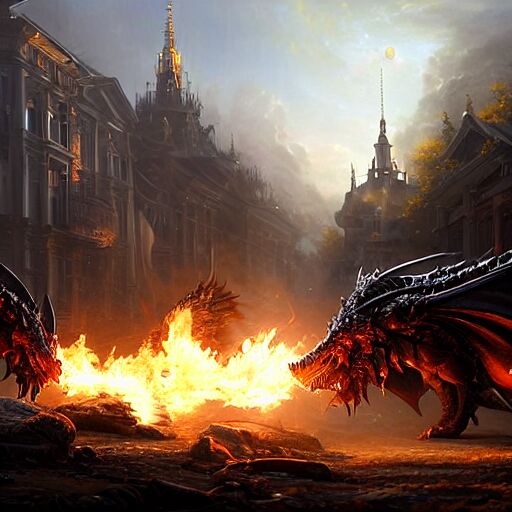}
        \includegraphics[width=0.16\textwidth]{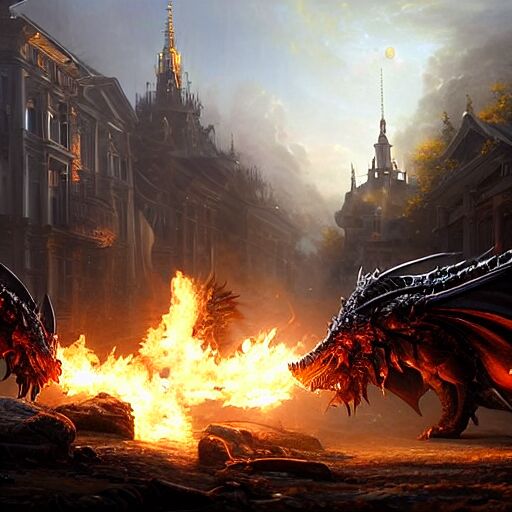}
    \end{subfigure}\hfill
    \begin{subfigure}[t]{\textwidth}
        \centering
        \includegraphics[width=0.16\textwidth]{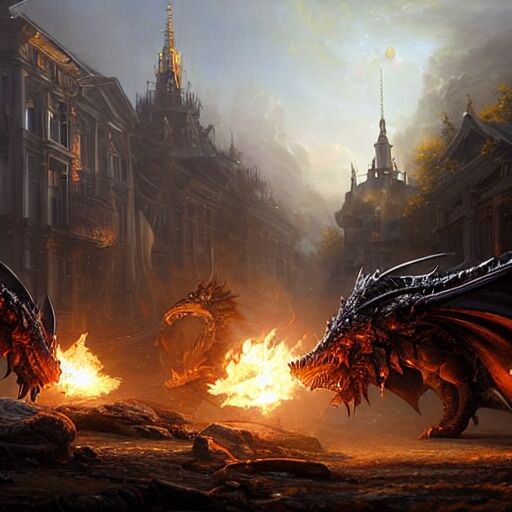}
        \includegraphics[width=0.16\textwidth]{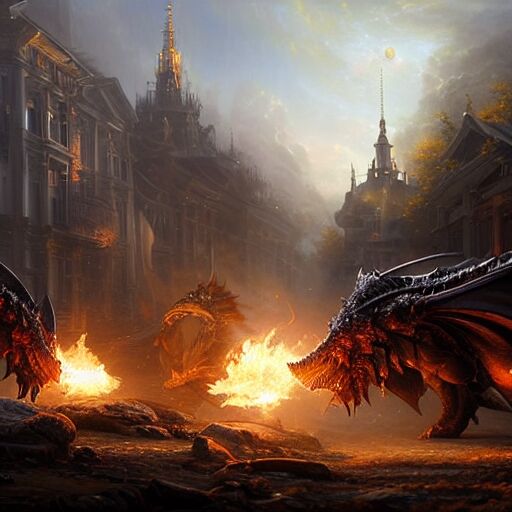}
        \includegraphics[width=0.16\textwidth]{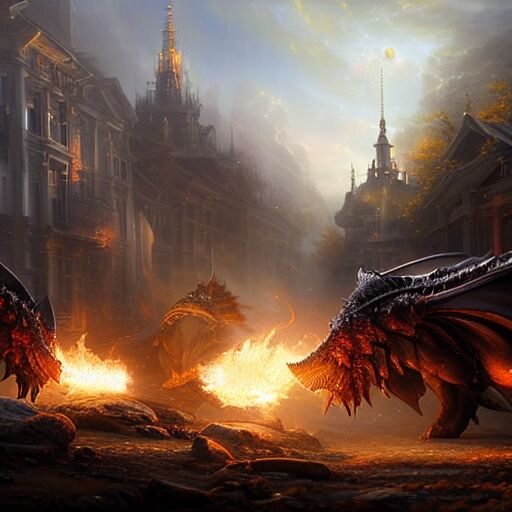}
        \includegraphics[width=0.16\textwidth]{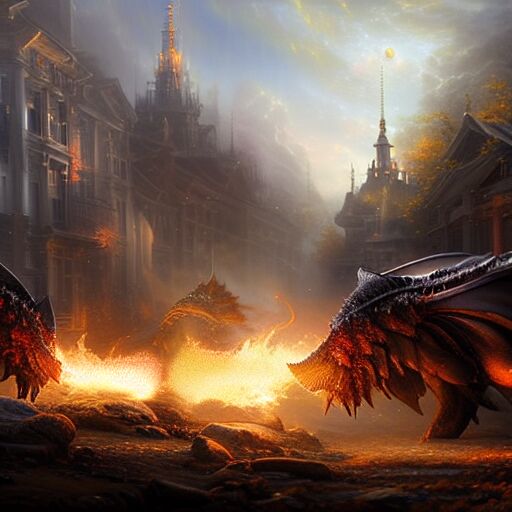}
        \includegraphics[width=0.16\textwidth]{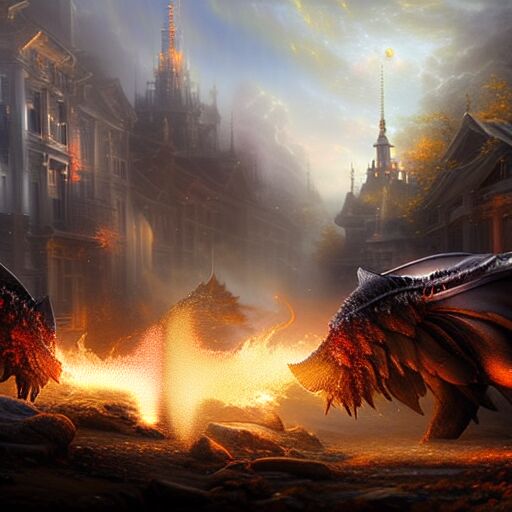}
        \includegraphics[width=0.16\textwidth]{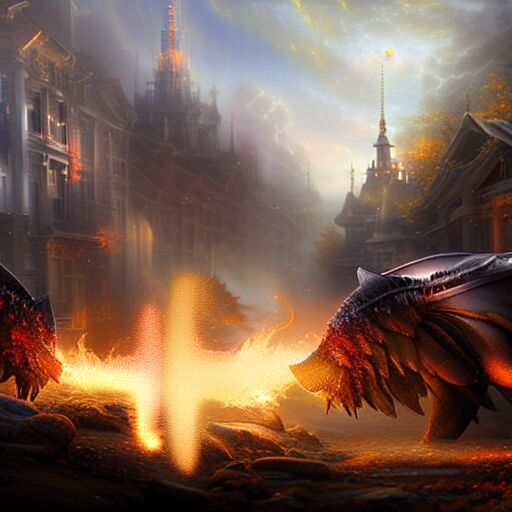}
    \end{subfigure}\hfill
    \begin{subfigure}[t]{\textwidth}
        \centering
        \includegraphics[width=0.16\textwidth]{figures/final/dragons/frame_000_2}
        \includegraphics[width=0.16\textwidth]{figures/final/dragons/frame_004_2}
        \includegraphics[width=0.16\textwidth]{figures/final/dragons/frame_008_2}
        \includegraphics[width=0.16\textwidth]{figures/final/dragons/frame_012_2}
        \includegraphics[width=0.16\textwidth]{figures/final/dragons/frame_016_2}
        \includegraphics[width=0.16\textwidth]{figures/final/dragons/frame_020_2}
    \end{subfigure}
    \caption{Ablation - Cross-Frame attention. First row: no cross frame attention; Second Row: Attend only to the initial frame; Third Row: Attend only to the previous frame; Fourth Row: Attend to the initial and preceding frame (ours).}
    \label{fig:dragons_abl_cross}
\end{figure*}

\begin{figure*}[!ht]
    \centering
    \begin{subfigure}[t]{\textwidth}
        \centering
        \includegraphics[width=0.16\textwidth]{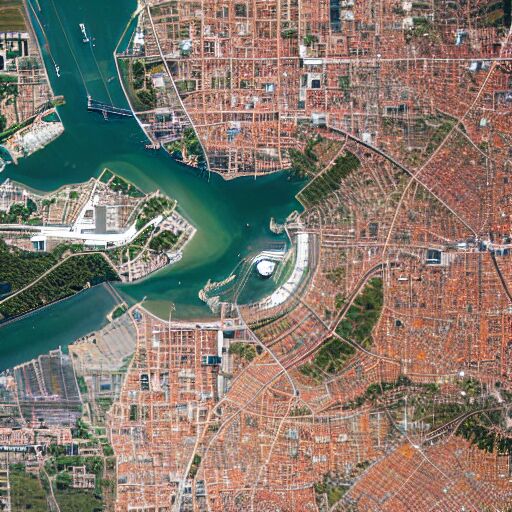}
        \includegraphics[width=0.16\textwidth]{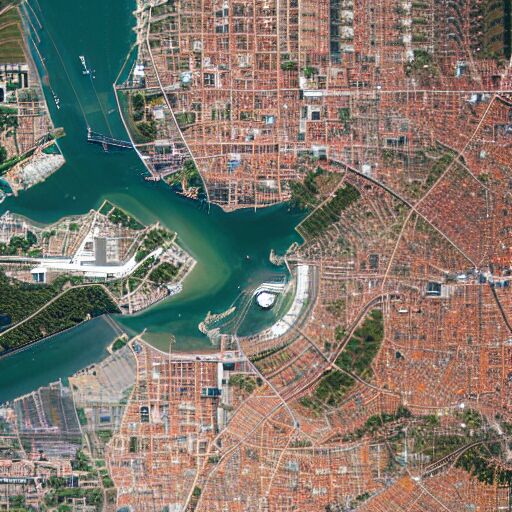}
        \includegraphics[width=0.16\textwidth]{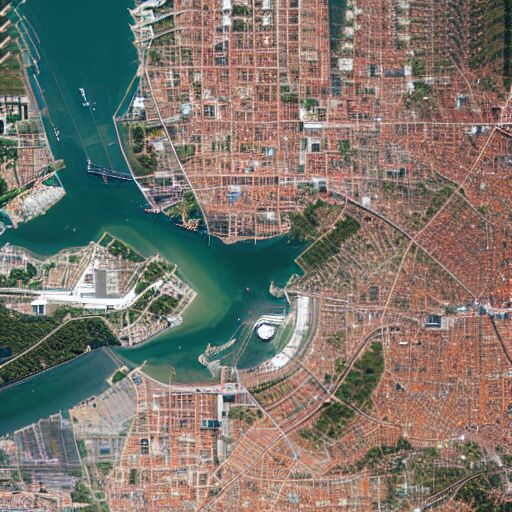}
        \includegraphics[width=0.16\textwidth]{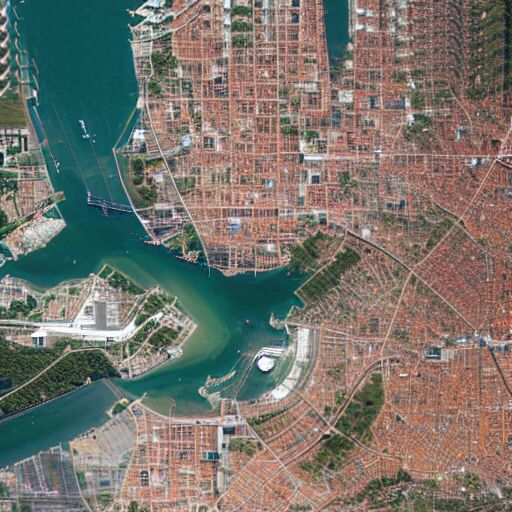}
        \includegraphics[width=0.16\textwidth]{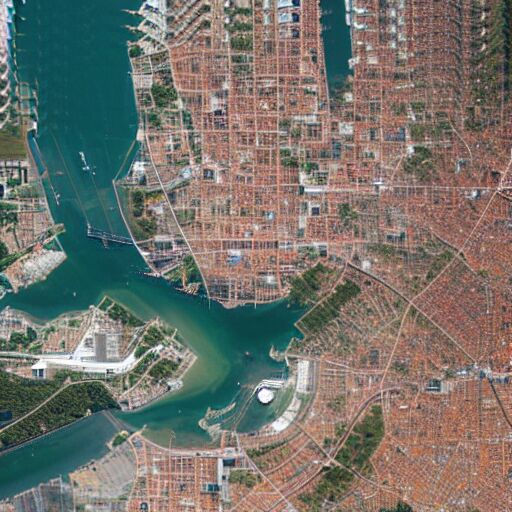}
        \includegraphics[width=0.16\textwidth]{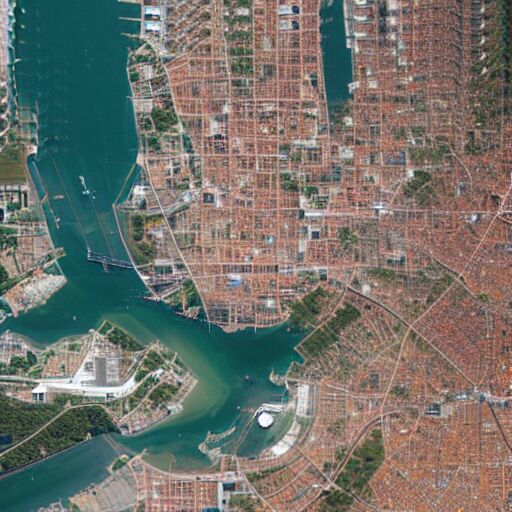}
    \end{subfigure}\hfill
    \begin{subfigure}[t]{\textwidth}
        \centering
        \includegraphics[width=0.16\textwidth]{figures/final/satellite/frame_000_2}
        \includegraphics[width=0.16\textwidth]{figures/final/satellite/frame_004_2}
        \includegraphics[width=0.16\textwidth]{figures/final/satellite/frame_008_2}
        \includegraphics[width=0.16\textwidth]{figures/final/satellite/frame_012_2}
        \includegraphics[width=0.16\textwidth]{figures/final/satellite/frame_016_2}
        \includegraphics[width=0.16\textwidth]{figures/final/satellite/frame_019_2}
    \end{subfigure}
    \caption{Ablation - Spatial-$\eta$. First Row: Spatial-$\eta$ on; Second Row: $\eta = 0$.}
    \label{fig:satellite_abl_spatialeta}
\end{figure*}

\begin{figure*}[!ht]
    \centering
    \begin{subfigure}[t]{\textwidth}
        \centering
        \includegraphics[width=0.16\textwidth]{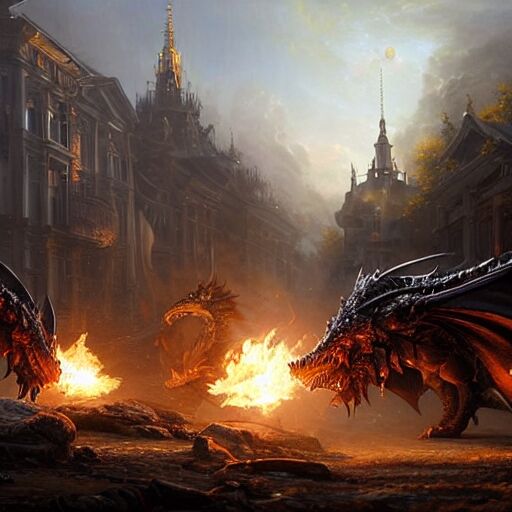}
        \includegraphics[width=0.16\textwidth]{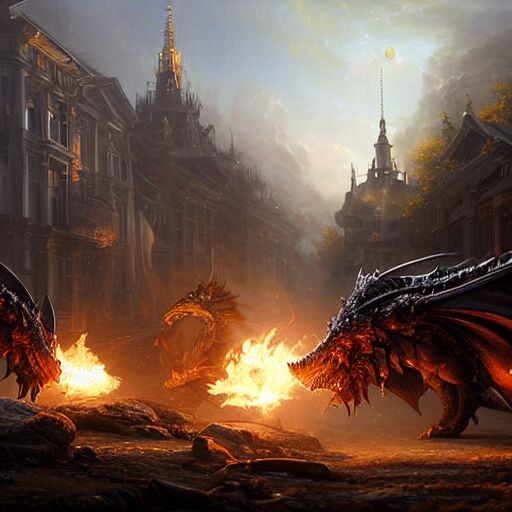}
        \includegraphics[width=0.16\textwidth]{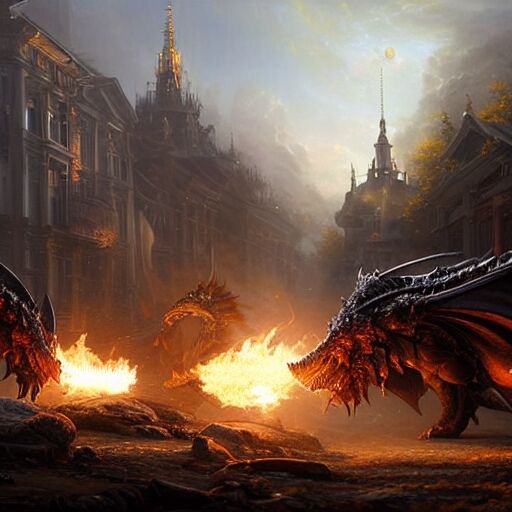}
        \includegraphics[width=0.16\textwidth]{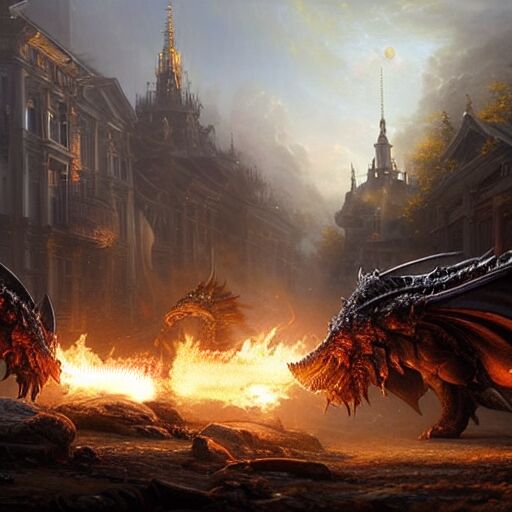}
        \includegraphics[width=0.16\textwidth]{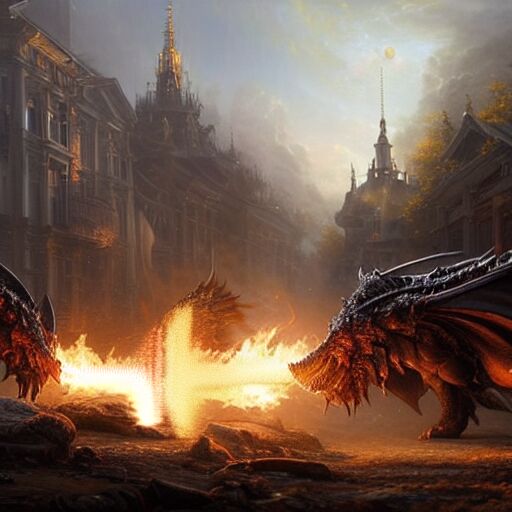}
        \includegraphics[width=0.16\textwidth]{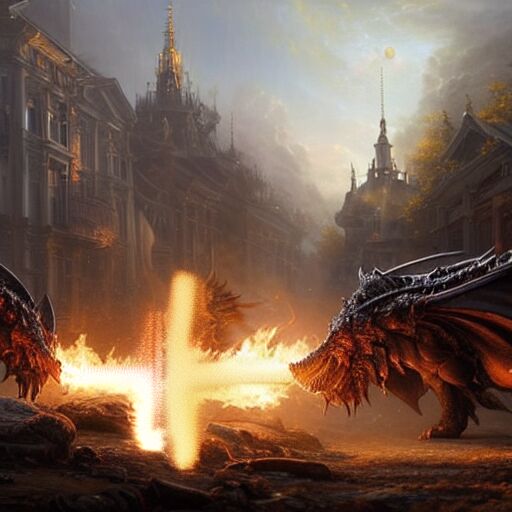}
    \end{subfigure}\hfill
    \begin{subfigure}[t]{\textwidth}
        \centering
        \includegraphics[width=0.16\textwidth]{figures/final/dragons/frame_000_2}
        \includegraphics[width=0.16\textwidth]{figures/final/dragons/frame_004_2}
        \includegraphics[width=0.16\textwidth]{figures/final/dragons/frame_008_2}
        \includegraphics[width=0.16\textwidth]{figures/final/dragons/frame_012_2}
        \includegraphics[width=0.16\textwidth]{figures/final/dragons/frame_016_2}
        \includegraphics[width=0.16\textwidth]{figures/final/dragons/frame_020_2}
    \end{subfigure}
    \caption{Ablation - Spatial-$\eta$. First Row: Spatial-$\eta$ on; Second Row: $\eta = 0$.}
    \label{fig:dragons_abl_spatialeta}
\end{figure*}

\begin{figure*}[!ht]
    \centering
    \begin{subfigure}[t]{\textwidth}
        \centering
        \includegraphics[width=0.16\textwidth]{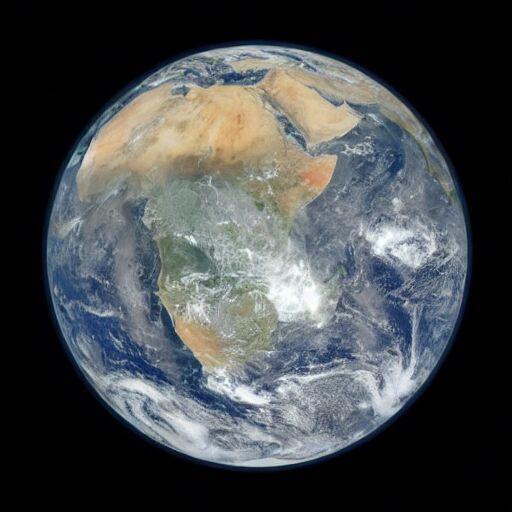}
        \includegraphics[width=0.16\textwidth]{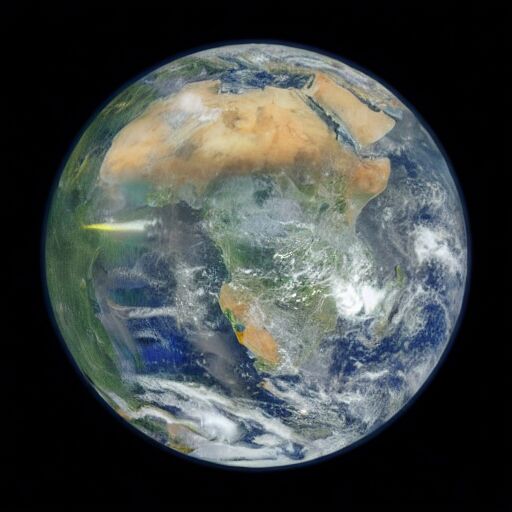}
        \includegraphics[width=0.16\textwidth]{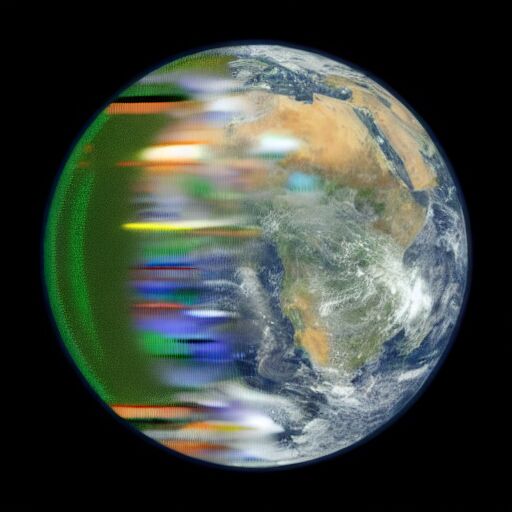}
        \includegraphics[width=0.16\textwidth]{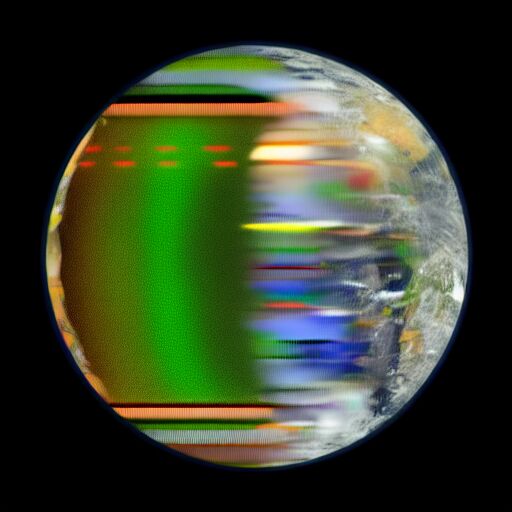}
        \includegraphics[width=0.16\textwidth]{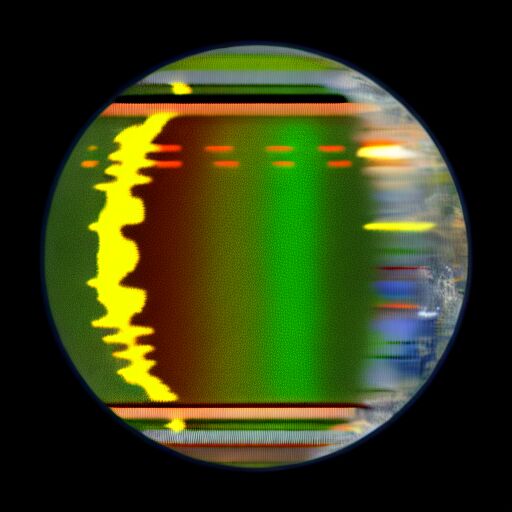}
        \includegraphics[width=0.16\textwidth]{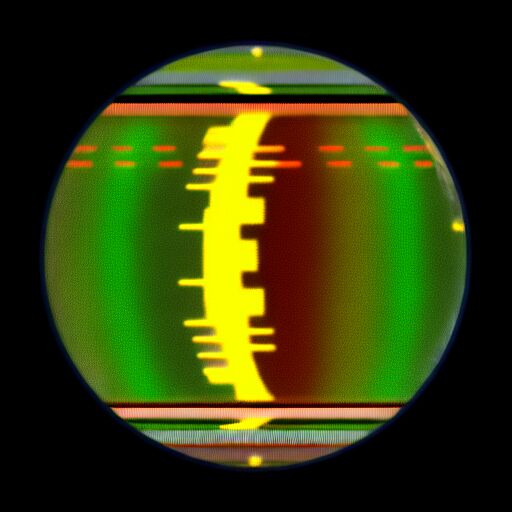}
    \end{subfigure}\hfill
    \begin{subfigure}[t]{\textwidth}
        \centering
        \includegraphics[width=0.16\textwidth]{figures/final/earth/frame_000_2}
        \includegraphics[width=0.16\textwidth]{figures/final/earth/frame_001_2}
        \includegraphics[width=0.16\textwidth]{figures/final/earth/frame_002_2}
        \includegraphics[width=0.16\textwidth]{figures/final/earth/frame_004_2}
        \includegraphics[width=0.16\textwidth]{figures/final/earth/frame_006_2}
        \includegraphics[width=0.16\textwidth]{figures/final/earth/frame_008_2}
    \end{subfigure}\hfill
    \caption{Ablation - Inversion Mechanism. First Row: Without Inversion; Second Row: With Inversion}
    \label{fig:earth_abl_inv}
\end{figure*}

\begin{figure*}[!ht]
    \centering
    \begin{subfigure}[t]{\textwidth}
        \centering
        \includegraphics[width=0.16\textwidth]{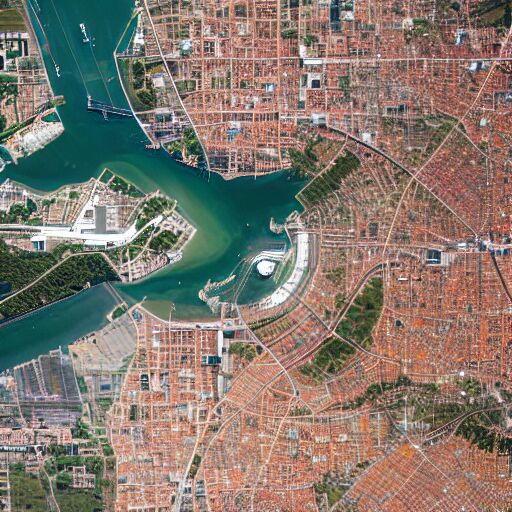}
        \includegraphics[width=0.16\textwidth]{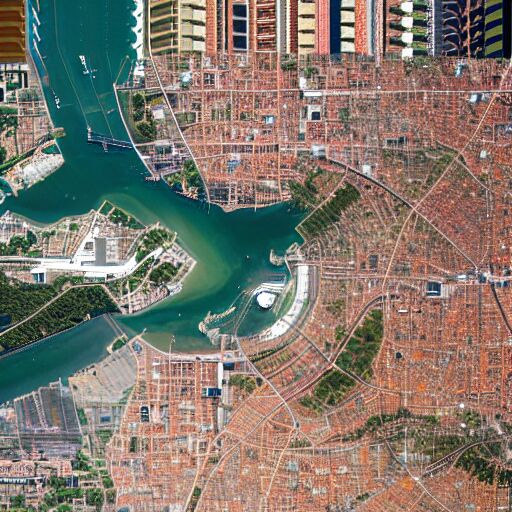}
        \includegraphics[width=0.16\textwidth]{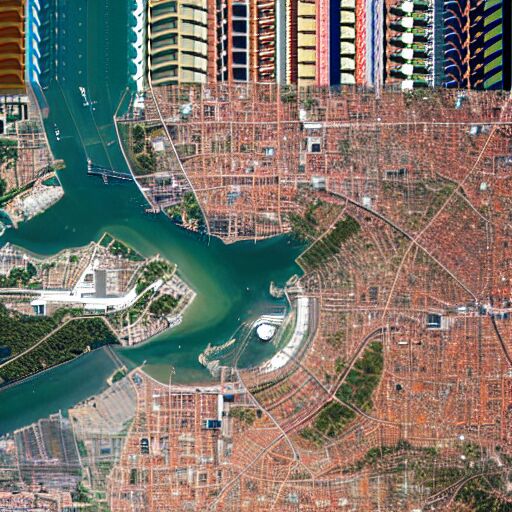}
        \includegraphics[width=0.16\textwidth]{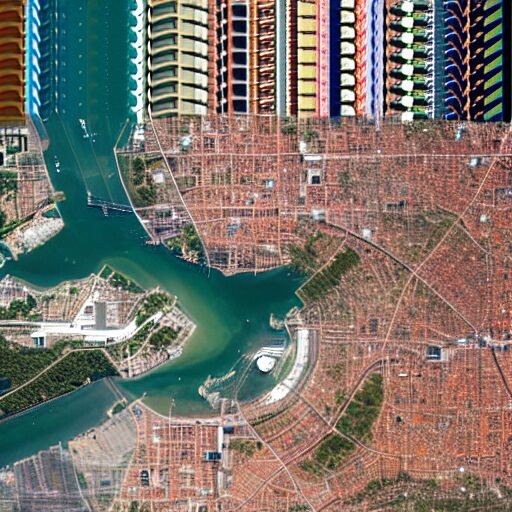}
        \includegraphics[width=0.16\textwidth]{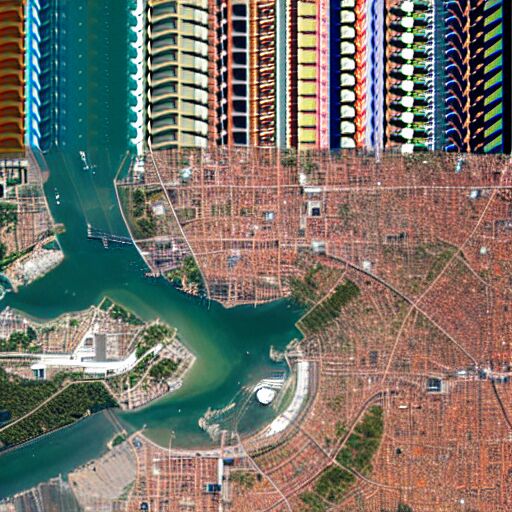}
        \includegraphics[width=0.16\textwidth]{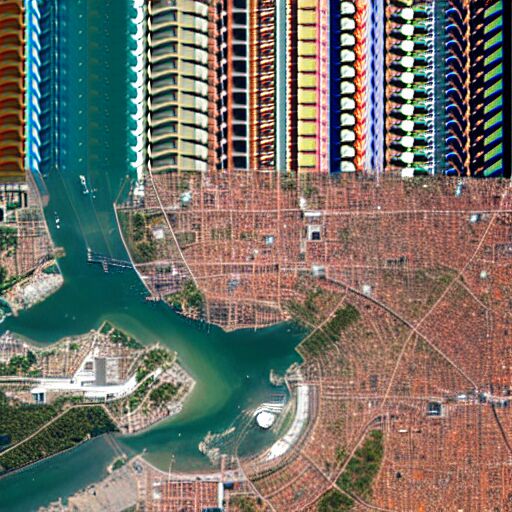}
        \end{subfigure}\hfill
    \begin{subfigure}[t]{\textwidth}
        \centering
        \includegraphics[width=0.16\textwidth]{figures/final/satellite/frame_000_2}
        \includegraphics[width=0.16\textwidth]{figures/final/satellite/frame_004_2}
        \includegraphics[width=0.16\textwidth]{figures/final/satellite/frame_008_2}
        \includegraphics[width=0.16\textwidth]{figures/final/satellite/frame_012_2}
        \includegraphics[width=0.16\textwidth]{figures/final/satellite/frame_016_2}
        \includegraphics[width=0.16\textwidth]{figures/final/satellite/frame_019_2}
    \end{subfigure}\hfill
    \caption{Ablation - Inversion Mechanism. First Row: Without Inversion; Second Row: With Inversion}
    \label{fig:satellite_abl_inv}
\end{figure*}

\begin{figure*}[!ht]
    \centering
    \begin{subfigure}[t]{\textwidth}
        \centering
        \includegraphics[width=0.16\textwidth]{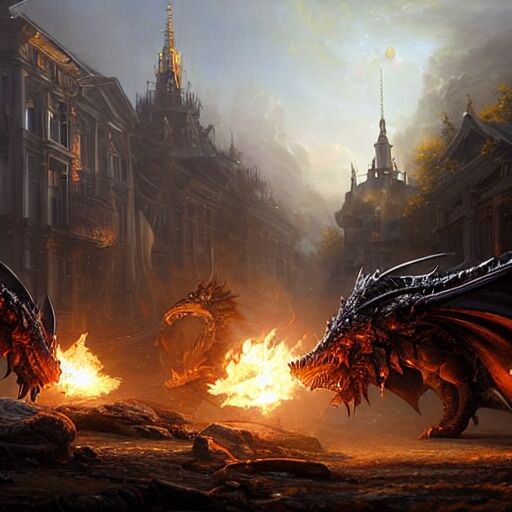}
        \includegraphics[width=0.16\textwidth]{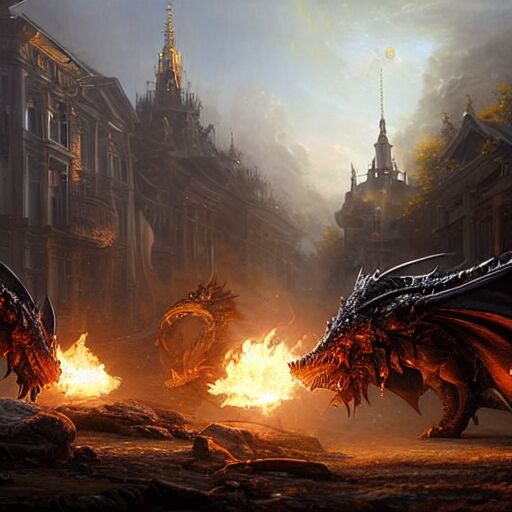}
        \includegraphics[width=0.16\textwidth]{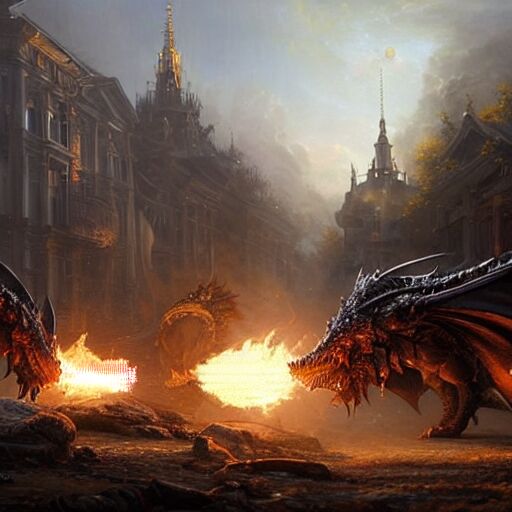}
        \includegraphics[width=0.16\textwidth]{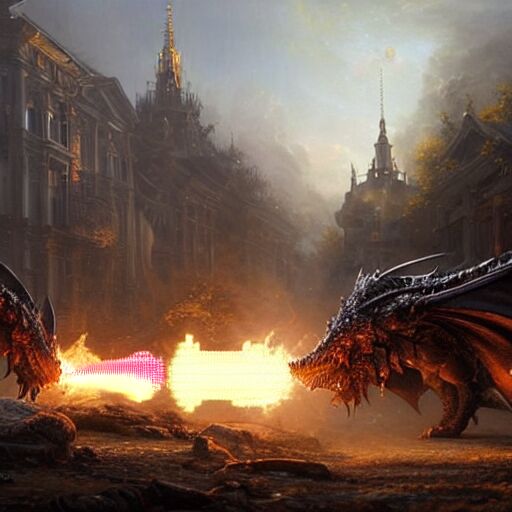}
        \includegraphics[width=0.16\textwidth]{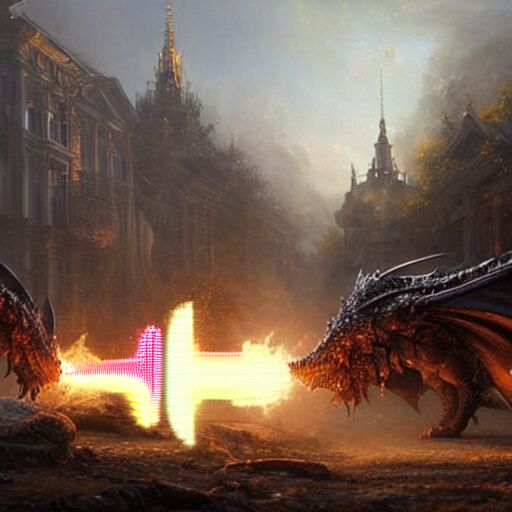}
        \includegraphics[width=0.16\textwidth]{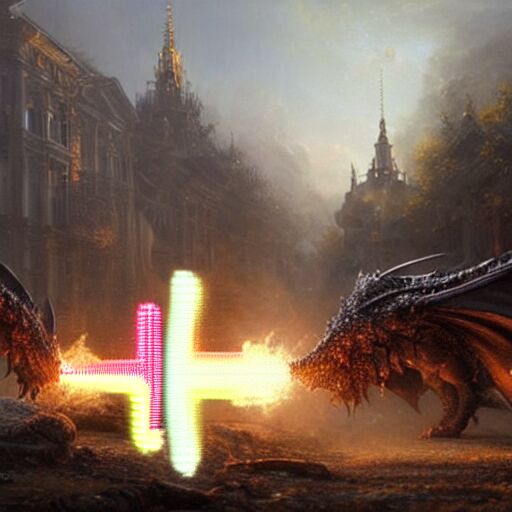}
    \end{subfigure}\hfill
    \begin{subfigure}[t]{\textwidth}
        \centering
        \includegraphics[width=0.16\textwidth]{figures/final/dragons/frame_000_2}
        \includegraphics[width=0.16\textwidth]{figures/final/dragons/frame_004_2}
        \includegraphics[width=0.16\textwidth]{figures/final/dragons/frame_008_2}
        \includegraphics[width=0.16\textwidth]{figures/final/dragons/frame_012_2}
        \includegraphics[width=0.16\textwidth]{figures/final/dragons/frame_016_2}
        \includegraphics[width=0.16\textwidth]{figures/final/dragons/frame_020_2}
    \end{subfigure}\hfill
    \caption{Ablation - Inversion Mechanism. First Row: Without Inversion; Second Row: With Inversion}
    \label{fig:dragons_abl_inv}
\end{figure*}

In this section we show the remaining ablations for the scene \textit{Earth}, and additional ablations on two new scenes: \textit{Dragons} and \textit{Satellite Scan}.
The ablations for cross frame attention mechanism can be found in Figures \ref{fig:satellite_abl_cross} and \ref{fig:dragons_abl_cross}.
The ablations of the Spatial-$\eta$ are shown in Figures \ref{fig:satellite_abl_spatialeta} and \ref{fig:dragons_abl_spatialeta}.
Moreover, we also show the contribution of the inversion mechanism in Figures \ref{fig:earth_abl_inv}, \ref{fig:satellite_abl_inv}, and \ref{fig:dragons_abl_inv}.

\clearpage
\subsection{Multiple Cross-Frame Attention Mechanism Ablation}

\clearpage
\subsection{Spatial eta}

\clearpage
\subsection{Inversion}
\label{sec:abl_inv}

\clearpage
\section{Obstacles, Different Physics and Additional Visual Results}
\label{sec:extraphysics}

In this section we showcase additional visual results of our method; all the generated videos can be found in the \textit{Supplementary Material}. In Fig. \ref{fig:glass} we show an example of a poured glass with \ourmethod (third row) and applying the same flow in image space (fourth row). 
In the first two rows of Fig. \ref{fig:glass} we show the results of the $\Phi$-flow physics simulator.
Note that we simulate both the fluid as a set of particles (\textit{Eulerian} simulations) in a specific position (blue balls in the first row of the figure) and two obstacles (orange objects) representing the glass and the jug. The corresponding optical flow that we used in \ourmethod is visualized in the second row of the figure.

As it can be seen, the optical flow applied to the image space produces some artefacts, such as deformations of the glass and the smoothness of the liquid due to the stretching of the pixels. On the other hand, when the same flow is applied to the noisy latent space through our method, the resulted video appears more realistic, avoiding such deformations. 

\begin{figure*}[!ht]
\centering

\begin{subfigure}[t]{\textwidth}
    \centering
    \includegraphics[width=0.16\textwidth]{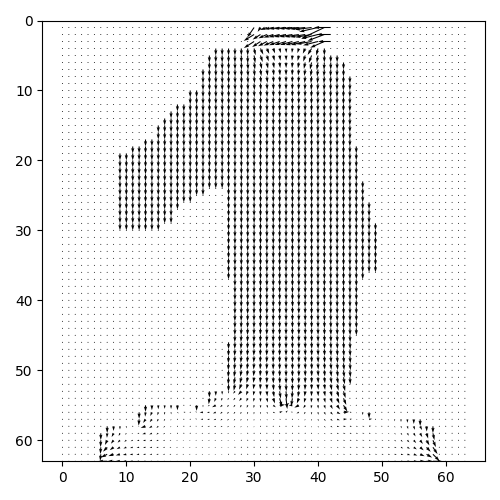}
    \includegraphics[width=0.16\textwidth]{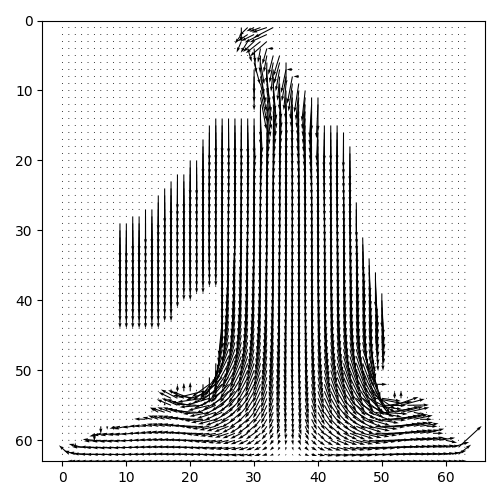}
    \includegraphics[width=0.16\textwidth]{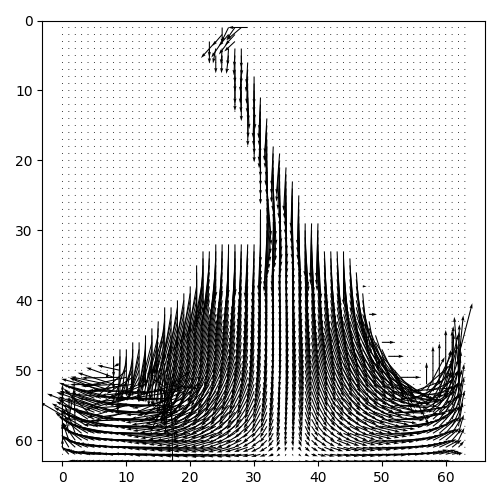}
    \includegraphics[width=0.16\textwidth]{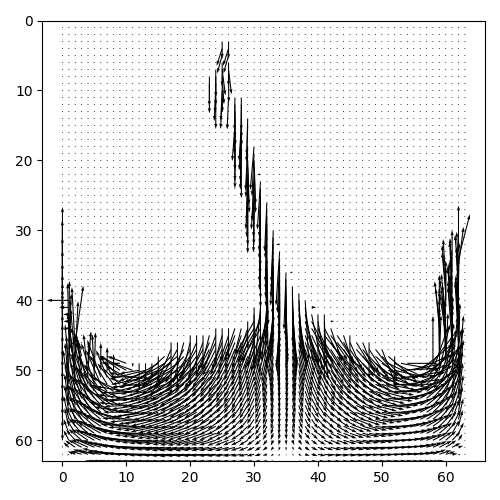}
    \includegraphics[width=0.16\textwidth]{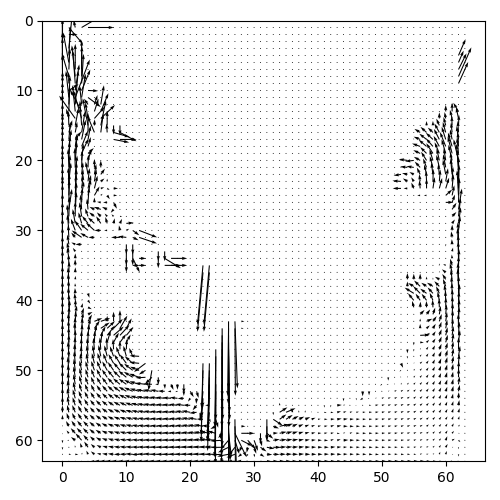}
    \includegraphics[width=0.16\textwidth]{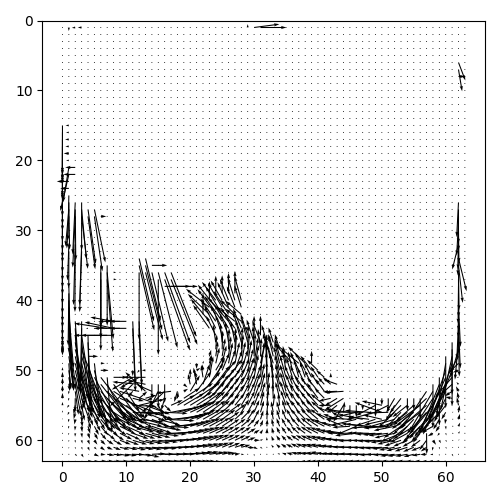}
\end{subfigure}\hfill

\begin{subfigure}[t]{\textwidth}
    \centering
    \includegraphics[width=0.16\textwidth]{figures/final/meltingman/frame_000_2}
    \includegraphics[width=0.16\textwidth]{figures/final/meltingman/frame_004_2}
    \includegraphics[width=0.16\textwidth]{figures/final/meltingman/frame_007_2}
    \includegraphics[width=0.16\textwidth]{figures/final/meltingman/frame_009_2}
    \includegraphics[width=0.16\textwidth]{figures/final/meltingman/frame_015_2}
    \includegraphics[width=0.16\textwidth]{figures/final/meltingman/frame_027_2}
\end{subfigure}\hfill

\begin{subfigure}[t]{\textwidth}
    \centering
    \includegraphics[width=0.16\textwidth]{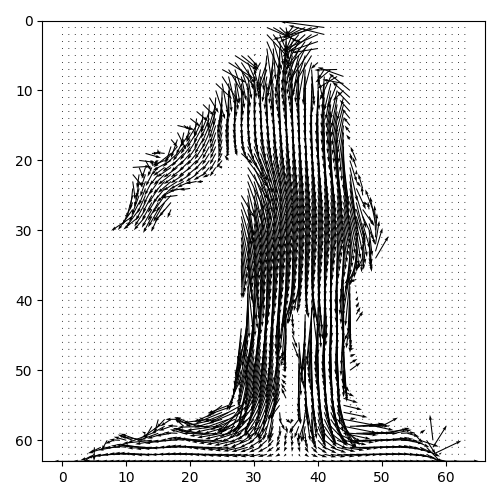}
    \includegraphics[width=0.16\textwidth]{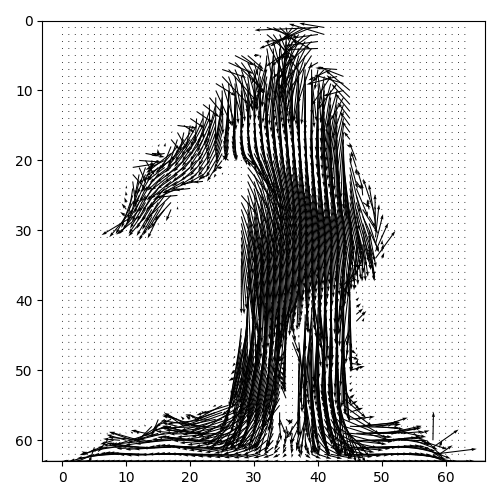}
    \includegraphics[width=0.16\textwidth]{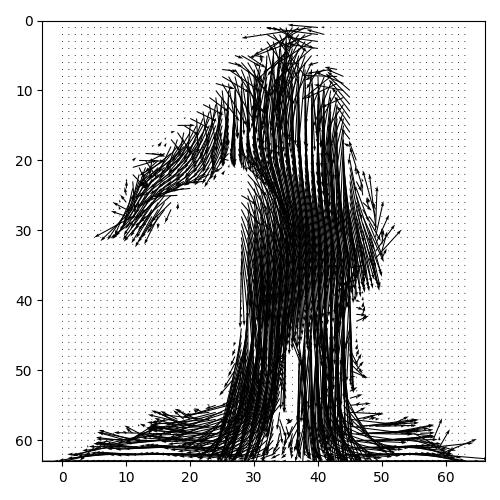}
    \includegraphics[width=0.16\textwidth]{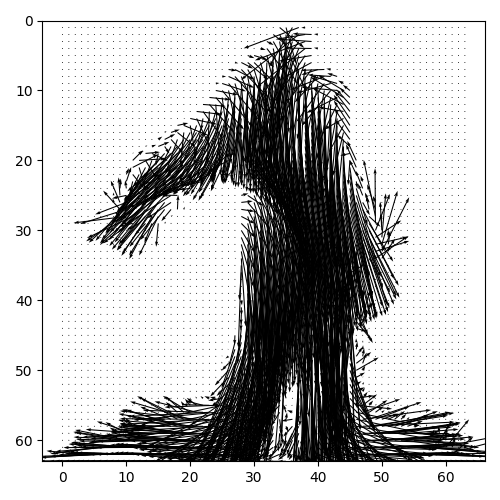}
    \includegraphics[width=0.16\textwidth]{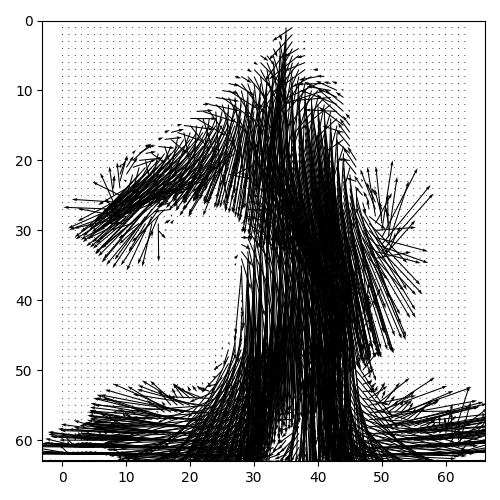}
    \includegraphics[width=0.16\textwidth]{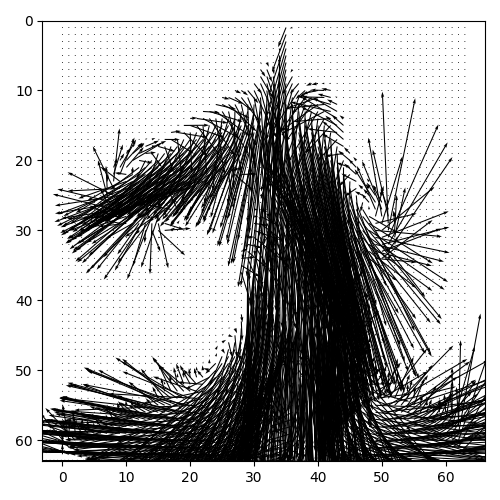}
\end{subfigure}\hfill

\begin{subfigure}[t]{\textwidth}
    \centering
    \includegraphics[width=0.16\textwidth]{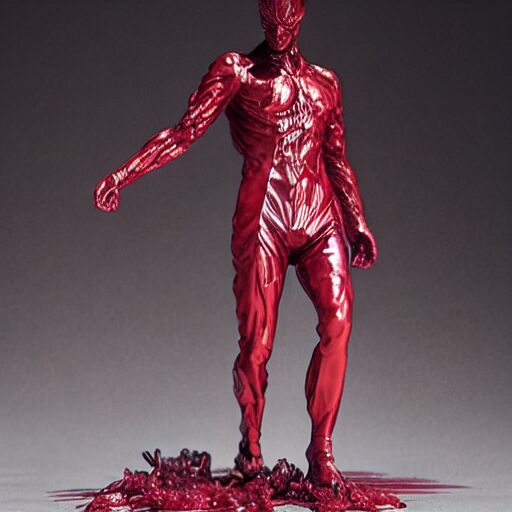}
    \includegraphics[width=0.16\textwidth]{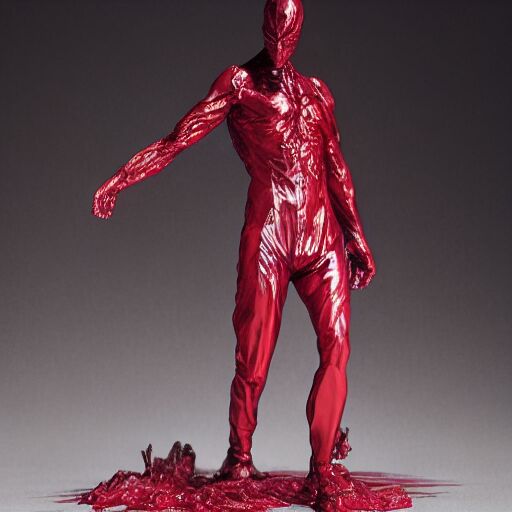}
    \includegraphics[width=0.16\textwidth]{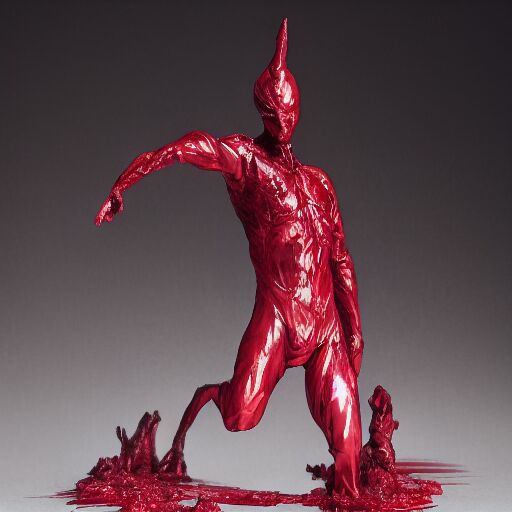}
    \includegraphics[width=0.16\textwidth]{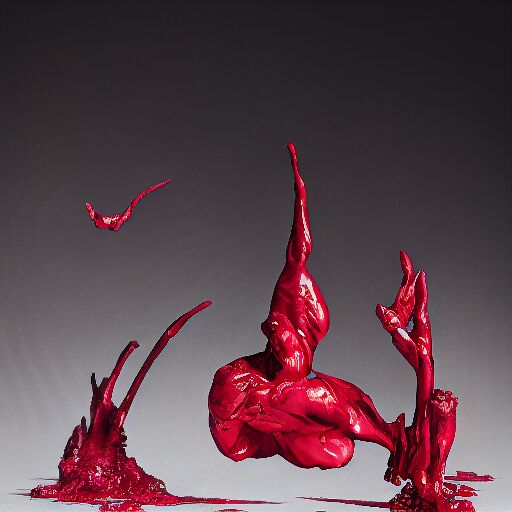}
    \includegraphics[width=0.16\textwidth]{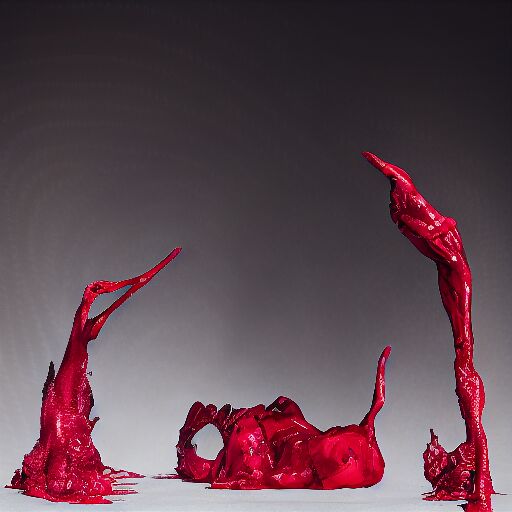}
    \includegraphics[width=0.16\textwidth]{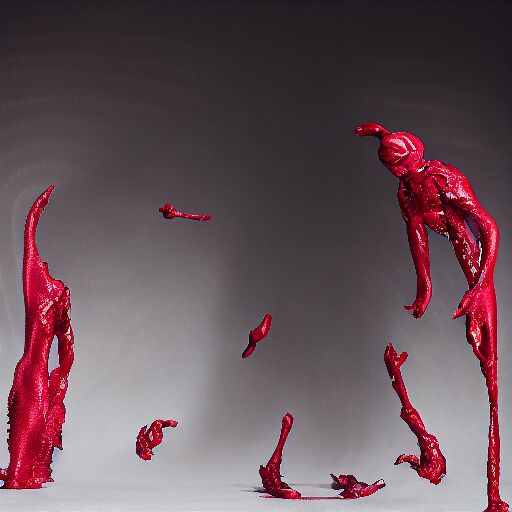}
\end{subfigure}\hfill

\caption{Smoke simulation: Evaporating man. First and second row: Optical flow and video generated by \ourmethod with an Eulerian simulation. Third and fourth row: Optical flow and video generated by \ourmethod with a Lagrangian simulation.}
\label{fig:evaporatingman}
\end{figure*}

Since $\Phi$-flow is able to adopt both \textit{Eulerian} and \textit{Lagrangian} numerical solvers, we show the corresponding videos in Fig. \ref{fig:evaporatingman} (second and fourth row).
While the former decomposes the fluid in a set of particles, the latter models the fluid in the entire space as a fluid field. In both cases we extract from the simulation the (eventually extrapolated) velocity field (first and third row in Fig. \ref{fig:evaporatingman}) and we use it as the optical flow in the latent space, resulting in two different videos.

\begin{figure*}[!ht]
\centering
\begin{subfigure}[t]{\textwidth}
    \centering
    \includegraphics[width=0.16\textwidth]{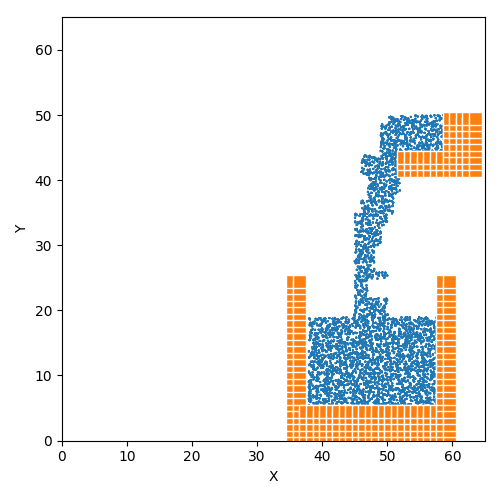}
    \includegraphics[width=0.16\textwidth]{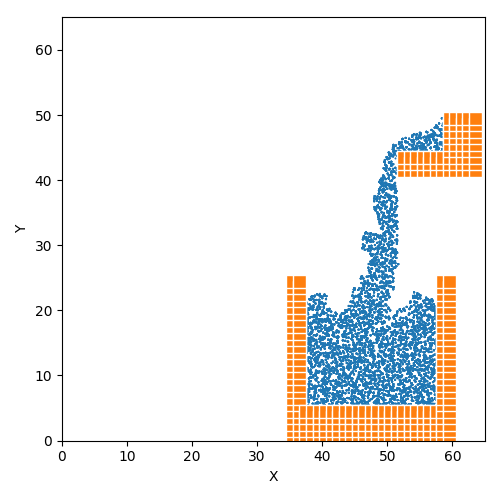}
    \includegraphics[width=0.16\textwidth]{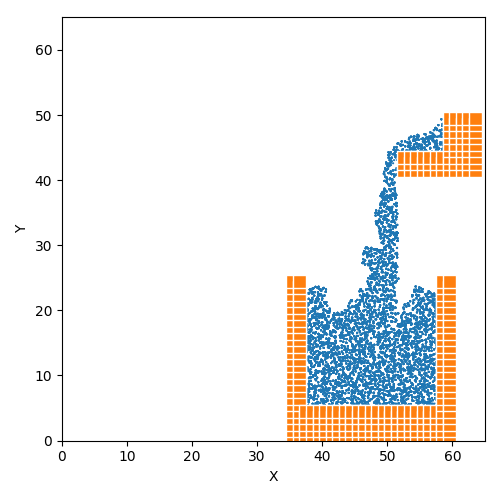}
    \includegraphics[width=0.16\textwidth]{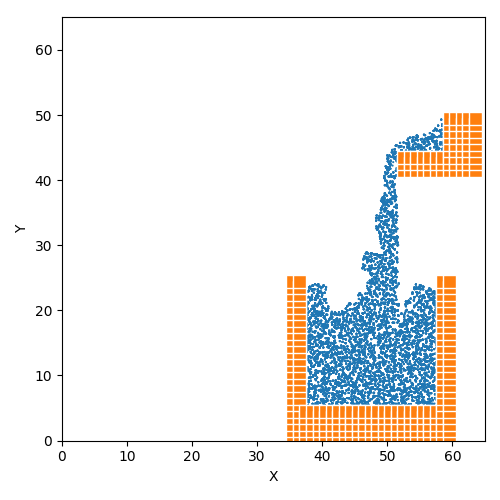}
    \includegraphics[width=0.16\textwidth]{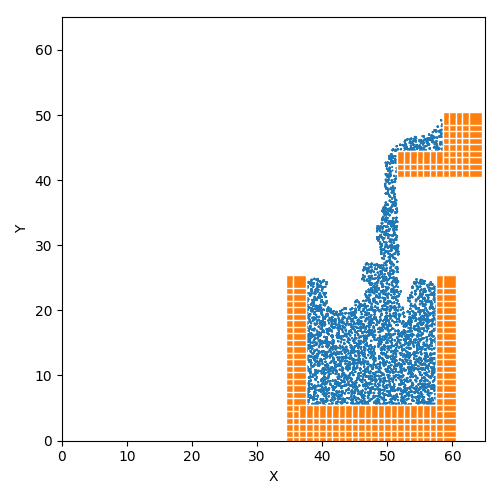}
    \includegraphics[width=0.16\textwidth]{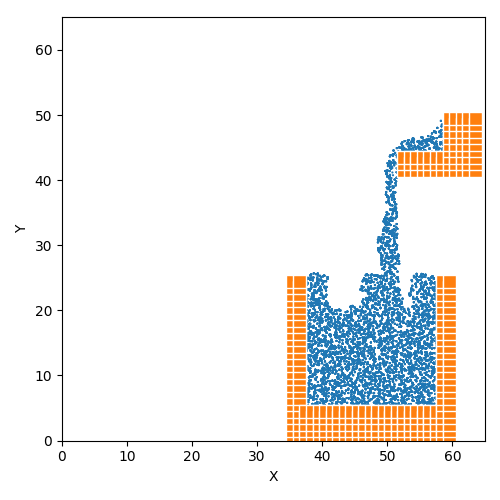}
\end{subfigure}\hfill

\begin{subfigure}[t]{\textwidth}
    \centering
    \includegraphics[width=0.16\textwidth]{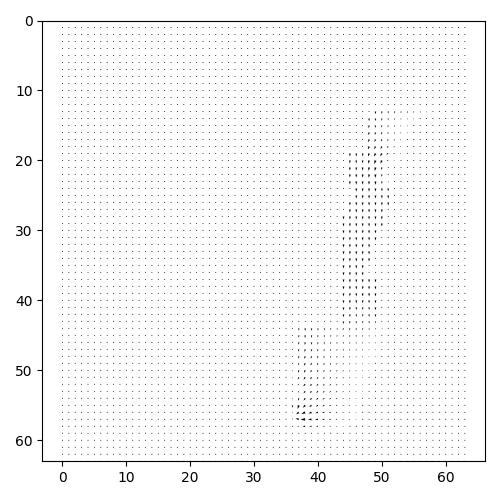}
    \includegraphics[width=0.16\textwidth]{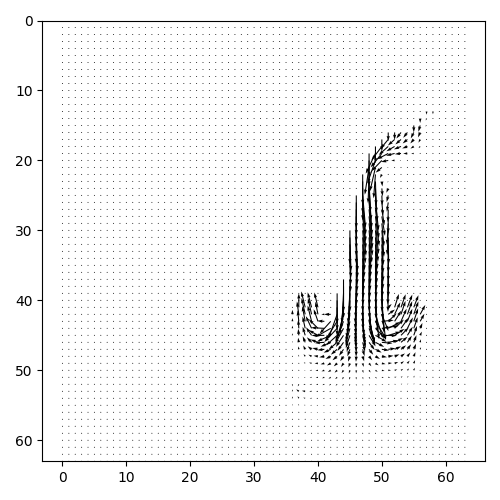}
    \includegraphics[width=0.16\textwidth]{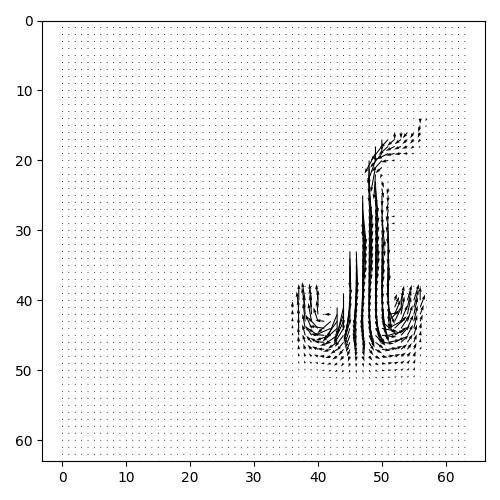}
    \includegraphics[width=0.16\textwidth]{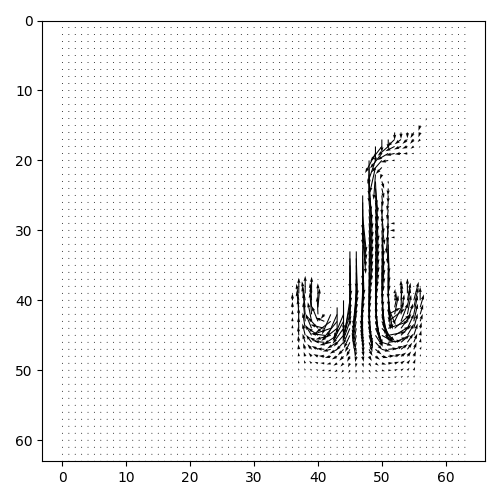}
    \includegraphics[width=0.16\textwidth]{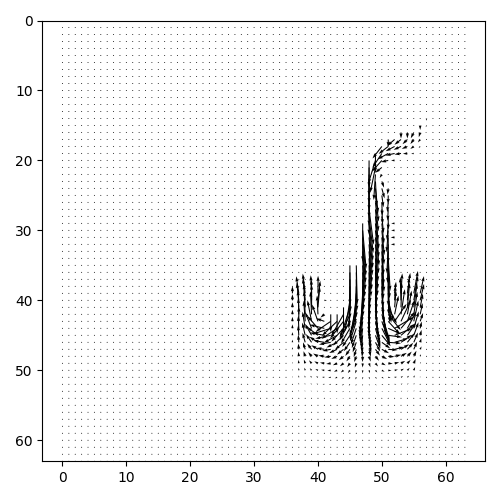}
    \includegraphics[width=0.16\textwidth]{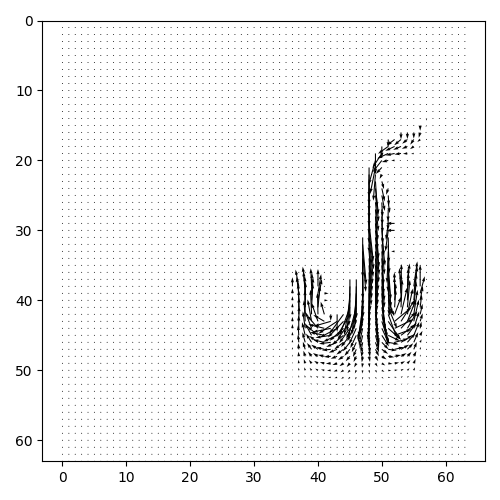}
\end{subfigure}\hfill

\begin{subfigure}[t]{\textwidth}
    \centering
    \includegraphics[width=0.16\textwidth]{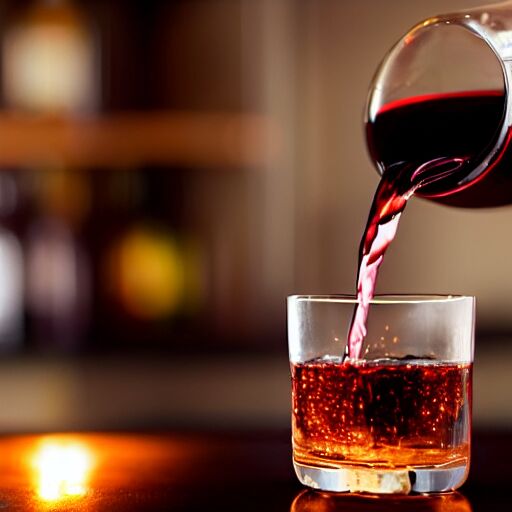}
    \includegraphics[width=0.16\textwidth]{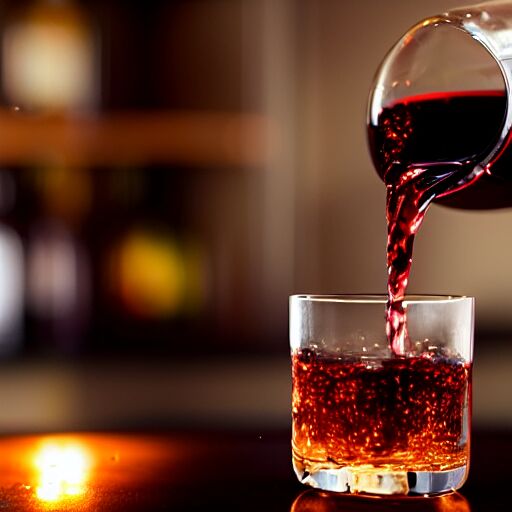}
    \includegraphics[width=0.16\textwidth]{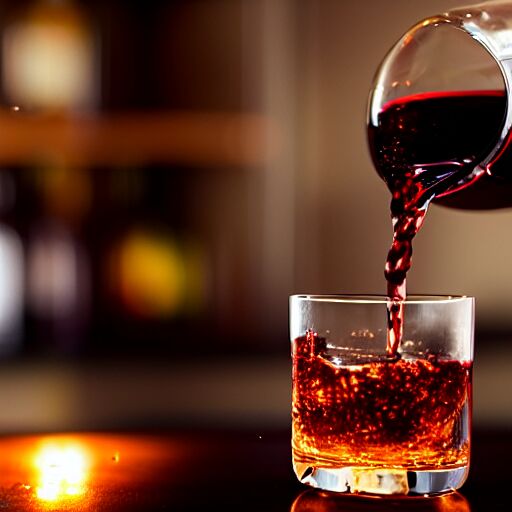}
    \includegraphics[width=0.16\textwidth]{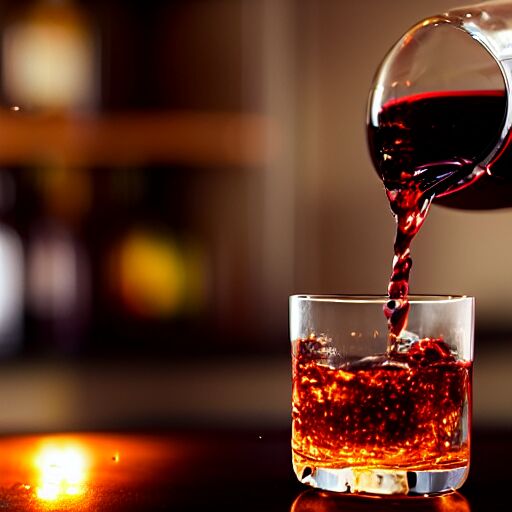}
    \includegraphics[width=0.16\textwidth]{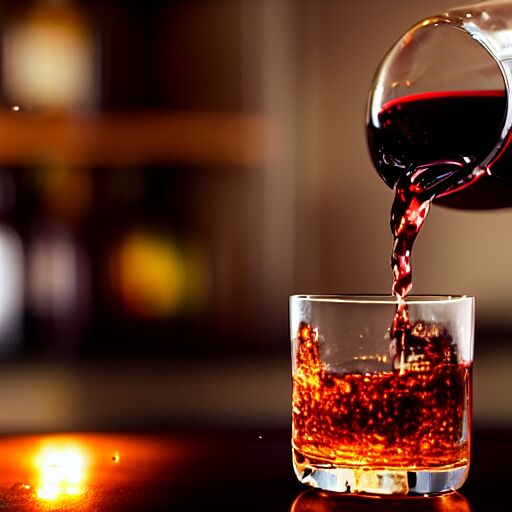}
    \includegraphics[width=0.16\textwidth]{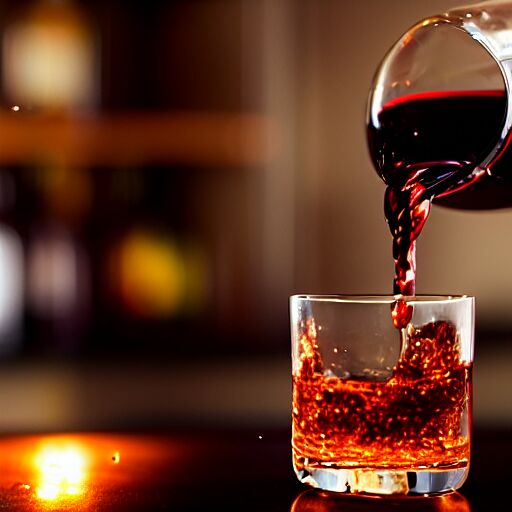}
\end{subfigure}\hfill

\begin{subfigure}[t]{\textwidth}
    \centering
    \includegraphics[width=0.16\textwidth]{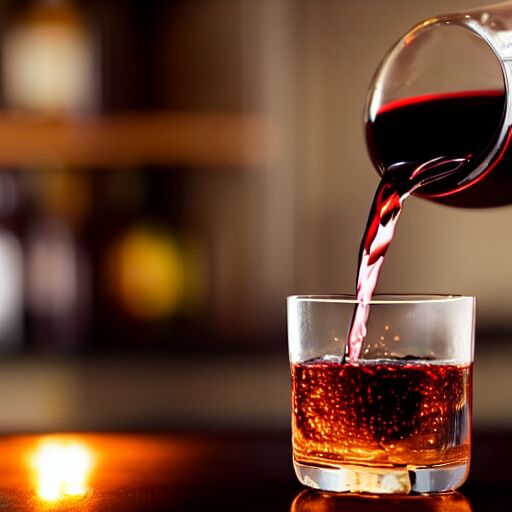}
    \includegraphics[width=0.16\textwidth]{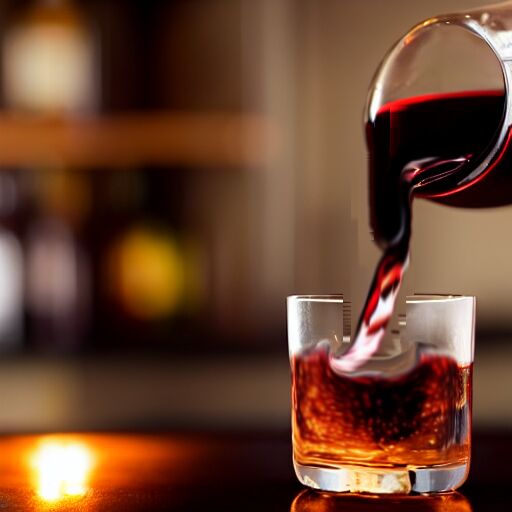}
    \includegraphics[width=0.16\textwidth]{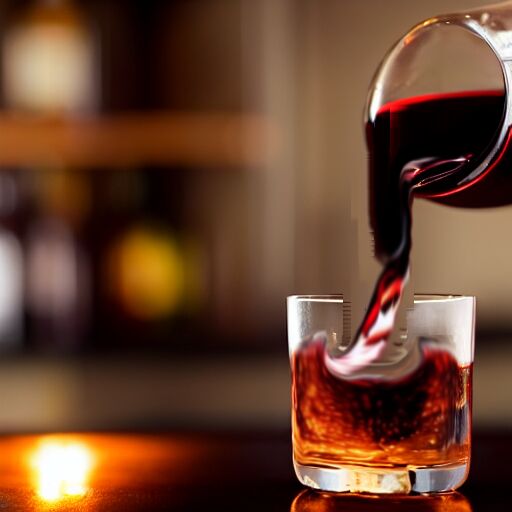}
    \includegraphics[width=0.16\textwidth]{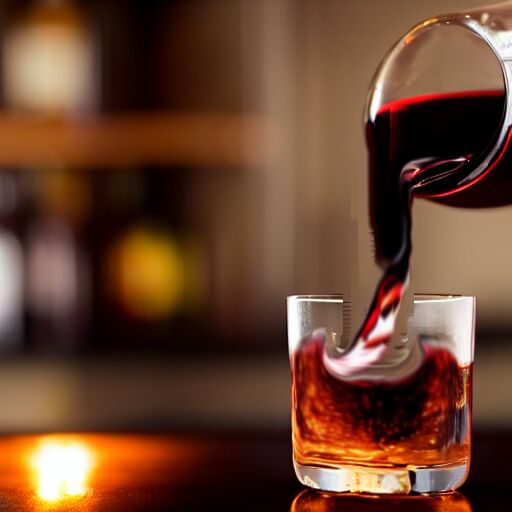}
    \includegraphics[width=0.16\textwidth]{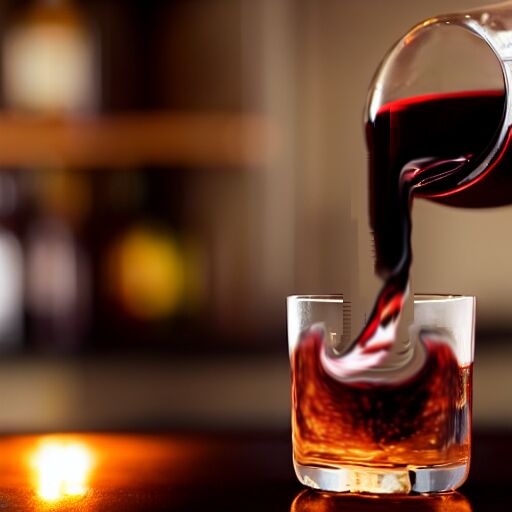}
    \includegraphics[width=0.16\textwidth]{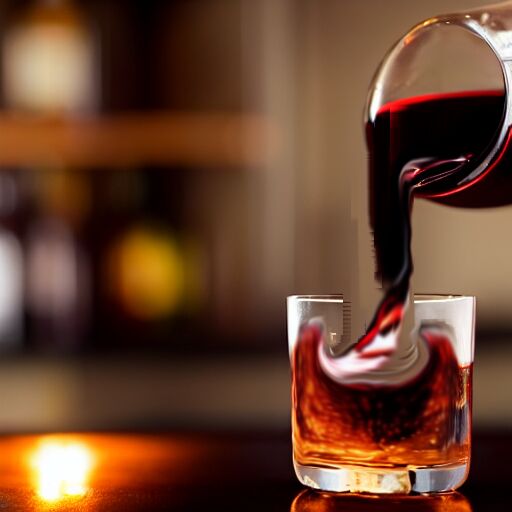}
\end{subfigure}

\caption{Fluid simulation: pouring drink. First row: Eulerian simulation performed with $\Phi$-flow; Second row: resulting optical flow; Third row: \ourmethod; Fourth row: resulting video when optical flow is applied directly to pixel-space.}
\label{fig:glass}
\end{figure*}

\clearpage
\section{Text Prompts}
\label{sec:prompt}

In this section we state the text prompts used in the generated videos for our method and T2V0.
Note that while \ourmethod is able to start from a real or generated image (with almost zero error for the real image reconstruction),  T2V0 needs a hyper-parameters tuning due to a high guidance scale (not supporting direct inversion of real images).

\begin{itemize}
    \item \textit{Fighting Dragons}: \virg{Two dragons fighting while breathing fires to each other. The flames are blazing and majestic light. Theatrical, character concept art by ruan jia, thomas kinkade, and  trending on Artstation.}
    \item \textit{Melting Man} (both versions): \virg{transparent man made by water and smoke, in style of Yoji Shinkawa and Hyung-tae Kim, trending on ArtStation, dark fantasy, great composition, concept art, highly  human  made of water and foam, in the style of Pierre Koenig, red pigment, pastel paint, pink color scheme}
    \item \textit{Satellite Scan}: \virg{a satellite image of a city}
    \item \textit{Revolving Earth}: \virg{a close up of a picture of the earth from space.}
    \item \textit{Flock of birds}: \virg{a small flock bird flying in the sky at the sunset}
    \item \textit{Pouring drink}: \virg{wine falling on a empty glass}
\end{itemize}
For the text prompts of \textit{Fighting Dragons} and \textit{Melting Man}  we leveraged MagicPrompt (for which we credit Gustavo Santana), a tool for rewriting simple text prompts to create more appealing starting images with Stable Diffusion.

For each example, the negative prompt $\mathcal{P}_{\emptyset}$ is equal to \virg{poorly drawn, cartoon, 2d, disfigured, bad art, deformed, poorly drawn, extra limbs, close up, b\&w, weird colors, blurry}

\section{Limitations and future work}
\label{sec:limitations}

In this section we discuss the limitations of the proposed approach. Being a zero-shot approach, \ourmethod relies on the pretrained text-to-image model, i.e. Stable Diffusion, and it can inherit some limitations from it, such as not exact DDIM inversion. Hence,  by exploiting other diffusion models we could improve our method as well. \\
Experimentally, we observed a global color shift, getting stronger in the last frames of the generated videos. We noted that the proposed MCFA strategy partially solved this, but a better solution could be attending to all the previous generated frames (albeit resulting in a memory and run-time complexity increase).  
Moreover, \ourmethod depends on the optical flow derived from physics simulations but there are some dynamics that may be difficult to simulate (e.g. the motion of a dancer), thus limiting the generality of the generated videos. However, we speculate that it might be possible to devise a generative model of optical flows conditioned on a starting frames and a prompt, while also being constrained by a physics simulator. This could readily provide inputs to \ourmethod and have the advantage of disentangling learning of motion from learning of content.
A future direction could also employ a better interaction between the image generator and the physics simulator, in order to have a closed feedback-loop framework leading to more physical fidelity in the generated frames. 
In this work we have shown videos generated by different physical simulations, but as future work we could also combine them to generate more complex scenes with different physics mixed together.

\section{Implementation Details and Licenses}
\label{sec:implementation}
We used the following hyperparameters throughout the work if not explicitely said otherwise. We set $\tau=400$, the number of inference steps (both for DDIM inversion and for inverse diffusion) is set to $200$ and the used model is \texttt{runwayml/stable-diffusion-v1-5} (license CreativeML Open RAIL-M).
All our experiments are done on a single NVIDIA A6000 (48GB); video generation runs in minutes (1-5min) on a single GPU. Our provided code is available under MIT license.
The \textit{Earth} image is a composite of six separate orbits taken on January 23, 2012 by the Suomi National Polar-orbiting Partnership satellite (Credit: NASA/NOAA).

\section{Broader Impact}
\label{sec:broader}
Synthetic video generation is a powerful technology that can be misused to create fake videos, hence it is important to limit and safely deploy these models. From a safety perspective, we emphasize that \ourmethod does not add any new restrictions nor does it relax any existing ones with respect to our base text-to-image model. Moreover \ourmethod, using existing text-to-image diffusion models, does not need extra training or adjustments. This means we avoid the large environmental costs associated with training new models. One possible broader impact of \ourmethod is its usage by scientists across various fields to visualize their simulations, thereby offering AI-based visualization of physical processes to a wider scientific audience.

\empty
\newpage
\clearpage



\newpage

\section*{NeurIPS Paper Checklist}

\begin{enumerate}

\item {\bf Claims}
    \item[] Question: Do the main claims made in the abstract and introduction accurately reflect the paper's contributions and scope?
    \item[] Answer: \answerYes{} 
    \item[] Justification: All the claims in the abstract and introduction sections are supported by experimental evidence throughout the paper and reflect the results of our method.

\item {\bf Limitations}
    \item[] Question: Does the paper discuss the limitations of the work performed by the authors?
    \item[] Answer: \answerYes{} 
    \item[] Justification: We discuss the limitations of the proposed work in the Appendix Section \ref{sec:limitations}.

\item {\bf Theory Assumptions and Proofs}
    \item[] Question: For each theoretical result, does the paper provide the full set of assumptions and a complete (and correct) proof?
    \item[] Answer: \answerNA{} 
    \item[] Justification: 

    \item {\bf Experimental Result Reproducibility}
    \item[] Question: Does the paper fully disclose all the information needed to reproduce the main experimental results of the paper to the extent that it affects the main claims and/or conclusions of the paper (regardless of whether the code and data are provided or not)?
    \item[] Answer: \answerYes{} 
    \item[] Justification: We report a detailed pseudocode \ref{alg:H} from which it is possible to reproduce the main experimental results of our work. The method with all hyperparameters and implementation details are present in the Appendix \ref{sec:implementation}.

\item {\bf Open access to data and code}
    \item[] Question: Does the paper provide open access to the data and code, with sufficient instructions to faithfully reproduce the main experimental results, as described in supplemental material?
    \item[] Answer: \answerYes{} 
    \item[] Justification: There is a link in the abstract pointing to the official project page, containing a public version of this code.

\item {\bf Experimental Setting/Details}
    \item[] Question: Does the paper specify all the training and test details (e.g., data splits, hyperparameters, how they were chosen, type of optimizer, etc.) necessary to understand the results?
    \item[] Answer: \answerYes{} 
    \item[] Justification: We specify the experimental setting in section \ref{subsec:experimental} and in the Appendix \ref{sec:implementation}

\item {\bf Experiment Statistical Significance}
    \item[] Question: Does the paper report error bars suitably and correctly defined or other appropriate information about the statistical significance of the experiments?
    \item[] Answer: \answerYes{} 
    \item[] Justification: We report mean and std of the results in Table \ref{tab:metrics}.

\item {\bf Experiments Compute Resources}
    \item[] Question: For each experiment, does the paper provide sufficient information on the computer resources (type of compute workers, memory, time of execution) needed to reproduce the experiments?
    \item[] Answer: \answerYes{}
    \item[] Justification: We specify the compute resource and time of execution in Appendix \ref{sec:implementation}
    
\item {\bf Code Of Ethics}
    \item[] Question: Does the research conducted in the paper conform, in every respect, with the NeurIPS Code of Ethics \url{https://neurips.cc/public/EthicsGuidelines}?
    \item[] Answer: \answerYes{} 
    \item[] Justification: The work respects the Code of Ethics.

\item {\bf Broader Impacts}
    \item[] Question: Does the paper discuss both potential positive societal impacts and negative societal impacts of the work performed?
    \item[] Answer: \answerYes{} 
    \item[] Justification: We discuss the broader impact of the work in Appendix \ref{sec:broader} 
    
\item {\bf Safeguards}
    \item[] Question: Does the paper describe safeguards that have been put in place for responsible release of data or models that have a high risk for misuse (e.g., pretrained language models, image generators, or scraped datasets)?
    \item[] Answer: \answerNA{} 
    \item[] Justification: 

\item {\bf Licenses for existing assets}
    \item[] Question: Are the creators or original owners of assets (e.g., code, data, models), used in the paper, properly credited and are the license and terms of use explicitly mentioned and properly respected?
    \item[] Answer: \answerYes{} 
    \item[] Justification: We cite all the used assets throughout the work and we report the correct version used for our experiments.

\item {\bf New Assets}
    \item[] Question: Are new assets introduced in the paper well documented and is the documentation provided alongside the assets?
    \item[] Answer: \answerYes{} 
    \item[] Justification: We release the code along with the documentation needed to run the experiments.

\item {\bf Crowdsourcing and Research with Human Subjects}
    \item[] Question: For crowdsourcing experiments and research with human subjects, does the paper include the full text of instructions given to participants and screenshots, if applicable, as well as details about compensation (if any)? 
    \item[] Answer: \answerNA{} 
    \item[] Justification: 

\item {\bf Institutional Review Board (IRB) Approvals or Equivalent for Research with Human Subjects}
    \item[] Question: Does the paper describe potential risks incurred by study participants, whether such risks were disclosed to the subjects, and whether Institutional Review Board (IRB) approvals (or an equivalent approval/review based on the requirements of your country or institution) were obtained?
    \item[] Answer: \answerNA{} 
    \item[] Justification: 

\end{enumerate}

\end{document}